\documentclass[12pt,a4paper]{report}
\usepackage[latin1]{inputenc}
\usepackage{algorithmic}
\usepackage{algorithm}
\usepackage{amsmath}
\usepackage{euscript}
\usepackage{mathrsfs}
\usepackage{amsfonts}
\usepackage{times}
\usepackage{amssymb}
\usepackage{multirow}
\usepackage{graphicx}
\usepackage{colortbl}
\usepackage{lipsum}
\usepackage{eqparbox}
\usepackage{fixltx2e}
\usepackage{stfloats}
\usepackage{caption}
\usepackage{cite}
\usepackage[numbers,sort&compress]{natbib}
\usepackage{url}
\usepackage{hyperref}
\hypersetup{
colorlinks = true,
urlcolor = blue,
linkcolor = blue,
citecolor = blue
}
\usepackage[a4paper,bindingoffset=0.2in,%
            left=0.4in,right=1.2in,top=1.0in,bottom=0.9in,%
            footskip=.25in]{geometry} 
\newcommand*\rot{\rotatebox{90}}
\begin{document}
\begin{titlepage}
\setlength{\unitlength}{1cm}
\begin{center}
\begin{figure}[!ht]
\begin{center}
\label{Neuron}
\end{center}
\end{figure}
\vspace*{70mm}
{\LARGE Bounded Fuzzy Possibilistic Method }\\[0.2cm]
{\LARGE of Critical Objects Processing}\\[0.2cm]
{\LARGE in Machine Learning}\\[4.5cm]
{\LARGE Hossein Yazdani}\\[2.5cm]
{\large  2020}
\end{center}
\end{titlepage}
%
%
%
%
%
%
\textbf{ \large Abstract}\\[0.8cm]
\parbox{15cm}{Unsatisfying accuracy of learning methods is mostly caused by omitting the influence of important parameters such as {\it membership assignments}, {\it type of data objects}, and {\it distance} or {\it similarity functions}. This Thesis proposes a new method, called \textit{Bounded Fuzzy Possibilistic Method} (BFPM) to address different issues that previous clustering or classification methods have not sufficiently considered in their membership assignments. In fuzzy methods, the object's memberships should sum to 1. Hence, any data object may obtain full membership in at most one cluster or class. Possibilistic methods relax this condition, but the method can be satisfied with the results even if just an arbitrary object obtains the membership from just one cluster, which prevents the objects' movement analysis and objects participation in other clusters. Whereas, BFPM differs from previous fuzzy and possibilistic approaches by removing these restrictions. Furthermore, BFPM provides the flexible search space for objects' movement analysis by allowing objects to obtain full membership in multiple or even in all clusters or classes. \\
Data objects are also considered as fundamental keys in learning methods, and knowing the exact type of objects results in providing a suitable environment for learning algorithms. The Thesis introduces a new type of object, called \textit{critical}, as well as categorizing data objects into two different categories: {\it structural-based} and {\it behavioural-based}. Critical objects are considered as causes of misclassification and miss-assignment in learning procedures. The Thesis also proposes new methodologies to study the behaviour of critical objects with the aim of evaluating objects' movements (mutation) from one cluster or class to another. The Thesis also introduces a new type of feature, called  \textit{dominant}, that is considered as one of the causes of misclassification and miss-assignments. Then the Thesis proposes new sets of similarity functions, called \textit{Weighted Feature Distance} (WFD) and \textit{Prioritized Weighted Feature Distance} (PWFD), to cover diversity in the vector and the feature spaces, in addition to handling the impact of dominant features.\\
The functionality of BFPM has been proved in geometry, set theory, and other disciplines, when learning methods should provide the important arithmetic operations on different domains. Different versions of BFPM algorithm have been compared with Fuzzy C-Means (FCM), modified FCMs, and advanced modified centroid-based methods. Validity and comparison indexes have been utilized to evaluate the accuracy of BFPM applied in clustering problems. Conventional fuzzy and possibilistic methods are compared with BFPM and BFPM-WFD, in terms of accuracy, fuzzification constant, objects' movements, different norms, and covering diversity and overlapping. In experimental sections, the proposed method has been applied and analysed in some real problems in medicine, risk management, anomaly detection, and some of the most well-known benchmark datasets. Promising results achieved in experimental research show that the method BFPM, proposed in the Thesis, ensures better accuracy than other learning methods due to taking into account the influences of critical objects and the impact of dominant features, besides considering mutation.}

\newpage
\tableofcontents
\listoffigures
\listoftables
%
\chapter{Introduction}
\label{Introduction-Chap}
\vspace*{-5mm}
\section{Motivation}
\label{Motivation}
Learning methods and mining approaches aim to collect useful knowledge from data in order to learn from discovered patterns. The main concern of such methods is whether the covered knowledge is accurate or in what extent we can rely on the discovered information. It is very important to find the crucial parameters that influence the quality of the discovered knowledge. Evaluating and partitioning data objects can be obtained using different techniques, where model-based method is one of them \cite{one}. In model-based approaches, data objects are categorized into different categories by attempting to optimize the fit between the data and some mathematical models when the data objects are generated by a mixture of underlying probability distributions \cite{two}. In other words, data objects are categorized based on different data models or data patterns. The data model can be extracted from a Gaussian mixture model, a regression-based model, or a proximity-based model \cite{three}. The accuracy of the learner strongly depends on the accuracy of the discovered patterns from data, where accuracy is mostly measured by calculating the  percentage of the correct labelled objects in classification problems and measuring the compactness and separations of objects with respect to clusters in clustering problems \cite{one}, \cite{two}. \\
In supervised and unsupervised methods, assigning memberships to objects is maintained by the learner based on the objects' behaviour and the learning functions \cite{four}, \cite{five}. Most of the learning methods make use of membership and similarity functions to assign the proper memberships to data objects \cite{six}. The assignment is not an easy task as there are different parameters that might affect the result, where overlapping, uncertainty, diversity, mutation, similarity and distance functions, and type of data object are some of the parameters, while overlapping occurs when an object participates in several clusters or the specific object cannot be easily assigned to just one cluster and uncertainty refers to condition that the memberships of any particular object with respect to clusters (or classes) are not completely clear. Diversity (covering the whole search space) in similarity assessments is also considered as evaluating objects with respect to the vector and the feature spaces. Mutation refers to the condition that objects move from one cluster or class to another by getting small changes in their feature spaces.\\
In social networks, each person can be a member of several societies, which these networks are considered as \textit{overlap clusters}. Transactions, online (or offline), should be analysed to check whether transactions are risky or not, or they can be harmful in the future (\textit{uncertainty}). Internet Service Providers (ISP(s)) receive transactions and need to analyze whether they are or will be affecting the security or not. Banking systems also need to evaluate their customers whether they are or will be risky or not. In medicine, people are interested to check their health whether they are healthy, and how likely they are going to be a member of diseases categories (\textit{mutation}).\\
Different approaches have been introduced to extract the most accurate information from datasets \cite{seven}. The methods considered the main parameters that affect the learning procedures such as data type, membership functions, and similarity functions. Crisp, fuzzy, probability, and possibilistic methods are the well-known methods in membership assignments. There are some difficulties to categorize objects in crisp clusters as the categories may overlap \cite{eight}. The most important factors in misclassification and miss-assignments are arisen from the objects that we are uncertain about their behaviour and properties. Fuzzy, probability, and possibilistic methods are introduced to cover uncertainty in learning methods, but there are also some drawbacks with these methods that are addressed in the following chapters.\\
Finding the desirable solutions for the drawbacks of these methods encouraged the author of this Thesis to introduce a new method to overcome the issues. Current fuzzy and probability clustering approaches include a condition that membership values sum to 1, meaning that any data object can obtain full membership in at most one cluster. Possibilistic clustering methods remove this restriction, but require good initializations in their early learning stages. On the other hand, learning approaches make use of similarity functions in their learning procedures and also in membership assignments. But, similarity functions mostly perform on the vector space, which in this case features can affect the final results. Some recent similarity functions that perform on both the feature and the vector spaces missed some properties of features, which leads to low accuracy. \\
Another important factor in learning procedures is the type of data objects. Normal objects can be categorized in a good way by almost all of the methods. Normal objects are those that follow the pattern of the data. But the accuracy of learning methods are influenced by some special objects, which the methods treat them as like as normal objects. For example, outliers which do not follow the pattern of the data should not be treated as normal objects, otherwise the accuracy of the method will not be acceptable. Type of data was not precisely considered by learning methods and most of the learning approaches consider all objects as normal objects or outliers. Due to the huge amount of data in recent years, redoing any learning algorithm is very costly, and using the sophisticated methods to cut the extra cost is mandatory. The method should also cover mutation analysis, diversity analysis (feature and vector spaces' analysis), and prediction and prevention strategies without redoing the learning procedures for many times.\\
In conclusion and in order to obtain the accurate result from learning methodologies, we need to make use of a comprehensive method that includes the accurate membership and similarity functions in addition to treat data objects based on their behaviour \cite{nine}. The selected membership function should consider all partial and full membership assignments for data objects with respect to all clusters, besides providing the facility to analyse objects' movements. The selected similarity functions should perform in both the feature and the vector spaces to handle the impact of features on final results. Lack of these abilities in learning procedures cannot be accepted by the recent and sophisticated methods that deal with big data and perform in high dimensional search spaces, where redoing any method can be costly and sometime impossible. \\
In brief, the issues with the methods in their learning strategies are the main reasons of the method proposed by this Thesis. Some of the main important parameters that influence the accuracy of learning methods (such as membership assignments, similarity functions, and data types) motivated the author of the Thesis to analyse the methods with respect to their functionalities in covering diversity, assigning memberships, measuring similarity, treating data types, handling mutation, proposing prediction and prevention strategies, and dealing with overlap conditions. In the final step, the Thesis aims to introduce a comprehensive method to cover the issues with other methods in their learning procedures in membership assignments, similarity measures, and treating different data types. The new proposed method is not an alternative or modifications of any learning methods, but instead it is a new method as a superset of other learning methods discussed in this Thesis.
%
%
\section{Goal of the Thesis}
\label{Goal of the Thesis}
\vspace*{-3mm}
The Thesis analyses the causes of misclassification and miss-assignments in learning methods in different stages of learning procedures such as membership assignments, similarity measurements, and treating different data types. The analysis resulted in observing some phenomena on misclassification, which consequently led to introducing a new type of object "critical" and a new type of feature "dominant" as some of the causes that were neglected by other methods.\\[0.12cm]
The goal of the Thesis is to propose a new method to improve the accuracy of partitioning process, by precise processing of  critical objects and dominant features. The new method is called Bounded Fuzzy Possibilistic Method (BFPM).\\[0.12cm]
The method is named Bounded for different reasons; the method puts some boundaries (constraints) on possibilistic methods to provide the most flexible search space, the method also bounds (ties) the fuzzy and possibilistic methods, and allows the objects to bound (jump) to other clusters (or classes). The proposed method aims to overcome the drawbacks of other methods that do not sufficiently take into account the influences of critical objects and dominant features on accuracy, where accuracy is mostly measured by calculating the percentage of the correct labelled objects \cite{four}. However the accuracy in clustering problems refers to the evaluation of the distance between the objects and the center of the clusters which is known as measuring separation and compactness of objects with respect to clusters' centers \cite{five}.\\
Fuzzy c-means algorithm as a well-known representative centroid-based clustering method is chosen for the analysis and verifications of the new method proposed by this Thesis. BFPM introduces a new membership function. The proposed membership function removes the constraints and limitations to provide a suitable environment as a search space for data objects. The proposed search space allows objects to show their potential abilities to participate in as much partitions as they can, which consequently results in obtaining better accuracy, in addition to facilitate the objects' movement analysis (mutation). The objects' movement analysis is very important in crucial systems such as medicine, security, risk management, and decision making systems, where any negligence leads to irreparable consequences. The new sets of similarity functions are proposed in this Thesis with the aim of covering diversity by addressing dominant features. The  similarity functions, called Weighted Feature Distance (WFD) and Prioritized Feature Distance Functions (PWFD), are introduced in different norms to detect such features and also to handle the impact of dominant features on final results, by performing in both the feature and the vector spaces.\\
In conclusion, the Thesis aims to prove the main hypothesis about the learning methodologies, whether there are other parameters rather than defined data types so far, similarity functions, and membership assignments that affect the results of learning approaches or not. To check whether the hypothesis is right, the Thesis evaluates if:
\vspace*{-2mm}
\begin{itemize}
\item  the membership assignments proposed up to date are adequate, or a new membership assignment should be introduced;
\vspace*{-2mm}
\item  objects can be just categorized into normal and outlier categories, or there are other types of objects that affect the results;
\vspace*{-2mm}
\item the accurate results can be obtained by similarity functions that perform in the vector space, or similarity functions should simultaneously consider both the vector and the feature spaces.
\end{itemize}
\vspace*{-2mm}
The new method will be verified on some real problems, and also on some benchmark datasets from different domains used in many experiments in high quality journal papers. According to the fundamental concept and implementation of the proposed method, some remarks should be noted:
\begin{itemize}
\vspace*{-2mm}
\item the method proposed by this Thesis is implemented by the author of the Thesis using Java,
\vspace*{-2mm}
\item datasets are chosen from Harvard medical school and an international bank, besides some datasets from UCI repository of the University of California as available benchmarks.\\
\end{itemize}
\newpage
\section{Thesis Organization}
\label{Thesis Organization}
\vspace*{-4mm}
The Thesis is organized in the following chapters. Chapter \ref{State-Chap} studies the recent and the most well known approaches in clustering and classification concepts. The chapter discusses about the centroid-based clustering methods by providing examples to illustrate the conditions that make the method vulnerable in their learning procedures. Some important parameters that affect the result of learning approaches such as similarity functions, different data types, and membership functions have been fully studied in this chapter. Similarity metrics, data types' taxonomies, and the most well-known membership assignments with the issues with their functionalities are discussed in this chapter. Functionality of Fuzzy C-Means (FCM) and other centroid-based clustering methods are explored in this chapter. The chapter also discusses the possibilistic methods by exploring the issues with the methods in centroid-based clustering approaches.  \\
Chapter \ref{Outstanding-Chap} introduces a new type of data object called "critical",  which was not sufficiently considered by other methods. This chapter illustrates some related examples from different disciplines in mathematics, geometry, medicine, security, economy, society, and education to explore the necessity of considering critical objects. The chapter aims to consider a new perspective on how learning methods should treat objects with respect to their types. \\
Chapter \ref{WFD-Chap} introduces a new type of features called "dominant" by discussing  different issues with similarity functions. The reasons of proposing similarity functions to cover diversity in both the feature and the vector spaces are studied in this chapter.  The chapter introduces new sets of similarity functions in different norms called "Weighted Feature Distance" (WFD) and "Prioritized Weighted Feature Distance" (PWFD) to overcome the drawbacks of similarity functions in handling the impact of features, specially dominant features. \\
Chapter \ref{BFPM-Chap} proposes a new method called "Bounded Fuzzy Possibilistic Method" (BFPM).
The functionality of BFPM is mathematically examined in this chapter. The issues with other methods have been studied in this chapter. The new membership function presented in this chapter overcomes the issues with other methods. The chapter also presents some new algorithms for supervised and unsupervised learning using different similarity functions and membership assignments in comparison with the proposed functions in this Thesis. \\
Chapter \ref{Measure-Chap} studies different accuracy and error measurements. Validity indices and some measurements on evaluating the accuracy, which are used in this Thesis, are explored in this chapter.  \\
Chapter \ref{Experiment-Chap} fully presents experimental verifications. The chapter provides different experiments on supervised and unsupervised learning strategies. Experiments on similarity and membership functions are also presented in this chapter. Critical objects and areas as well as the ability of learning methods to treat critical objects are discussed in this chapter. The chapter compares the functionality of different fuzzy and possibilistic methods with BFPM in terms of accuracy, fuzzification constant, dealing with overlapping, covering uncertainty, and mutation analysis. The obtained results from different validity functions on proposed methods have been depicted in this chapter.\\ 
Chapter \ref{Analysis-Chap} analyses the results of the proposed methods and functions. The chapter also evaluates the results of the proposed algorithms in different aspects: dealing with critical objects and ability of providing the flexible environment for objects to participate in more partitions. \\
Chapter \ref{Conclusion-Chap} presents the final conclusions on the proposed methods and functions. The chapter demonstrates the achievements of the proposed method and the introduced similarity and membership functions in comparison with other methods.\\
The proposed method has been already published in the Fuzzy Sets and Systems Journal \cite{nine} as well as other ideas in different papers cited at the end of the Thesis \cite{hundred-six}, \cite{hundred-ten, hundred-eleven, hundred-twelve, hundred-thirteen, hundred-fourteen, hundred-fifteen, hundred-sixteen, hundred-seventeen}, \cite{hundred-thirty-six}.\\ 
%
\chapter{Clustering and Classification Methods}
\label{State-Chap}
\section{Clustering Methods}
\label{Compared Clustering Methods}
Clustering is a form of unsupervised learnings to separate data objects into different groups or clusters based on the similarity between objects in datasets. Clustering is also called data segmentation in some applications, specially in image processing, with regard to its functionalities to partition a large dataset into different groups. The most important properties of clustering are: \textit{scalability}, \textit{ability to deal with different types of features (attributes)}, \textit{discovery of clusters with arbitrary shape}, \textit{minimal requirements for domain knowledge to determine input parameters}, \textit{ability to deal with noisy data}, \textit{insensitivity to the order of input records}, \textit{working in high dimensional search space}, \textit{constraint-based}, and \textit{interpretability and usability} \cite{eight}. In clustering problems, there is no label class or learning steps to guide the algorithm in advance. The algorithm must attempt to learn based on the similarity between the objects in a cluster and dissimilarities from the objects of the other clusters.\\
There are two main types of clustering approaches: soft and hard (crisp) clustering algorithms, which soft clustering can be categorized in different types. It is difficult to categorize the objects in crisp clusters as clusters may overlap \cite{eight}. There are some issues with crisp methods on membership assignments in dealing with uncertain conditions. Soft clustering methods are developed to deal with uncertainty and to overcome the issues with crisp methods. Crisp, fuzzy, probability, and possibilistic are some of the most common approaches that learning methods make use of them in their membership assignments. Each one of these methods provides different search spaces for data objects while the flexibility of the provided search spaces is different from one method to another. Clustering methods can be applied on different types of data such as \textit{intervals}, \textit{binary}, \textit{categorical}, \textit{ordinal}, \textit{ratio}, \textit{mixed-types}, and \textit{vectors}, where all of these types can be presented by numerical forms \cite{eight}. Assume a set of $n$ objects represented by $ O = \lbrace O_1, O_2, \;  ... \; , O_n \rbrace$ in which each object is typically introduced by numerical $feature-vector$ data that has the form $X \;= \; \lbrace x_1,... \;, x_d \rbrace \; \subset R^d $, where $d$ is the dimension of the search space or the number of features. A $c \times n$ partition (membership) matrix is often represented as a matrix $U = [u_{ij}]_{c\times n}$, and a cluster can be represented by a vector from the matrix U, where $u$ represents a membership value, $j$ is the index to present the $j^{th}$ object in the dataset, $i$ is the index to present the $i^{th}$ cluster, and $c$ is the number of clusters. Crisp clusters are non-empty, mutually-disjoint subsets of $O$, Eq. (\ref{Crisp-M}).\\
\begin{equation}
\nonumber
	\label{Crisp-M}
M_{hcn} = \bigg\lbrace U \; \in \Re^{c\times n}| \; u_{ij} \; \in \lbrace 0,1 \rbrace, \; \;  \forall i,j;
\end{equation}
\begin{equation}
 0 < \sum_{j=1}^n u_{ij} < n,  \; \;  \forall i; \; \;  \sum_{i=1}^c u_{ij} =1, \; \;  \forall j  \; \; \bigg\rbrace
\end{equation}
$u_{ij}$ is the membership of the object $O_j$ in cluster $i$. If the object $O_j$ is a member of cluster $i$, then $u_{ij} \;  = \; 1;$ otherwise, $u_{ij} \;  = \; 0$. Fuzzy or probability membership assignments allow each object to have a partial membership in more than one cluster \cite{ten}, \cite{eleven}. The idea was to come up with some solutions that crisp methods have difficulties to handle uncertainty and partial membership assignments. This condition is stated in Eq. (\ref{fuzzy-Assign}), where objects may obtain a partial nonzero membership in several clusters, but only a full membership in one cluster \cite{twelve}.
\begin{equation}
\nonumber
\label{fuzzy-Assign}
M_{fcn} = \bigg\lbrace U \; \in \Re^{c\times n}| \; u_{ij} \; \in [ 0,1 ], \; \; \forall i,j;
\end{equation}
\begin{equation}
0 < \sum_{j=1}^n u_{ij} < n, \;\; \forall i; \;\;  \sum_{i=1}^c u_{ij} =1, \;\; \forall j \; \; \bigg\rbrace
\end{equation} \\
In other words and based on Eq. (\ref{fuzzy-Assign}), each column of the partition matrix must sum to 1, $(\sum_{i=1}^c u_{ij} =1)$. Some fuzzy-decision tree methods and fuzzy hierarchical methods have been introduced to provide more relaxed search spaces for data objects in comparison with crisp methods. Possibilistic method has been introduced to relax the condition in fuzzy methods by providing a more flexible search space for data objects, Eq. (\ref{Possibilistic-M}) \cite{thirteen}. \\
\begin{equation}
\nonumber
\label{Possibilistic-M}
M_{pcn} = \bigg\lbrace U \; \in \Re^{c\times n}| \; u_{ij} \;\; \in [ 0,1 ], \;\;  \forall i,j;
\end{equation}
\begin{equation}
0 < \sum_{j=1}^n u_{ij} \leq n, \; \forall i; \;\;  \max_{1\leq i\leq c}\; u_{ij} >0, \;\; \forall j \; \; \bigg\rbrace
\end{equation}
In Eq. (3), the condition $(\sum_{i=1}^c u_{ij} =1)$ is relaxed by substituting it with $(\underset{1\leq i\leq c}{max}\; u_{ij} >0)$. Based on Eq. (\ref{Crisp-M}), Eq. (\ref{fuzzy-Assign}), and Eq. (\ref{Possibilistic-M}), it is easy to see
that all crisp partitions are subsets of fuzzy partitions, and a fuzzy partition is a subset of a possibilistic partition:\\
\begin{equation}
 M_{hcn} \subset M_{fcn} \subset M_{pcn}\\
\end{equation}
The major clustering methods (either hard or soft methods) can be categorized into the following categories: \textit{partitioning}, \textit{hierarchical}, \textit{density-based}, \textit{grid-based}, and \textit{model-based} methods.
%
\begin{itemize}
\item {\bf Partitioning methods} \\
These methods assign \textit{n} data objects into \textit{k} (or $c$) clusters, where $(k<n)$. Partitioning methods work on the similarity between objects, while the procedure of measuring the similarity is different from one approach to another. The most well-known partitioning methods are k-means, k-medoids, and fuzzy c-means \cite{fourteen}. These methods work based on the centroid-based technique, that use a centroid (prototype) as a cluster-core to cluster data, where in each iteration the centroid is being updated by considering the objects of each cluster. Fig. \ref{Centroid_Cluster} shows a general idea of centroid-based technique used in partitioning method.
\begin{figure}[!ht]
\begin{center}
\leavevmode\fbox{\parbox[b][7cm][s]{140mm}{
\vfill\footnotesize {\includegraphics[width=14cm,height=7cm]{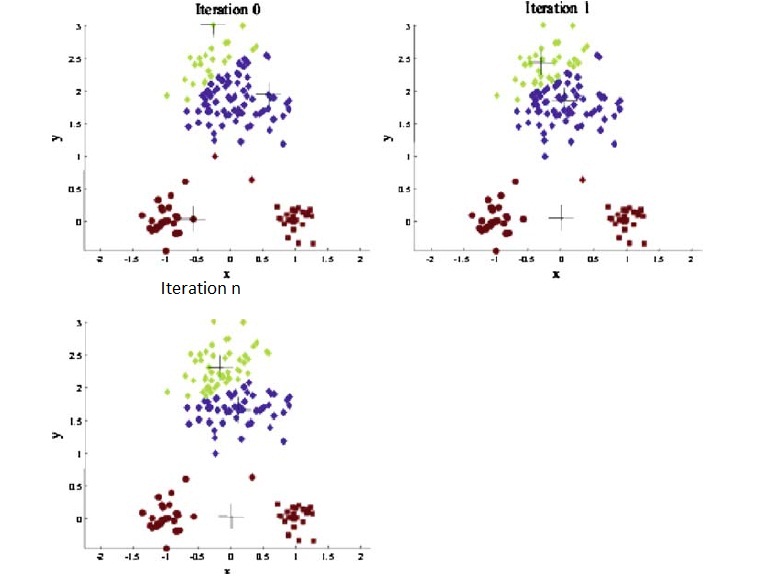}}\vfill}}
\caption{General idea of centroid-based technique used in partitioning method. In each iteration the centroids are being updated \cite{fifteen}.}
\label{Centroid_Cluster}
\end{center} 
\end{figure}
\item {\bf Hierarchical methods} \\
These kinds of methods create the hierarchical structures of the given datasets. The method can be categorized either in agglomerative (bottom-up approach) or divisive (top-down) form. The former approach starts with an object to form a cluster. The method merges the clusters or objects that are close to each other. The procedure continues until all the objects are categorized into one topmost cluster or the algorithm reaches the termination condition. The divisive approach starts with all objects as one cluster. The method splits the objects into smaller clusters until each object is assigned to one cluster or the termination condition is reached. The methods work in opposite directions, agglomerative methods start from the leaves to the root and the divisive methods work from the root to the leaves. Redoing the hierarchical methods is impossible with respect to the processing steps (merge or split). These types of methods are mostly used for crisp strategies.
\item {\bf Density-based methods}\\
The methods work based on the notation of objects density. The cluster grows until the objects in neighbourhood meet the threshold. Such a model can be used as a filter model to remove the noise (outlier) from clusters. DBSCAN (Density-Based Spatial Clustering of Applications with Noise) and DENCLUE (DENsity-based CLUstEring) are the well known density methods. The former one works based on density-connectivity and density-reach-ability, while the latter one performs based on density distribution functions. DBSCAN searches on the search space for the object $p$ in $\varepsilon-neighbourhood$. The values for $\varepsilon$, as a distance between $p$ and other objects, are not fixed for all problems. If the number of objects in neighbours is greater than the threshold, then the algorithm creates a new cluster for $p$ as a core. The parameter's initialization needs to be addressed by the user, which makes the algorithm complicated in high dimensional search spaces. OPTICS (Ordering Points to Identify the Clustering Structure) is introduced to solve the difficulties of implementation of DBSCAN method.
%
\item {\bf Grid-based methods}\\
Grid methods use the grid structure to categorize the objects to a finite number of clusters or cells. The fast processing time is one of the advantages of this kind of clustering method, which works based on the number of clusters regardless of the number of objects. STING (STatistical INformation Grid) is a grid-based method that divides the search space into some rectangular cells. The method followed the idea from hierarchical strategy to cluster objects. Each cell can be partitioned into smaller rectangular cells for the next layer. The method has many advantages such as: efficiency, parallel processing, and query-independent. Once the algorithm runs on the dataset, it computes all the statistical parameters regarding each cell and it does not need to calculate them for several times. Grid-based structure allows the method to process objects with respect to the cells in parallel by storing all the statistical information in each cell separately. The method can respond to the queries with respect to each cell very fast.
\item {\bf Model-based methods} \\
These methods design the models with respect to spatial distributions of objects for clusters and find the best fit of objects for each cluster. Conceptual and Expectation-Maximization are some of the model-based methods. In general, each cluster can be represented by a parametric probability distribution \cite{sixteen}, and the question is which probability distribution can be used to represent the clusters, where the parameter estimation is an essential step in such a model. Fig. \ref{EM} illustrates the objects clustered by the model-based method using Normal (Gaussian) distribution. Conceptual clustering methods make use of probabilistic descriptions to cluster objects into two steps: in the first step the objects are clustered based on the clustering methods and in the second step characterization is performed to find the characteristic description for each group. COBWEB is one of the popular conceptual-based clustering methods.  

\begin{figure}[!ht]
\begin{center}
\leavevmode\fbox{\parbox[b][5cm][s]{110mm}{
\vfill\footnotesize {\includegraphics[width=11cm,height=5cm]{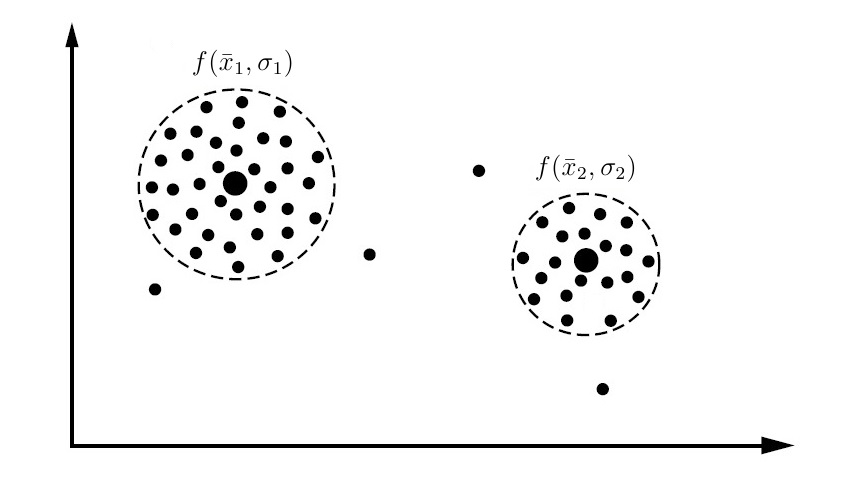}}\vfill}}
\caption{Normal distribution with $ f(\bar{x}_1,\sigma_1)$ and $f(\bar{x}_2,\sigma_2)$ (mean and standard deviation) used by EM (Expectation-Maximization) model-based clustering method.} 
\label{EM}
\end{center}
\end{figure}
\end{itemize}
%
%
\subsection{Prototype-Based or Centroid-Based Methods}
\begin{figure}[!ht]
\begin{center}
\leavevmode\fbox{\parbox[b][4.6cm][s]{120mm}{
\vfill\footnotesize {\includegraphics[width=12cm, height = 4.6cm]{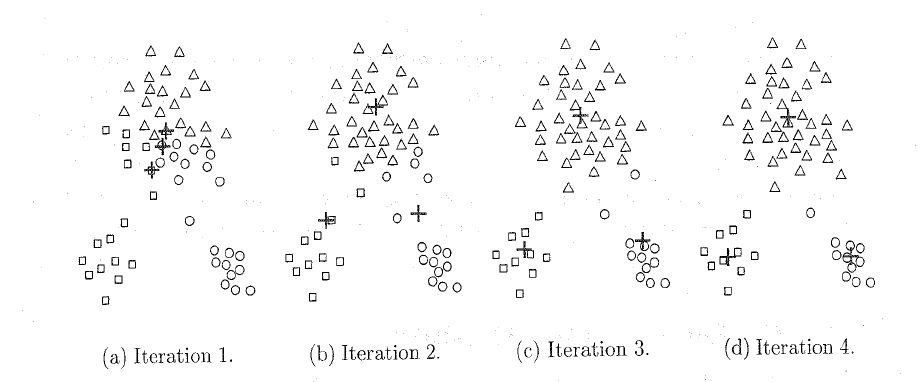}}\vfill}}
\caption{Selecting centroids in an appropriate condition.}
\label{Centroid-1}
\end{center}
\end{figure}

\begin{figure}[!ht]
\begin{center}
\leavevmode\fbox{\parbox[b][4.6cm][s]{120mm}{
\vfill\footnotesize {\includegraphics[width=12cm, height = 4.6cm]{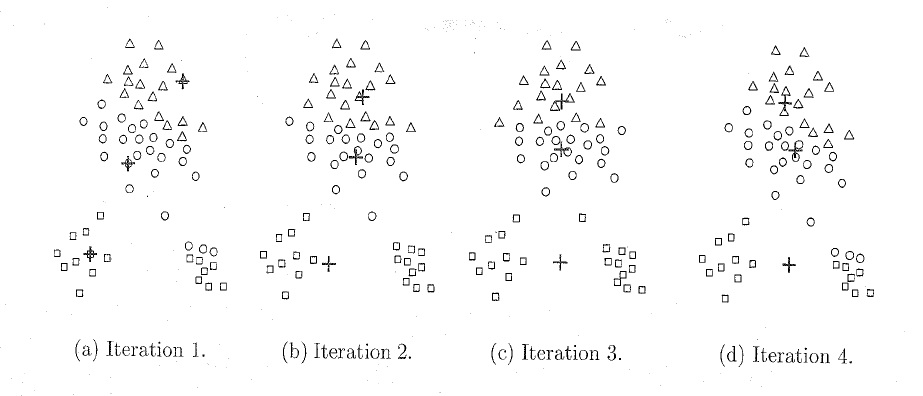}}\vfill}}

\caption{Selecting centroids in an inappropriate condition.}
\label{Centroid-2}
\end{center}
\end{figure}

Centroid-based clustering is an essential learning approach that is also used in data mining, pattern recognition, and statistical analysis. Centroid-based methods partition the data into clusters according to the similarities between the objects and the centroids, which help the procedure of extracting new information (or knowledge discovery) for new patterns. In centroid-based clustering (fuzzy c-means, k-means) each centroid is updated in each iteration. But, updating the centroids in each iteration in some cases causes miss-assignments. If the centroids are wrongly selected at the beginning of the process or the procedure goes wrong, the error will be significantly increased in the next iterations that affects the procedure of choosing new centroids for the next iterations. The proper and wrong selection of centroids are depicted by Fig. \ref{Centroid-1} and Fig. \ref{Centroid-2} respectively. The centroids that seem to be well determined by the algorithm with respect to one feature space might be inappropriate from other feature perspectives. This is one of the reasons of miss-assignments and miss-clustering \cite{seventeen}. To deal with the miss-assignments, some methods apply different methodologies to pick the most proper centroids in initializations' steps instead of selecting the centroids randomly. 
%
%
%
%
\subsubsection{K-means Algorithm}
\begin{figure}[!ht]
\begin{center}
\leavevmode\fbox{\parbox[b][5cm][s]{120mm}{
\vfill\footnotesize {\includegraphics[width=12cm,height=5cm]{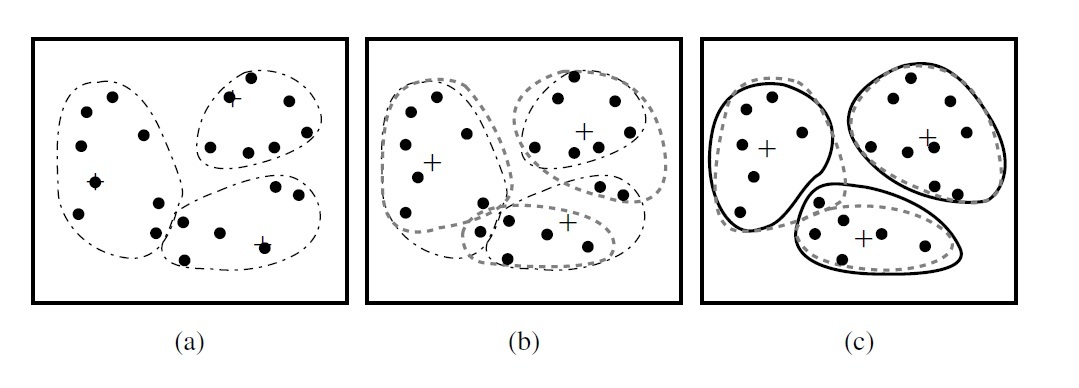}}\vfill}}
\caption{General idea of K-means algorithm to update centroids and objects of each cluster. The steps are from (a) to (c), and in each iteration the centroid is updated based on the mean value of objects in each cluster \cite{eight}.}
\label{Centroid-updates}
\end{center}
\end{figure}

The method clusters \textit{$n$} objects into \textit{$k$} clusters. The K-means algorithm evaluates the objects based on the mean value of each cluster known as centroid (or centra, or prototype), on the way that the intra-cluster similarity is high and the inter-cluster similarity is low. The algorithm works based on the following steps. First, it randomly selects $k$ objects as centroids. Then the objects will be measured based on the similarity between the centroids and the objects. Fig. \ref{Centroid-updates} shows the general idea of how the centroid is being updated in each iteration.
The selected similarity function compares data objects for being grouped into the clusters based on their distances. The closest objects to the centroid are clustered with the centroid. The algorithm calculates the square-error in each iteration, shown by Eq. (\ref{K-mean-Similarity}).
\begin{equation}
\label{K-mean-Similarity}
Err =\sum_{i=1}^k  \sum_{j=1}^{n} | O_j - C_i|^2 
\end{equation}
where \textit{$Err$} is the sum of the square-error of the \textit{$j^{th}$} object in the \textit{$i^{th}$} cluster from the \textit{$c_i$} centroid. The algorithm runs until the centroids do not change \cite{two}. The method updates the centroids in each iteration based on the mean value of the objects in each cluster. \\
%
%
%
%
\subsubsection{FCM (Fuzzy C-Means) Method}
There are two important types of FCM Algorithm: one is based on the fuzzy partition of the sample set and another is on the geometric structure of sample set of kernel base method \cite{eighteen}, which in this Thesis fuzzy partitioning is mostly considered. The main FCM function can be defined as follow, Eq. (\ref{General-FCM}).\\
\begin{equation}
\label{General-FCM}
J_m(U,V) \; = \; \sum_{i=1}^c \sum_{j=1}^n \; u_{ij}^m \;|| O_j - V_i||_{A} ^2 \; ;
\end{equation}
%
where U is the $(c \times n)$ partition matrix, $ V= \lbrace v_1, v_2,..., v_c \rbrace $ is the vector of $c$ cluster centers in $ \Re^d , \; m  > 1 $ is the fuzzification constant, and $ ||.||_A $ is any inner product A-induced norm, where the inner product of two vectors $v_1$ and $v_2$ in $d$ dimensions is the standard vector dot product defined as $\sum_{i=1}^d v_{1i} \; v_{2i}$ and the A-induced norm can be any norm, which for instance, the second norm $||v_1||_2$ is defined as $||v_1||_2 = \sqrt{v_1 \; v_1}$ \cite{eight}.
\subsection{Kernel-Based Methods}

In the kernel FCM, the dot product $ \Big{(} \phi(O).\phi(O) = k(O,O) \Big{)}$ is used to transform feature-vector $O$, for non-linear mapping function $\Big{(}\phi : \; O \longrightarrow \phi(O) \; \; \in  \; \; \Re^{d_k} \Big{)}$, where $d_k$ is the dimensionality of the feature space. Eq. (\ref{kFCM}) presents a non-linear mapping function for Gaussian kernel \cite{nineteen}.\\

\begin{equation}
\label{kFCM}
J_m(U;k) = \sum_{i=1}^c \Big{(} \sum_{j=1}^n \sum_{k=1}^n (u_{ij}^m u_{ik}^m \; \; D_k(O_j,O_k))/2 \sum_{l=1}^n u_{il}^m \Big{)} 
\end{equation}\\

\begin{figure}[!ht]
\begin{center}
\leavevmode\fbox{\parbox[b][8cm][s]{130mm}{
\vfill\footnotesize {\includegraphics[width=13cm, height=8cm]{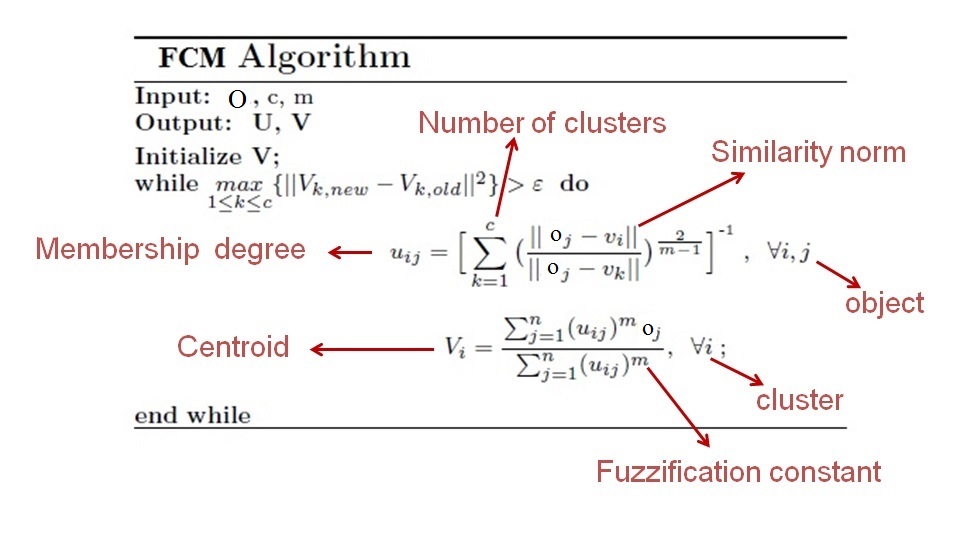}}\vfill}}
\caption{The general concepts of fuzzy c-means (FCM) algorithm \cite{one}.}
\label{centroid}
\end{center}
\end{figure}

\hspace*{-7.5mm} where $U \in M_{fcn}, \;\; m > 1$ is the fuzzification parameter, and  $D_k(O_j,O_k)$ is the kernel base distance \cite{twenty} (replaces Euclidean distance function) between the $j^{th}$ and the $k^{th}$ feature-vectors as, Eq. (\ref{kernel distance}).\\

\begin{equation}
\label{kernel distance}
D_k(O_j,O_k) = k(O_j,O_j) + k(O_k,O_k) - 2k(O_j,O_k)
\end{equation}
Now, let us discuss about the parameters that affect the accuracy of methods by evaluating a common algorithm in clustering problems. As Fig. \ref{centroid} shows, the procedure of partitioning objects $(O)$, based on the centroid-based clustering methods, depends on some parameters such as membership assignments, similarity functions, and the type of objects. In each iteration, the membership values $(U, u_{ij})$ for each object with respect to each cluster $(C_i)$ will be updated based on similarity functions that have been used \cite{twenty-one}. This is clear that we need to choose the most accurate membership and similarity functions to obtain the precise results. On the other hand, we see that the type of object is the main key in each algorithm as the values of memberships will be calculated for data objects.
In conclusion, membership assignments, similarity functions, and data types are some of the most important parameters that affect the accuracy of clustering methods. This Thesis proposes new methodologies in membership assignments and similarity functions to obtain the better results in addition to introduce a new type of data object as one of the causes that leads the methods to improper accuracy. Before moving to propose the new methodologies, let us review some related works, data types, similarity functions, and some classification methods that will be used in experimental verification. The proposed methodologies, membership assignments, similarity functions, a new type of feature, and a new type of object are considered and applied by this Thesis in both supervised and unsupervised learning strategies. 
\section{Classification Methods}
Classification is a form of supervised learning that performs in a two-step process. Let us recall the set of $n$ objects represented by $ O = \{ O_1, O_2, \;  ... \; , O_n \}$, which each data object is typically described by numerical $feature-vector$ data that has the form $X \;= \; \{ x_1,... \;, x_d\} \; \subset R^d $ in $d$ dimensional search space. It should be noted that classification methods also deal with different types of objects and attributes. In classification, the dataset is mostly divided into two parts: a learning set $ O_L = \{ O_1, O_2, \;  ... \; , O_l \}$ and a testing set $ O_T = \{ O_{l+1}, O_{l+2}, \;  ... \; , O_n \}$. In binary classification as like as crisp methods, each object can be a member of just one class, where data objects are classified based on a class label $x_c$. A class is a set of objects with values $\{ u_{ij}\}$, where $u$ represents a membership value, $j$ is the the index to present the $j^{th}$ object in the dataset and $i$ is the index to present the $i^{th}$ class (the indexes are chosen in the similar way of clustering definition). Similarity and membership assignments in supervised and unsupervised problems are almost similar as each data object needs to be compared and assigned a partial or full membership from each class. Membership functions consider similarity functions to evaluate data objects. \\

%
%
\begin{figure}[!ht]
\begin{center}
\leavevmode\fbox{\parbox[b][6.5cm][s]{130mm}{
\vfill\footnotesize {\includegraphics[width=13cm,height=6.5cm]{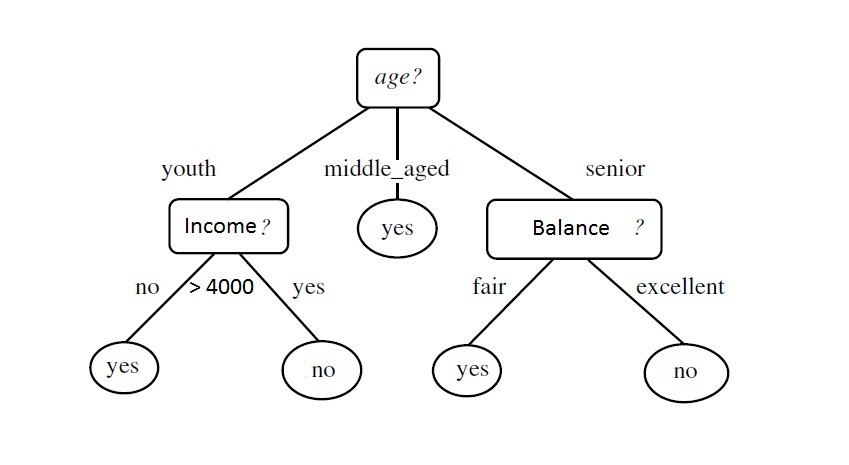}}\vfill}}
\end{center}
\caption{A decision tree for the loan assignment to bank customers. The tree checks whether the customer is risky or not.}
\label{DTree_Loan}
\end{figure}
\hspace*{-7.5mm} Decision tree and neural networks have been commonly used in classification problems. Decision tree algorithm is a flowchart-like tree that each internal node denotes an attribute (feature) \cite{twenty-two}. Fig. \ref{DTree_Loan} shows how decision tree decides to label any object. Each internal node is designed to test the particular attribute. Each branch represents an outcome of the test on the upper node and each leaf node (or terminal node) holds a class label. The top most node is known as a root node. The decision tree presented in the figure is made for the loan application from an international bank to evaluate whether the giving loan to a particular customer is risky or not. There are some measures and techniques such as \textit{Information Gain},
 \textit{Gain Ratio}, and \textit{Gini Index} to select the root and attributes of the tree to obtain the best partition for data objects. The basic method of decision tree is used in several classification algorithms, such as ID3, C4.5, and CART. The main disadvantage of this type of classification method is that the method cannot deal with very large datasets easily. On the other side, the algorithms occupy the huge amount of memory, so this drawback makes the algorithms uncomfortable for large datasets \cite{eight}.\\ Artificial Neural Network (ANN) is another concept that has been used in supervised learning strategies. ANN is a set of connected nodes to each other (input/output units). The link between two nodes contains a weight, which helps the network to learn by adjusting the weight with respect to the previous data object \cite{twenty-three}. Back-Propagation and Feed-Forward neural networks have commonly used in different disciplines. ANN contains several layers: input, hidden, and output layers. The hidden layer mostly works as a black box and may contain more than one layer. The weights on links between nodes are being updated by each learning step. In training phase of back-propagation neural network, the algorithm compares the values of the output layer with the expected values to calculate the error occurred by the assigned weights. This error will affect the weights of the previous nodes till the values of output nodes get close and closer to the desirable output values. This procedure continues for all objects in the training dataset. After this step, the network stabilizes the weights assigned to each link and the neurons. The network is ready to work for data objects of the testing dataset. Time and cost are two main issues on these types of learning methods.\\
But in feed-forward neural network, the weights and neurons are being affected by the objects from the training dataset. These assignments vary based on the type of classification problems. Sometimes a range of output values can be considered to classify the objects on the test phase, but some methods update the weights between nodes to make the network ready to classify the objects on the test phase \cite{twenty-four}. As like as the discussion for supervised methods, membership assignments, similarity functions, and the type of objects are still the center of attention in learning procedures. Obtaining the proper results can be done by providing the most accurate membership assignments, selecting the proper similarity functions, and knowing the exact type of objects in advance.  
%
%
\section{Uncertainty in Clustering and Classification Methods}
As mentioned, in some cases objects cannot be certainly categorized in one category. To deal with this uncertainty several methods such as fuzzy, rough, probability, and possibilistic approaches have been utilized. Fuzzy or probability methods have become popular and have been also compared with other approaches. The most important factor in each method is how to assign memberships to objects with respect to each class, path, problem, solution, or cluster. According to different methodologies, dealing with uncertainty is a crucial concept that needs to be addressed by  methods, rather than using crisp methods. Most of the well known categories that make use of uncertainty can be considered as follow:
\begin{itemize}
\item inaccurate measurements,
\item random occurrences,
\item vague descriptions.
\end{itemize}
These categories describe the uncertainty in deterministic, probabilistic, and fuzzy models respectively \cite{twenty-five}, \cite{twenty-six}. The process is deterministic whenever the outcome of a process can be completely described by the absolute certainty. In other situation, if the outcome of a process is random then the process is probabilistic, and if the element of uncertainty is neither caused by measurements error nor by random occurrence then we are dealing with a source of fuzziness. The uncertainty in fuzzy systems arises into two types:
\begin{itemize}
\item different answers to the same question for same antecedents but different consequents,
\item parameters in membership functions are interpreted differently by different people.
\end{itemize}
Fuzzy algorithms in supervised methods have been widely discussed in several papers, apart from statistical classification methods. Allowing each object to participate in more than one category is the main key of the approaches that deal with uncertainty. The fuzzy method is also applied in combination with other methods in order to obtain better results. For example, Fuzzy SVM approach on credit-risk is discussed in \cite{twenty-seven}. Uncertainty and overlapping are the most important parameters that influence the outcomes of the methods. In certain conditions where there is no overlapped objects, most of the methods can partition objects in a desirable way, but accuracy of the methods declines when the methods face overlapped objects. The accuracy of learning methods can be evaluated based on the ability of the method in uncertain conditions. In overlapped clusters, the accuracy of the method is acceptable when overlapped objects are being assigned to proper clusters.\\
Crisp methods allow objects to participate in just one cluster, which is suitable for most approaches that need to differentiate and assign objects just in one cluster, $(u_{ij}=0)$ or $(u_{ij} =1)$, where $u$ is the membership function to assign memberships to the $j^{th}$ object for the $i^{th}$ cluster. These types of approaches have difficulties to provide a flexible search space to deal with mutation (object's movement). Fuzzy or probabilistic methods are as same as crisp methods with more flexibility in the way that each object can obtain memberships from more than one set. The range of memberships, which can be obtained by each object, is between zero and one ($ 0 \leq u_{kj} \leq 1 $). If an object $O_j$ obtains $ u_{kj} =1$ from a particular cluster $k$, then $O_j$ is fully a member of $k$. If  $u_{kj} =0 $, then $O_j$ is not a member of the $k^{th}$ cluster at all, and if $( 0 < u_{kj} < 1 )$, then $O_j$ is partially a member of the $k^{th}$ cluster. According to above description, members are able to obtain memberships from several clusters, but they are not able to obtain  full membership of more than one cluster. In fact, if they obtain a full membership of one cluster, then they are not able to participate in other clusters. Several membership and similarity functions have been used in several methods to provide better performance in overlapped and uncertain conditions, where some of them are more sophisticated in comparison with others. In follow, some methodologies that have been proposed to deal with overlapping are explored. 
    \begin{itemize}
    \item {\bf Removing overlapped conditions.}\\
Some approaches have been introduced to keep all objects completely separated from each other. Applying such methodologies helps the objects to be partitioned in different categories easily. Support Vector Machine (SVM), Hierarchical methods, and decision trees are the most well known approaches from this category \cite{twenty-eight}. These approaches are useful for some datasets, but the accuracy of these methods is not always good as the characteristic of the overlapped object is changed or has not been fully considered. In other words, these methods have challenges to assign the proper memberships to overlapped objects, and they can be used in cases that objects are supposed to be completely separated. In recent years and due to the huge amount of data in social networks, big data, and high dimensional datasets, separating objects from each other or removing overlapped conditions not only is not easy but also is not necessary as objects should be analysed with respect to all clusters.   
\item {\bf Using membership assignments that able to cover uncertainty.}\\ 
As mentioned, removing overlap conditions or replacing overlapped objects with new objects did not provide the desirable accuracy. To obtain better accuracy in uncertain conditions, some fuzzy, probability, and possibilistic methods have been introduced \cite{twenty-nine}, \cite{thirty}. These methods aim to categorize the objects into the proper cluster by evaluating their partial memberships with respect to different clusters. In general, these methods allow objects to obtain partial memberships from more than one cluster. This strategy allows the methods to consider overlapping instead of separating all the objects without considering their participations in other clusters.
\item {\bf Using machine learning methods to deal with overlapping.}\\
    According to the huge amount of data in different disciplines and based on the variety of problems that need to be addressed by overlapped objects, different approaches from machine learning \cite{thirty-one}, statistical methods, and artificial intelligence methods such as particle swarm \cite{fifteen}, ant colony \cite{thirty-two}, genetic algorithms \cite{thirty-three}, Bayesian \cite{thirty-four}, regression \cite{thirty-five}, and evolutionary algorithms \cite{thirty-six} have been applied. These methods perform well for some specific datasets with respect to the type of partitioning problems, but the accuracy of the proposed methods are not acceptable for other datasets while performing on overlapped objects.    
\item {\bf Using combined methods.}\\  
According to above discussion, methods have been combined with each other to obtain better results. Most of the statistical methods, fuzzy, probability, and possibilistic methods have been joined with machine learning methods such as fuzzy SVM \cite{thirty-seven}, fuzzy decision tree \cite{thirty-eight}, fuzzy regression \cite{thirty-nine}, fuzzy neural network \cite{forty}, fuzzy possibilistic \cite{forty-one}, fuzzy hierarchical \cite{forty-two}, fuzzy genetic \cite{forty-three}, Bayesian Network-based \cite{forty-four}, and fuzzy type-II \cite{forty-five}. The methods have been also used in both centroid and kernel based strategies to obtain better results, where different centroid-based approaches (such as fuzzy c-means \cite{forty-six}, k-means algorithms \cite{forty-seven}, and centroid-based classification) and kernel-based approaches (such as Multiple Kernel Fuzzy, Fuzzy and Possibilistic Kernel c-means, Relaxed Exponential Kernels, and Gaussian kernel-based fuzzy c-means) have been introduced \cite{forty-eight}.  
    \end{itemize}
\vspace*{-2mm}    
As a result, similarity functions, membership assignments, and the type of data are considered as the most important factors in learning procedures regardless of the type of learning methods.
\vspace*{-2mm}
\section{Data Types}
\label{Data-FCM-Analysis}
As discussed above, learning methods need to come up with some solutions to deal with uncertainty and overlapping. The main causes of miss-classifications and miss-assignments in partitioning methods are overlapped conditions, where the methods are not able to assign overlapped objects to proper partitions. To deal with the participations of data objects in one or more clusters (overlapped), first we need to identify the properties of different types of data objects. For this purpose, several books and papers have been introduced. Data objects are considered as fundamental keys in the methods that without the objects the learning and mining algorithms are meaningless. Objects basically direct the accuracy of the selected algorithm in case if they are extracted from inappropriate groups. Knowing the exact type of object leads the investigators to providing a suitable environment for learning algorithms. Supervised and unsupervised learning methods propose some membership functions to categorize data objects and solutions with respect to behaviour of each data category. So far, data objects are categorized into two main categories, either outlier or normal objects. \\
Outliers are those that do not fit the model of the data (data pattern). Unlike outliers that do not fit the model of data, normal objects fit into one model of data in the datasets. The normal objects can be easily processed by the methods with respect to the data patterns. The procedure of outlier detection is very complicated as some data objects might be known as outliers from one dimension or one feature space, but these particular objects can be considered as normal objects from different dimensions. The importance of considering the type of data objects is being highlighted when the methods treat different types of objects to obtain better accuracy. For example, treating outliers like as normal objects massively skews the final results, and also paying less attention to objects that cause overlapping may influence the final results. In big data applications, knowing the exact type of objects and selecting the most accurate similarity and membership assignments are crucial to cut the extra costs, besides obtaining better performance. Redoing the learning algorithms on big data platform is not reasonable and sometimes is not possible. To prevent redoing learning algorithms, we need to study the type of data in advance. Providing the relaxed environment for data objects is the most desirable goal of learning algorithms. \\
Supervised and unsupervised methods have introduced sophisticated functions in their processing stages, and within each stage, data objects as the fundamental keys should be evaluated. This evaluation starts from object recognition till  assigning the proper memberships to each data object. As discussed, data objects are the basis of datasets, where learning methods and data mining techniques extract the knowledge from them. To obtain the most accurate results, we need to pay more attention to data type in learning and mining algorithms. Due to the growth of data in recent years in different disciplines, the importance of objects' recognitions in different types is inevitable. Sophisticated objects discussed in this section help to avoid the cost of redoing learning and mining techniques caused by mixing the objects up with each other.
\vspace*{-2mm}
\subsection{Data Type Taxonomies}
 Data mining and learning methods evaluate data objects based on their patterns (\textit{descriptive}, \textit{predictive}, and \textit{prescriptive}) \cite{eight}. According to learning procedures and mining functionalities, the type of data objects should be considered as each type of object has its own characteristics, behaviour, and different effects on the final results. For instance, outlier(s) can be considered as very interesting objects in anomaly detection \cite{forty-nine}. On the other hand, outliers do not play any role in other applications as they are considered as noise \cite{fifty}. Due to the fact that each type of data object has different effects on the final results, the Thesis aims to look at different types of data objects from different perspectives. Data objects can be named as: \textit{record}, \textit{point}, \textit{pattern}, \textit{event}, \textit{sample}, \textit{observation}, \textit{entity}, \textit{case}, or \textit{vector} \cite{two}. A dataset is a set of objects. There are different types of datasets with respect to the type of data objects. Each data object is a set or a collection of some components named as: \textit{attribute}, \textit{variable}, \textit{characteristic}, \textit{field}, \textit{identifier}, \textit{outcome}, \textit{feature}, or \textit{dimension}. Datasets can be categorised in different categories based on their dimensionality, sparsity, and resolution. The most well known dataset categories are: \textit{Relational}, \textit{Transactional}, \textit{World Wide Web}, \textit{Flat Files}, \textit{Data Streams}, and \textit{Data Warehouses} \cite{fifty-one}. Types of attributes are also categorized into some categories such as: \textit{Nominal}, \textit{Ordinal}, \textit{Interval}, and \textit{Ratio}. In general data objects (either outlier or normal object) are categorized into single variable or simultaneously two or more variables. 
%
%
\begin{itemize}
\item{\bf {Uni and Multivariate Data Objects}\\}
Let's start with the simplest definition for data objects and categorize them into single variable or simultaneously two or more variables \cite{fifty-two}.
\begin{itemize}
\vspace*{-1mm}
\item{\bf{Univariate Data Object:}\\}
Observations on a single variable on datasets $O= \lbrace O_1, O_2,..., O_n\rbrace$, where $n$ is the number of single variable observations $(O_j)$. Univariate data objects can be categorized into two groups: 
\begin{itemize}
\item categorical or qualitative \cite{fifty-three},
\item numerical or quantitative.
\end{itemize} 
\item{\bf {Multivariate Data Object:}\\}
Observations on a set of variables on $O= \lbrace O_1, O_2,..., O_n\rbrace$ dataset, where $n$ is the number of observations, $O_j= \lbrace x_{j1}, x_{j2},..., x_{jd}\rbrace$, and $d$ is the number of variables or dimensions. Each variable can be a member of the above mentioned categories. 
\end{itemize}
%
%
\vspace*{-2mm}
\item{ \bf {Complex Data Objects}}\\
The growth of data in various types prevents data taxonomies to categorize objects into mentioned categories. Most of the methods need to distinguish all sophisticated objects to introduce more efficient methodologies and algorithms in mining and aggregation techniques. Complex and advanced categories are two main types for sophisticated objects. These objects not only have sophisticated structures, they also need advanced techniques, storages, and algorithms for being saved, retrieved, and analysed. Complex objects are categorized in:
\begin{itemize}
\vspace*{-2mm}
\item{ {structured data object}},
\item{ {semi-structured data object}},
\item{ {unstructured data object}},
\item{ {spatial}},
\item{ {hypertext}},
\item{ {multimedia}}.
\end{itemize}
%
%
\vspace*{-2mm}
\item {\bf {Advanced Data Objects}\\}
Due to the growth of data in recent years, the cost of using legacy techniques to evaluate objects using some measurements (similarity or dissimilarity, ...) is very high. To reduce the amount of cost of running these measurements, some evaluation techniques such as similarity functions between histograms, graphs, networks, and so on are introduced. For example, for graph objects, not only the objects of each graph are not evaluated individually, but also the whole graph will be evaluated as an object. In sophisticated algorithms, each one of these structures is known as a new type of objects. In this category, objects are nested to each other and even the structured type is considered as a new type, that are categorized:
\vspace*{-2mm}
\begin{itemize}
\item { {sequential patterns}},
\item { {graph and sub-graph patterns}},
\item{ {objects in interconnected networks}},
\item{ {data stream or stream data}},
\item{ {time series}}.
\end{itemize}
\end{itemize}
%
%
\vspace*{-5mm}
\subsection{Outlier and Outlier Detection}
Objects from this category can be any object from above mentioned categorizes (complex, advanced, univariate, and multivariate data objects). Outliers are very crucial and important in datasets, because they have abilities to change the results of selected learning algorithms. 
%
\vspace*{-3mm}
\subsubsection{Outlier}
\begin{figure}[!ht]
\begin{center}
\leavevmode\fbox{\parbox[b][5cm][s]{120mm}{
\vfill\footnotesize {\includegraphics[width=12cm,height=5cm]{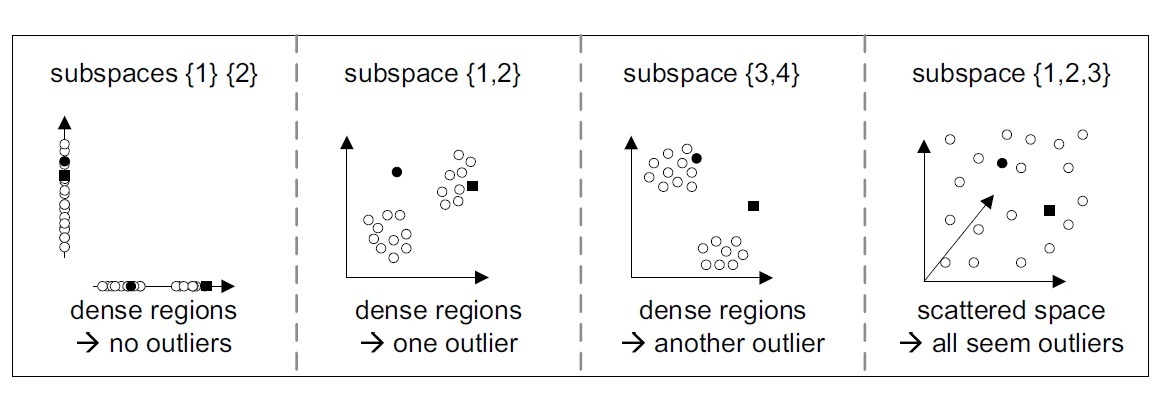}}\vfill}}
\caption{Objects known as outliers in one search space might not be outliers in another search space. Data objects might have different behaviour with respect to each feature space \cite{fifty-seven}.}
\label{Outliers_Search_Space}
\end{center}
\end{figure}

A dataset may contain objects that do not fit the model of the data and do not obey the discovered patterns, which are called outliers \cite{fifty-four}. Outliers are important because they might change the behaviour of the model as they are far from the discovered patterns and are mostly known as noise or exceptions. Outliers are useful in some applications such as fraud and anomaly detection approaches as the rare cases are more interesting than the normal cases \cite{fifty-five}. Outlier analysis is another concept of learning and mining approaches known as \textit{outlier mining} \cite{fifty-six}. The procedure of outlier detection is an issue in learning procedures, because a data object may be considered as an outlier in one dimensional search space (based on specific feature), but can be recognized as a normal object in another search space. In fact, outlier detection in high dimensional search spaces is a very complicated procedure. Fig. \ref{Outliers_Search_Space} illustrates outliers from the features point of view, which presents the issues with outlier detecting approaches.
%
%
\subsubsection{Outlier Detection}
There are several approaches to detect outliers such as:
\begin{itemize}
\item{\bf{Statistical Approaches}\\}
These methods evaluate objects based on the distribution or probability models to detect outliers, which can be categorized in: 
\begin{itemize}
\item {block model}: outliers are removed or considered as consistent,
\item {consecutive (sequential) model}: this method evaluates the least likely object as an outlier and if this object is recognized as an outlier, then the other objects with extreme values are considered as outliers, otherwise the next least extreme object will be evaluated.
\end{itemize}
As mentioned, the outlier detection approaches try to find any distribution for data objects to detect outliers, but in high dimensional search spaces selecting any feature or a subset of features to detect outlier is not an easy task. 
%
\item{\bf{Distance Based Approaches}\\}
These methods measure the distances between objects by considering objects as outliers that are very far from the rest. The methods overcome the issues with statistical models. Main models of this type are:
\begin{itemize}
\item { index-based algorithm},
\item { nested loop algorithm}, 
\item { cell-based algorithm}.
\end{itemize}
\item{\bf{Density Based Approaches}\\}
The previous methods have some drawbacks as they depend on the distribution of the given dataset. The density-based methods evaluate the objects based on the density distribution to detect outliers.
\item{\bf {Deviation Based Approaches}\\}
These methods identify outliers by evaluating the object's characteristics in a group and if objects deviate from these characteristics, then they are considered as outliers. Eq. (\ref{dissimilarity}) evaluates dissimilarities between objects in a group and the mean value, where $n$ is the number of objects, $\bar{O}$ is the mean value, and $O_j$ can be any object in $d$ dimensional search space.
\begin{equation}
\label{dissimilarity}
\frac{1}{n} \sum_{j=1}^n (O_j- \bar{O})^2
\end{equation}
 \end{itemize}
\section{Similarity Functions}
\label{Similarity-FCM-Analysis}
In machine learning and data science, regardless of the type of methods, distance or similarity function is one of the most important parameters that affect the results. In order to find a suitable solution for the given problem, most of the methods compare the given problem with other problems. This methodology tells that the solution for the most similar problem can be the desired solution for the given problem. Due to the variety of the methods, some similarity functions in different types for different problems have been introduced. Similarity functions have been compared with each other by learning approaches, in terms of metrics and covering diversity (considering the vector and the feature spaces). Dissimilarity space or distance metric space is a pair $(O,\delta)$, where $O=\lbrace O_1, ..., O_n \rbrace $ is a set of data objects $ (O \times O \rightarrow \Re) $ in which $ (n\geq 2) $ and $ \big{(} \delta(O_j,O_l) \big{)}$ is a metric on $X$ or dissimilarity function between $(O_j $ and $ O_l )$ objects with the following properties \cite{fifty-eight}, \cite{fifty-nine}:
\begin{itemize}
\item $ \Big{(} \delta(O_j,O_l) \geq 0 \Big{)} \; \; (nonnegativity)$  
\item $ \Big{(} \delta(O_j,O_l) = \delta(O_l,O_j) \Big{)} \; \; (symmetry)$
\item $ \Big{(} \delta(O_l,O_j) = 0 \; \; if \; and \; only \; if \; l=j \Big{)} \;\;(reflexivity \; or \; 	identity) $
\item $ \Big{(} \delta(O_j,O_l) \leq \delta(O_j,O_k)  + \delta(O_k,O_l)\Big{)} \; \; $  \textit{(triangle  inequality)}
\end{itemize}
Spaces known as metric or semi-metric include the concept of distance. In metric spaces different objects can have distance 
\begin{center}
$ \Big{(} \delta(O_j,O_l) = 0 \;\;\; \; for  \; \;\;\; (O_j\neq O_l)\Big{)} $
\end{center}
but in semi-metric we have \cite{fifty-nine}.
\begin{center}
 $\Big{(} \delta(O_j,O_l) = 0  \;\;\;\; if \; and \; only \; if \;\;\; (O_j=O_l)\Big{)}$ 
 \end{center}
Distance functions for a vector $X$ in linear subspace $L$ are associated with \textit{norms} or length of a vector $O \in \Re^d$ as $ ||O|| = \sqrt{(O,O)}$, (e.g., $L_1$, $L_2$, $L_p$, and $L_\infty$ norms). In other definition, a vector norm $||O||$ is any mapping from $\Re^d$ to $\Re$ with the following three properties. 
\begin{itemize}
\item  $||O|| > 0, \; \; if \; \; O \neq 0$
\item $|| \alpha O|| = |\alpha| \;  ||O||, \;\; for \;\;  any \;\;  \alpha \in \;\;  \Re $
\item $||O_j + O_l||  \leq ||O_j|| + ||O_l||$
\end{itemize}
for any vector $ O \in \Re^n$.\\
\vspace*{-4mm}
\begin{itemize}
\item The $L_1$-norm (or 1-norm): $\Big {(}||O||_1 = \sum_{j=1}^d |O_j| \Big{)}$
\item  The $L_2$-norm (2-norm, or Euclidean norm): $\Big{(} ||O||_2 = \sqrt{\sum_{j=1}^d O_j^2} \; \Big{)}$
\item  The infinity norm (or max-norm):  $\Big{(} ||O||_{\infty} = \underset{1 \leq j \leq d}{max}  |O_j| \Big{)}$
\item $L_p$ norm: $\Big{(}||O||_p = \Big{(}\sum_{j=1}^d |x_j|^p \Big{)}^{\frac{1}{p}} \Big{)}$ 
\end{itemize}
There are different metrics in similarity (or dissimilarity) concepts and the Thesis presents the most common metric spaces in Table \ref{Metrics} \cite{fifty-nine}. Due to the variety of the methods, several similarity functions in different types and norms for different problems have been introduced. Some of the well known functions can be categorized into some families, which are briefly discussed in the following sections.
\begin{table*}[!ht]
\setlength\tabcolsep{1.0pt}
\caption{Distance/Similarity Metrics.}
\vspace*{-4mm}
\label{Metrics}
\begin{center}
\begin{tabular}{| l | l |  }
\hline
\textbf{\scriptsize Direct Metric} & \parbox{14.5cm}{\scriptsize \begin{equation} |\delta(O_j,O_l)| \leq |\delta(O_j,O_l)| + |\delta(O_l,O_k)| \;\;\;\;\;\;\;
  \forall \; O_j,O_l \; \in  O :\;\;\; \delta(O_j,O_l) = -\delta(O_l,O_j) \end{equation}} \\
 \hline
 \textbf{\scriptsize Quasi-Metric} & \parbox{14.5cm}{\scriptsize \begin{equation} \delta(O_j,O_l) \leq \delta(O_j,O_k) + \delta(O_k,O_l)  \;\;\;\; \forall \; O_j,O_l,O_k \; \in  X :\;  \delta(O_j,O_l) \geq 0 ; \; \;   if \; and \; only \; if \; (O_j=O_l)\end{equation} }  \\
 \hline
 \textbf{\scriptsize Ultrametric} & \parbox{14.5cm}{\scriptsize \begin{equation} \delta(O_j,O_l) \leq max \Big{(} \delta(O_j,O_k),\delta(O_k,O_l) \Big{)} \end{equation} } \\
 \hline
 \textbf{\scriptsize P-Metric} & \parbox{14.5cm}{\scriptsize \begin{equation} \delta(O_j,O_l) \leq \delta(O_j, O_k) + \delta(O_k, O_l) -\delta(O_j, O_k)\delta(O_k, O_l) \;\;\;\;\; \forall \; O_j,O_l,O_k \; \in  O \longmapsto [0,1] \end{equation}} \\
 \hline
 \textbf{\scriptsize Protometric} &  \parbox{14.5cm}{\scriptsize \begin{equation} \delta(O_j,O_l) \leq \delta(O_j, O_k) + \delta(O_k, O_l) - \delta(O_k, O_k) \end{equation} }\\
 
\hline
\multirow{2}{*}{\textbf{\scriptsize Partial Metric}} & \parbox{14.5cm}{\scriptsize \begin{equation} \nonumber \delta(O_j, O_l) \leq \delta(O_j, O_k) + \delta(O_k, O_l) - \delta(O_k, O_k) ; \end{equation}} \\
 & \parbox{14.5cm}{\scriptsize \begin{equation}  (O_j=O_l) \;  if \;  \delta(O_j,O_j)= \delta(O_j,O_l) = \delta(O_l, O_l) = 0 ; \; \delta(O_j, O_j) \leq \delta(O_j, O_l) \end{equation}} \\
\hline
\textbf{\scriptsize 4-Point}{\tiny Inequality} &  \parbox{14.5cm}{\scriptsize \begin{equation} \delta(O_j,O_l) + \delta(O_k, O_u) \leq  max \Big{\lbrace} \big{(} \delta(O_j, O_k) + \delta(O_l, O_u) \big{)}, \big{(} \delta(O_j, O_u) + \delta(O_l,O_k)\big{)} \Big{\rbrace} \;\;\;  \forall \; O_j,O_l,O_k,O_u \; \in  O   \end{equation}}  \\
  \hline
\multirow{2}{*}{ \textbf{\scriptsize $\delta -$ hyperblic}} &  \parbox{14.5cm}{\scriptsize
   \begin{equation} \nonumber \delta(O_j,O_l) + \delta(O_k, O_u) \leq 2\gamma + max \Big{\lbrace} \big{(} \delta(O_j,O_k) + \delta(O_l,O_u) \big{)}, \big{(} \delta(O_j, O_u) + \delta(O_l, O_k)\big{)} \Big{\rbrace}  \end{equation}}  \\
   & \parbox{14.5cm}{\scriptsize
   \begin{equation}  \forall \; O_j,O_l,O_k,O_u \; \in  O ; \;\; and \;  \gamma \geq 0 \end{equation}}\\
  \hline
 \textbf{\scriptsize Ptolemaic}  &  \parbox{14.5cm}{\scriptsize \begin{equation} \delta(O_j, O_l)\delta(O_u, O_k)\leq \delta(O_j, O_u)\delta(O_l, O_k) + \delta(O_j, O_k)\delta(O_l, O_u) \end{equation} } \\
  \hline
\end{tabular}
\end{center}
\end{table*}  
%
%
\vspace*{-7mm}
\section{Related Work}
\label{related-work}
\vspace*{-2mm}
As discussed above, the methods aim to come up with some solutions to deal with uncertainty and overlapping. One of the main causes of misclassification and miss-assignments in partitioning methods are overlapped conditions, where the methods are not completely able to partition overlapped objects. In order to detect overlapping in learning procedures, several methods and algorithms have been proposed. Sarlynuce et al. \cite{sixty} proposed a new algorithm named as "SONIC" to detect overlapping in learning procedures. Authors provided a survey of several overlapping detection approaches in their paper, by focusing on social network problems and assuming that each person can participate in more groups of friends, families, and societies. The authors worked on dynamic community and overlapping on their members. Authors paid more attention to detect overlap members in social networks by monitoring their new vertices on proposed graphs.\\
Xie et al., \cite{sixty-one} categorized different overlapping methods into different categories named as: "Clique percolation", "Link partitioning", "Local expansion and optimization", "Fuzzy detection", and "Agent based". Authors paid less attention to the occurrence of the problems related to overlapping, instead, they mostly focused on overlap detection strategies. Authors knew there are some members that are capable of participating in other groups, where they mainly worked to detect objects in each group. Das at el., \cite{sixty-two} proposed a new approach in overlapping detection problems. Authors made use of genetic algorithms to analyse overlapping in clustering problems. The same scenario was introduced by Whang st al., \cite{sixty-three} using the seed expansion technique instead of genetic algorithms. Authors worked on detection analysis with a brief study on the behaviour of data objects in overlap clustering. \\
In general, to deal with the participation of data objects in one or more clusters (overlapping), first we need to identify the properties of different types of data objects. For this purpose, several solutions have been introduced. Data objects are considered as fundamental keys in learning methods that without the objects the mining algorithms are mostly useless. Objects basically direct the accuracy of the selected algorithm in case if they are extracted from inappropriate groups. Knowing the exact type of object leads the investigators to providing a suitable environment for learning algorithms. Supervised and unsupervised  methods propose some membership functions to categorize data objects and solutions with respect to behaviour of each data category. So far, data objects are categorized into two main categories, either outlier or normal objects. Outliers are those that do not fit the model of the data (data pattern). Unlike outliers that do not fit the model of data, normal data objects fit into one model of data from datasets. Normal objects can be easily processed with respect to the data patterns.\\
Aggarwall \cite{three} worked on model-based methodologies by considering the models of the data as the bases of data engineering in outlier detection. The author gathered useful information related to probabilistic, statistical, linear, proximately-based, and information theoretic models. Using the model-based approaches in outlier analysis is fully discussed in the paper by working on several approaches such as statistical tail confident test, depth-based, deviation-based, angel-based, distance-distribution-based: (cell-based, index-based, and reverse nearest-based), and density-based models: (local outlier factor, local correlation integral, histogram-based, and kernel density estimation). \\
Campos et al., \cite{sixty-four} discussed an empirical study of outlier detection. Authors made use of unsupervised methods to evaluate data objects in detection procedures. They also studied some outliers' detection measurements. As outliers do not follow any model presented by data, outliers are mostly considered as noise and in most applications and methods these objects are being removed from datasets. But in some special applications such as anomaly detection, fraud detection, and security these objects are considered as special objects in the learning procedures. Nguyen et al., \cite{sixty-five} made use of outlier detection techniques to check any abnormality for securing the systems from any cyber attack in supervised methods. The same strategy has been used by Kannan et al., \cite{fifty-five} to detect any intrusion in networks by using fuzzy methods in their procedure. Both methods work based on feature selection strategies to detect abnormalities in different search spaces. Su et al., \cite{forty-nine} and Yu et al.,\cite{fifty-six} tried to use learning algorithms to detect any abnormality in networks, where in the former paper, authors made use of a combined method of genetic and fuzzy approaches, but in the latter one authors just used a fuzzy method in their detection's procedures.\\
In big data applications, knowing the exact type of objects and selecting the most accurate similarity and membership assignments are crucial to cut the extra costs and obtaining better performance. Redoing the learning algorithms in big data is not reasonable and sometimes is impossible, and for preventing such circumstances, we need to study the type of data in advance. Providing the relaxed environment for data objects is very crucial for learning algorithms. Regarding membership assignments in learning procedures, different methodologies and algorithms have been proposed. Dunn \cite{sixty-six} proposed a method called ISODATA to apply fuzzy methodology in clustering problems, which the idea was followed by Bezdek \cite{sixty-seven} to work on fuzzy c-means in partitioning methods. In fuzzy membership functions data objects may have partial nonzero membership in several clusters, but only full membership in one and only one cluster as ($\sum_{i=1}^c u_{ij} =1$).\\
Havens et al. \cite{forty-six} introduced some algorithms in both centroid and kernel based methodologies to improve the accuracy of the partitioning algorithms using fuzzy methods. Authors applied the algorithms in large datasets with the aim of comparing the results of their algorithms on different datasets. They discussed about some parameters such as similarity and membership functions that affected the accuracy of their proposed algorithms. Authors tried to overcome some of the issues in similarity functions by assigning some weights to data objects. Wang et al. \cite{fifty-seven} made use of the same strategies in clustering problems by adding some weights to the features in their clustering procedures. They used the main centroid-based fuzzy c-means algorithm to update members of clusters and the centroids. They also faced the issues with membership assignments in partitioning methods. Authors aimed to handle the influences of similarity functions, as causes of miss-assignments, by adding weights to some features. They were able to improve the accuracy of the proposed algorithms in comparison with the previous methodologies. \\
Huang et al. \cite{sixty-eight} discussed about kernel-based methods using fuzzy theory. Authors used fuzzy c-means algorithm, in addition to find the most important features using eigenvectors. Their method, called as multi kernel fuzzy clustering, was designed to handle the problems by spherical clusters (normalized vectors). Different values of fuzzification constant have been used in their algorithms in their membership assignments. Eschrich et al. \cite{sixty-nine} studied the speed and complexity of fuzzy methods. Authors tried to introduce some algorithms to speed up the procedure of clustering algorithms. The main goal of the paper was to reduce the dimensionality of the search space to obtain the results in a better and faster way. Authors found that there are some issues with learning procedures that affect the final results. They also found that similarity and membership functions are being trapped by some features so they tried to use feature selection techniques to remove those features from the learning procedures. The issues with similarity and membership functions were temporarily handled for some datasets by removing some features from the datasets.\\
Hatawaya et al. \cite{seventy} compared fuzzy and probabilistic methods in different criteria in clustering problems. The authors applied these methods in large datasets by introducing some algorithms as centroid-based approaches. Authors paid more attention to speed of their algorithms. They mostly discussed how to deal with large datasets with less attention to similarity and membership functions. They also used different strategies in their similarity procedures without discussing deeply about the issues that influence the results. Kosko \cite{seventy-one} discussed about fuzziness in comparison with probability. Uncertainty can be covered by both methodologies according to the paper, but there were some discussion points that fuzzy sets and methods are different from probability methods. The paper did not cover the issues that both of these methods face with, but instead the methods were compared. The main idea of the paper was to statistically explore how these methods work. All these methods and algorithms have been proposed to cover overlapped clustering and uncertainty in order to increase the accuracy of the introduced methods in partitioning problems, as covering uncertainty is a key factor for the proposed methods.\\
Decision trees and hierarchical methods made use of crisp theory in their partitioning procedures. Some other methods such as support vector machine (SVM) work on dimensionality of the search space by moving from higher dimensional search space to lower or vice versa in order to get the objects separated from each other \cite{seventy-two}. To deal with overlapping, introduced decision tree and hierarchical methods were combined by the other methods such as fuzzy and  probability methods. Probability and fuzzy theory have been used in some partitioning methods in their membership assignments. Cannon at al. \cite{seventy-three} proposed a method to improve the accuracy of fuzzy method with assigning some weights and good implementation. Xu at el. \cite{seventy-four} studied the fuzzy methods by exploring different types of partitioning approaches without discussing the issues with fuzzy methods in details. Fuzzy membership functions have been widely used in combination with other machine learning methods \cite{seventy-five}. In probability methods \cite{seventy-six} as same as fuzzy methods, objects are allowed to participate in other clusters to cover overlapping. \\
As discussed earlier, there are some challenges with proposed methodologies on partitioning problems, dealing with uncertainty, which encouraged investigators to come up with new ideas. Linda et al., \cite{six} proposed a new method by applying fuzzy type-II (interval-valued type-II and generalized type-II) in membership assignments. Their algorithms made use of upper and lower boundaries $[ \underline{u_{ij}} (x) \; , \; \overline{u_{ij}} (x) ] $ in fuzzification procedures. All the presented methods aimed to come up with some ideas to deal with uncertainty. Fuzzy, probability, and possibilistic methods have become more popular.\\
Possibilistic theory, studied by Zadeh \cite{twenty-five}, discusses about possibility of happing of events. The possibilistic theory has been used by Krishnapuram et al., \cite{thirteen}, which consequently led to possibilistic clustering method (PCM), presented by Eq. (\ref{Possibilistic-M}). Authors' ideas were based on the same ideas from previous methods by providing more flexibilities. Some examples have been proposed in their paper to demonstrate how data objects should be clustered with less limitations (relaxing the fuzzy condition). Discussed problems in their paper became more popular and gradually became the basis for new partitioning strategies in recent methods. The method provided a flexible and wider search space for partitioning problems in comparison with previous methods. The proposed algorithms made use of the possibilistic theory to partition data objects in order to allow the object to participate in more clusters. The proposed methods obtained desirable results in some datasets, but for some criteria and some datasets the results of other methods were better.\\
Pal et al. \cite{seventy-seven} applied possibilistic method in fuzzy c-means algorithms. As there were some issues with membership assignments in fuzzy and probability methods that affect the final results, the proposed method have been used to allow members to participate in other clusters. There were some arguments on the paper on improving membership assignments. Authors used the same centroid-based method (fuzzy c-means) by using possibilistic membership assignments. Anderson et al. \cite{seventy-eight} compared fuzzy, probability, and possibilistic methods in partitioning criteria. The authors made use of some validity indices to compare the accuracy of each one of the proposed partitioning methodologies. The paper discussed about fuzzy validity functions to check the ability of methods on their procedures. Authors studied different membership assignments by exploring some differences in membership functions that lead the method to obtaining better results. Impact of membership functions on final results and partitioning methods were briefly discussed in this paper.\\
Havens et al. \cite{seventy-nine} made use of the possibilistic method in their membership assignments to speed up the functionalities of the algorithms. Authors made use of the basic algorithm of fuzzy c-means to update the centroids and the objects' membership degrees in each iteration. Their membership assignments resulted in obtaining better outcomes. In similar way, Zarandi et al. \cite{eighty} made use of possibilistic and fuzzy methods to obtain better results. They also used the basic fuzzy c-means algorithm to update centroids and memberships in each cluster. The authors moved one more step forward to use different fuzzy membership functions in their clustering procedures. They used fuzzy type-II in comparison with fuzzy type-I. The idea was to show that the methods need to work more on uncertain conditions to provide more sophisticated methods in partitioning problems. Authors found that the membership function is one of the most crucial parameters in learning and partitioning problems so they decided to use different types of fuzzy methods to overcome the issues.\\
In general, possibilistic method is more flexible in comparison with fuzzy, probability, and crisp methods that helps investigators to deal with uncertainty. But some argued about the possibilistic method when the method was used in fuzzy c-means (centroid-based) algorithm. As a result, their implementations on such an algorithm need proper constraints and also require good initializations, otherwise the accuracy and the results are not reasonable. Challenges on possibilistic methods have been studied by Barni et al. \cite{eighty-one}. The authors implemented the possibilistic method in combination with fuzzy c-means algorithm. They found that the algorithm is being trapped by null solutions. The PCM method has some challenges in their implementation procedures. The idea of covering uncertainty and diversity by using a comprehensive membership function encouraged authors to use possibilistic method with good initializations in the first stages of learning procedures. According to PCM condition $(\underset{1 \leq i \leq c}{max} \; u_{ij} > 0)$ the trivial null solutions should be handled by some modifications on membership assignments. Investigators agreed that membership assignments should be precisely considered in learning procedures, but there were some challenges on how to use a method in partitioning problems. Vanisri \cite{twenty} introduced an algorithm to overcome the issues on trivial null solutions by changing the objective function as Eq. (\ref{General-1}). \\
\begin{equation}
\label{General-1}
J_m(U,V) = \; \sum_{i=1}^c \sum_{j=1}^n \; u_{ij}^m \;|| O_j - V_i||_{A} ^2 \;  + \sum_{i=1}^c \eta_i \sum_{j=1}^n (1- u_{ij})^m
\end{equation}
where $\eta_i$ are suitable positive numbers. The author believed that membership assignments should be relaxed enough for partitioning problems. Xenaki et al. \cite{eighty-two} proposed a new idea by using the same strategy to handle the issues with possibilistic membership assignments by proposing the objective function as Eq. (\ref{ANAP}) to make the algorithm free of null solutions.\\
\begin{equation}
\label{ANAP}
J_m(U,V) = \; \sum_{j=1}^n \Big{[}\sum_{i=1}^c \; u_{ij}^m \; \Big{|}\Big{|} O_j - V_i \Big{|}\Big{|}_{A} ^2 \Big{]} \;  + \eta_i \sum_{i=1}^c  (1- u_{ij})^m
\end{equation}
where $\eta_i$ is calculated based on the following equation.\\
 \begin{equation}
 \eta_i = \frac{\sum_{i=1}^c u_{ij} \; \Big{|}\Big{|} O_j - V_i \Big{|}\Big{|}}{\sum_{i=1}^c \; u_{ij}}, \; \; \forall j. 
 \end{equation}
Yang et al. \cite{eighty-three} made use of the same idea to initialize the method in an automatic way. The authors believed that $U_{pcm}$ can obtain different values as the implementation of PCM might be different. Authors tried to propose a new strategy to handle the issues by the proper assumptions, which consequently, AM-PCM (Automatic Merging Possibilistic Clustering Method) has been proposed by the authors to overcome the issues with possibilitic methods. Authors addressed that PCM method strongly depends on good initializations. Masulli et al. \cite{eighty-four} introduced a new approach named as "soft transition techniques"  to overcome the same problem with PCM methods in partitioning problems. Authors worked on initialization under uncertain conditions in soft clustering. Honda et al. \cite{eighty-five} compared some different approaches that make use of possibilistic membership assignments in their learning procedures adapted for fuzzy c-means functions and algorithms. Authors were able to improve the accuracy of these methods by introducing a new approach named as "PCM-II". Altintakan et al. \cite{eighty-six} aimed to make their algorithms to be free of initializations in early stages using PCM methodology, where initializations cannot be justified for all situations and problems. The authors claimed that initializations can vary with respect to the nature of each problem. Therefore, finding the best initializations in addition to the complexity of partitioning strategies mostly makes the learning procedures more complicated. \\
In almost all of these methods, all objects are being evaluated and treated in the same way. One of the reason behind the issue with the learning methods in membership assignments might be that the methods treat all the objects in a same way and the methods pay less attention to different types of data objects. Type of data is a fundamental key that needs to be considered accordingly. Different objects have different behaviour and based on their properties the methods should dynamically provide the suitable search space for data objects. Besides the type of data, similarity function is another parameter that affects the accuracy of methods. Chen et al., \cite{eighty-seven} proposed similarity metrics and they studied different norms used in similarity functions. The authors found that similarity functions have some pros and cons that they can be used in different conditions, but some of them are useful in some scenarios, where others are introduced for other datasets. The same situation has been discussed with Cha \cite{eighty-eight} by studying different similarity and distance functions. The author compared different similarity functions using some measurements related to probability density functions. As a result, the author found that similarity functions have different functionalities, which might affect the results of the methods. According to functionality of similarity functions and their influences on the methods, some sophisticated similarity functions have been proposed.\\
Cunningham \cite{fifty-nine} gathered some similarity functions from four different categories as case-based reasoning. These categories were named as Direct metrics, Transformation-based, Information-theoretic, and Emergent similarity functions. Deza et al., \cite{eighty-nine} studied several similarity and dissimilarity functions by providing their functionalities and also their pros and cons. Different metrics and norms are fully discussed by the authors. Authors studied weighted distance functions to cover some challenges with similarity and distance functions. They discovered that distance functions are not always suitable for similarity purposes, as distance functions mostly work on the vector space with less attention to feature spaces. The issues with functionalities of similarity functions in the vector space have been observed by Huang at el. \cite{ninety}, where they proposed a new unsupervised method by assigning weights to variables in their learning procedures. The same idea has been proposed by Jing at el., \cite{ninety-one} by assigning some predefined weights to variables in learning procedures. Similarity functions can be also categorized into different groups: "power distances", "distribution laws", "correlation distances", and "other distances", presented by Table \ref{Third Taxonomy}. 
\begin{table}[!ht]
\caption{Distance and Similarity Taxonomy.}
\label{Third Taxonomy}
\vspace*{-7mm}
\setlength\tabcolsep{1.0pt}
\begin{center}
\begin{tabular}{| l | l | l | l |  }
\hline
{\bf Power Distances} & {\bf Distribution Laws} & {\bf Correlation Distances} & {\bf  Other Distances}   \\
\hline
\hline
{Euclidean} & {Bhattacharya Coefficient}& {Spearman} & {Geodesic Distance}\\
 {Mahalanobis} & {$x^2$- distance} & {Kendell $\tau$ rank Correlation} & {Dice Similarity}\\
 {Simplified Mahalanobis} & {Modified $x^2$- distance} & {Pearson Product-moment}  & {Linear Kernel}\\
 {Heterogeneous Distance} & {Entropy} & {Cosine Similarity} & \\
 {Manhatan Distance} & {Standard Entropy} & & \\
  & {Conditional Entropy} & & \\
  & {Jensen Distance} & & \\
  & {Entropy} & & \\
  & {Information Gain} & & \\
  & {Kullback-Leibler} & & \\
  & {Hellinger Metric} & & \\
  & {Hellinger-Sengar Metric} & & \\
  & {Likelihood} & & \\
  \hline
\end{tabular}
\end{center}
\end{table}  
\vspace*{-5mm}
Similarity functions can be categorized into some families, which some of them are depicted by Table \ref{Second Taxonomy}, presented by Eq. (\ref{Euc11}) \textendash Eq. (\ref{Additive}) \cite{eighty-eight}. Minkowski family is also known as power distances used to measure distances between vectors. Almost all the power distances use the Minkowski family functions to measure the distance between vectors in $(L_1$ city block), $(L_2$ Euclidean), $(L_n$ Minkowski), and $(L_{}\infty$ Chebyshev) norms. This taxonomy is divided into two categories: \textit{vector} and \textit{probabilistic} approaches. $P$ and $Q$ represent data objects or probability measures in $d$ dimensional search space. $D(P,Q)$ and $S(P,Q)$ are distance and similarity functions respectively. Lvovich family discusses about functions introduced on $L_1$ family: Soresen, Gower, Soregel, Kulczyski, Canberra, and Lorentzian. Eq. (\ref{Gower}) is introduced to normalize the search space boundaries by dividing the equation by $R_j$ as a range of population in the dataset with respect to features. The method scales down the search space by dividing the equation by $d$ as the number of dimensions. Some asymmetric distance functions \big{(}Pearson, Eq. (\ref{Pearson}) \cite{ninety-two}, Neyman, Eq. (\ref{Neyman}) \cite{ninety-three}, ...\big{)} and symmetric version of those functions \big{(}squared $x^2$, Eq. (\ref{Squared}) \cite{ninety-four} and probabilistic symmetric $x^2$, Eq. (\ref{Probabilistic}) \cite{ninety-five}, ... \big{)} have been proposed.
\begin{table}[!ht]
\setlength\tabcolsep{1.0pt}
\hspace*{-10mm}
\caption{Distance functions between objects or probability measures.}
\vspace*{-4mm}
\label{Second Taxonomy}
\begin{center}
\begin{tabular}{ | c | c | c |  }
\hline
\multirow{4}{*}{\textbf{Minkowski}} 
	& {\small Euclidean } & \parbox{10.5cm}{\small \begin{equation} \label{Euc11} D_E = 
	{\Large{(}\sum\nolimits_{j=1}^d | P_j - Q_j|^2 \Large{)}^{\frac{1}{2}}} \end{equation}}\\
	\cline{2-3}
	& {\small City Block   \cite{ninety-six}}  & \parbox{10.5cm}{\small  \begin{equation} D_{CB} = {\sum\nolimits_{j=1}^d |P_j - Q_j|} \end{equation} }  \\
	\cline{2-3}
	& {\small Minkowski  \cite{ninety-six}}  & \parbox{10.5cm}{\small \begin{equation}  D_{MK} = {\Large{(}\sum\nolimits_{j=1}^d |P_j - Q_j|^p \Large{)}^{\frac{1}{p}}}  \end{equation}} \\
	\cline{2-3}
	& {\small Chebyshev \cite{ninety-seven}}  & \parbox{10.5cm}{\small \begin{equation}  D_{Cheb} = max_j |P_j - Q_j|  \end{equation}}\\
\hline
\multirow{4}{*}{\textbf{Lvovich}} 
 & {\small Sorensen \cite{ninety-eight}}& \parbox{10.5cm}{\small \begin{equation}  D_{Sor}=\frac{\sum_{j=1}^d |P_j - Q_j|}{\sum_{j=1}^d (P_j + Q_j)} \end{equation}} \\
		\cline{2-3}
 {\textbf{Family}} & {\small Gower  \cite{ninety-nine}} & \parbox{5.4cm}{\small \begin{equation} \label{Gower} D_{Gov}= \frac{1}{d} \sum\nolimits_{j=1}^d \frac{|P_j - Q_j|}{R_j}  \end{equation}} 
  \parbox{4.4cm}{\small \begin{equation} D_{Gov}= \frac{1}{d} \sum\nolimits_{j=1}^d |P_j - Q_j|  \end{equation}} \\
  		\cline{2-3}
  \textbf{$L_1$ Family}		& {\small Soergel \cite{hundred}} & \parbox{10.5cm}{\small \begin{equation} D_{Sg}= \frac{\sum_{j=1}^d |P_j - Q_j|}{\sum_{j=1}^d max(P_j, Q_j)}  \end{equation}}  \\
  		\cline{2-3}
   		& {\small Kulczynski \cite{hundred-one}} & \parbox{10.5cm}{\small \begin{equation} D_{Sg}= \frac{\sum_{j=1}^d |P_j - Q_j|}{\sum\nolimits_{j=1}^d min(P_j, Q_j)}  \end{equation}} \\
   		\cline{2-3}
   		& {\small Canberra  \cite{hundred-one}} &  \parbox{10.5cm}{\small \begin{equation} D_{Can}= \sum\nolimits_{j=1}^d \frac{ |P_j - Q_j|}{ P_j + Q_j}  \end{equation}} \\
   		\cline{2-3}
   		& {\small Lorentzian \cite{hundred-one}} &  \parbox{10.5cm}{\small \begin{equation} D_{Lor}= \sum\nolimits_{j=1}^d ln( 1+ |P_j - Q_j| )\end{equation}}\\
\hline
\multirow{8}{*}{\textbf{$x^2$ Family}}        
	& {\small Squared  Euclidean} & \parbox{10.5cm}{\small \begin{equation} \label{Eucl} D_{SE} = \sum\nolimits_{j=1}^d (P_j -Q_j)^2 \end{equation}}\\
		\cline{2-3}
	& {\small Pearson   $x^2$    \cite{ninety-two}}  & \parbox{10.5cm}{\small \begin{equation} \label{Pearson} D_{P} = \sum\nolimits_{j=1}^d \frac{(P_j -Q_j)^2}{Q_j} \end{equation}} \\
	\cline{2-3}
    & {\small Neyman    $x^2$   \cite{ninety-three}}   & \parbox{10.5cm}{\small \begin{equation} \label{Neyman} D_{N} = \sum\nolimits_{j=1}^d \frac{(P_j -Q_j)^2}{P_j} \end{equation}}\\
    \cline{2-3}
 \textbf{$L_2$ Family}   & {\small Squared $x^2$ \cite{ninety-four}}  &  \parbox{10.5cm}{\small \begin{equation} \label{Squared} D_{SQ} = \sum\nolimits_{j=1}^d \frac{(P_j -Q_j)^2}{P_j+Q_j} \end{equation}}\\
    \cline{2-3}
    & {\small Probabilistic \cite{ninety-five}} &  \parbox{10.5cm}{\small \begin{equation} \label{Probabilistic} D_{PSQ} = 2 \sum\nolimits_{j=1}^d \frac{(P_j -Q_j)^2}{P_j+Q_j} \end{equation}}\\
    \cline{2-3}
    & {\small Divergence \cite{ninety-five}}   &  \parbox{10.5cm}{\small \begin{equation} D_{Div} = 2 \sum\nolimits_{j=1}^d \frac{(P_j -Q_j)^2}{(P_j+Q_j)^2} \end{equation}}\\
    \cline{2-3}
    & {\small Clark \cite{hundred-one} }      &  \parbox{10.5cm}{ \small \begin{equation} D_{Clk} = {\Large{(}\sum\nolimits_{j=1}^d \big{(} \frac{(P_j - Q_j)}{(P_j+Q_j)} \big{)}^2 \Large{)}^{\frac{1}{2}} } \end{equation}}\\
    \cline{2-3}    
     & {\small Additive $x^2$ \cite{hundred-one}} & \parbox{10.5cm}{\small \begin{equation} \label{Additive} D_{Ad} = \sum\nolimits_{j=1}^d  \frac{(P_j -Q_j)^2 (P_j + Q_j)}{(P_jQ_j)}  \end{equation}} \\		
\hline
\end{tabular}
\end{center}
\end{table}
More families of similarity functions are presented by Table \ref{First Taxonomy}. There are also other useful distance functions such as distance functions on histograms, signatures, and probability density that will not be discussed in this Thesis. For more information regarding distance functions, metrics and different methods on similarity functions please see \cite{eighty-nine}.
\begin{table*}[!ht]
\setlength\tabcolsep{0.9pt}
\caption{Taxonomy of Similarity Families.}
\label{First Taxonomy}
\begin{center}
\begin{tabular}{| l | l | l | l | l | l | l | l | }
\hline
\textbf{\small Minkowski} & \textbf{\small Lvovich} & \textbf{\small Fidelity} & \textbf{\small $x^2$ Family} & \textbf{\small Intersection} & \textbf{\small Inner Product} &  \textbf{\small Entropy} & \textbf{\small Combination}  \\
\hline
\hline
 {\small Euclidean} &  {\small Sorensen} & {\small Fidelity}  &  {\small Euclidean} & {\small Intersection}  &  {\small Inner Product}  &  {\small Kullback} & {\small Teneja}  \\
{\small City Block  } & {\small Gower}   & {\small Bhatacharya}  & {\small Pearson $x^2$} & {\small Wave Hedges}  & {\small Harmonic {\tiny Mean}}  & {\small Jeffreys} & {\small Kumar-{\tiny Johnson}}\\
{\small Minkowski}  & {\small Soergel}  & {\small Hellinger}  & {\small Neyman $x^2$}     & {\small Czekanowski}  & {\small Cosine}  &  {\small K-{\tiny Divergence}} & {\small AVG$(L_1, L_{})$}\\
{\small  Chebyshev}  &  {\small Kulczynski}  & {\small Matusita} & {\small Squared $x^2$} & {\small Motyka}  & {\small PCE}  & {\small Jensen,{\tiny Shannon}}  &    \\
  & {\small Canberra}   & {\small Sq.-Chord}  & {\small Probabilistic} & {\small Kulczynski}   & {\small Jaccard}  &  {\small Jensen,}{\tiny Difference}   & \\
 &    {\small Lorentzian}  & &  {\small Divergence}   & {\small Ruzicka}   &  {\small Dice} &  {\small Topsoe}     &  \\
  & & & {\small  Clark } & {\small Tanimoto}   & &  & \\
 & & & {\small  Additive $x^2$ } &  & &  & \\
  \hline
\end{tabular}
\end{center}
\end{table*}  
Discussed distance functions mostly work on the vector space, but in high dimensional search space we need to cover feature spaces to prevent miss calculation in learning procedures. Some recent similarity functions performed on both the vector and the feature spaces with some challenges. Some achievements from \cite{hundred-two}, \cite{hundred-three} show that how similarity functions are crucial in learning methods that need to be well evaluated. Modha et al. \cite{hundred-two} considered features of each object as heterogeneous and homogeneous types which should be addressed in similarity functions. The authors provided some weighted distance functions in kernel-based methods to obtain promising results. Salzberg \cite{hundred-three} assigned weights to Euclidean distance function to obtain better results using nested generalized examplar (NGE). Some other weighted similarity functions have been proposed to cover diversity in both the vector and the feature spaces, but there are some issues with those similarity functions that will be fully discussed in Chapter \ref{WFD-Chap}. 
%
%
\section{Conclusion and Proposed Solutions}
As mentioned in this chapter, it is clear that accuracy of the methods depends of different parameters that less consideration for these parameters results in improper accuracy. Some of the most important parameters namely data type, membership function, and similarity functions have been fully discussed in this chapter. Generally, most of the learning approaches need to evaluate the membership assignments and similarity functions in their learning procedures. Obtaining proper accuracy is not doable without providing the accurate membership function, applying proper similarity functions, and treating objects with respect to their nature. So the goal of this Thesis is to improve the accuracy of the methods, by considering the important parameters that affect the results. In addition, this Thesis defines some hypotheses regarding membership assignments, similarity measurements, and type of data, whether learning methods need to consider new membership and similarity functions by considering data types, or the existing membership and similarity functions and data types are sufficient to obtain the reasonable results. To answer the hypothesis, Thesis fully analyses the issues on membership functions by illustrating some examples in the wider perspectives. The necessity of considering a comprehensive membership function on data objects is being highlighted when intelligent systems are used to implement mathematical equations. With regard to learning methods, this Thesis plans to answer the following question: 
\begin{itemize}
\item which learning method can provide both prediction and prevention strategies, besides partitioning capabilities and differentiating between different types of data objects? 

\end{itemize}
According to the question, the Thesis aims to find an accurate answer by introducing a new method and functions in learning procedures. Extracting knowledge from data is a sophisticated procedure that needs to be addressed by a comprehensive method. Regardless of the type of learning approaches, membership assignments can be selected from the most common membership functions, crisp, fuzzy, probability, and possibilistic so far. Some combinations of these methods can be used in different approaches. Obtaining the desirable accuracy extremely depends on how the methods are capable of covering diversity, dealing with overlapping, analysing mutation, providing the flexible search space for objects, and considering all types of objects in their learning procedure. As a result, this Thesis aims to answer the following question with respect to membership assignments:
\begin{itemize}
\item what type of membership function can provide a flexible searching space for all type of data objects to  facilitate the procedure of objects' behaviour analysis and objects' movement analysis?
\end{itemize}
The Thesis also aims to answer the following question related to different types of data objects:
%
\begin{itemize}
\item is it sufficient to categorize objects into just normal objects and outliers, and how learning methods and membership functions can be influenced by different types of objects?
\end{itemize}
And finally, the following question will be answered by the Thesis according to similarity functions: 
%
\begin{itemize}

\item what type of feature can affect the result of similarity functions, and how to handle the impact of such features?
\end{itemize}
%
%
%
%
\chapter{Critical Objects}
\label{Outstanding-Chap}
In most of the machine learning, data science, and statistical books, we can find useful information for data objects known as outliers. But in this chapter, the Thesis discusses about a new type of objects that is called \textit{Critical}. Critical objects are very important in many applications as they do not behave like as normal objects. The behaviour of these objects are completely different from other objects with respect to the type of algorithms and methods used in learning procedures. In following sections, the Thesis explores how these objects must be addressed to overcome the issues with other methods that pay less attention to critical objects. To cover the idea, some examples from different disciplines and domains are explored. Note that, the new type of object can be from any type of the presented objects in the previous chapter: univariate, multivariate, complex, and advanced objects.\\
%
%
%
%
\section{Definition of Critical Objects}
\label{outstanding}
Unlike outliers, a dataset may contain objects that do fit the model of the data and obey the discovered patterns in more than one, even all groups, classes, or clusters. These data objects are important, because they do not change the behaviour of the model as they are in the discovered patterns and are mostly known as full members. These special objects that are called "\textit{Critical}" cannot be removed from any group that they participate in. However, if they obtain small changes even in one dimension they can easily move from one cluster to another, which is the another property of critical objects. Following examples in economy, society, education, and also from the mathematical and geometrical point of view emphasize the necessity of considering critical objects in datasets. Let us recall the set of objects $O= \lbrace O_1, O_2, ..., O_n \rbrace$ in $d$ dimensional search space presented as $X = \lbrace x_1, x_2, ..., x_d\rbrace$ and a set of data models (functions) for $c$ clusters $C= \lbrace1, 2, ..., c \rbrace$ from the given dataset as $F= \lbrace f_1(x), f_2(x), ..., f_c(x)\rbrace$, $f: \; X \mapsto Y$ $; \;\; Y \subseteq\Re$.\\[0.2cm]
$O_j$ is declared as critical object if $O_j$ obtains the following condition, Eq. (\ref{Outstanding-data1}).
 \begin{equation}
 \nonumber
 \label{Outstandingdata}
 O_j \; \; is \; critical \;\;\; if \;\;\;\; 
 \end{equation}
 
 
 \begin{equation}
 \label{Outstanding-data1}
 {\Big \lbrace}  \;\;\;  \exists i, i^{'} \in C \Big{|} \;\;\; \Big{|} f_{i}(O_j) \; - \; f_{i^{'}}(O_j) \Big{|} < \varepsilon  {\Big \rbrace}
 \end{equation}
 
\begin{center}
and for some specific and very crucial domains we have Eq. (\ref{Critical-data}).\\[0.1cm]
\end{center}

%

\begin{equation}
\label{Critical-data}
 {\Big \lbrace}  \;\;\;  \exists i, i^{'} \in C \Big{|} \;\;\;  \Big{|} f_{i}(O_j) \; - \; f_{i^{'}}(O_j) \Big{|} =0  {\Big \rbrace}
 \end{equation}\\[0.2cm]
 
%
\hspace*{-7.5mm} In data science, these critical objects can be described as follow, Eq. (\ref{Outstanding-data2}). $O_j$ is known as a critical object if it obtains the partial or full membership degrees from more than one or even all groups or clusters. \\

\begin{equation}
\label{Outstanding-data2}
 {\Big \lbrace}  \;\;\;  \exists i, ..., i^{'} \in C \Big{|} \;\;\; 0\; < \; u_{ij} \leq 1 \;\;\;\;\; \&  \;\;\; ... \;\; \& \;\; 0\; < \; u_{i^{'}j} \leq 1  {\Big \rbrace}
\end{equation}\\
where $u_{ij}$ is the membership degree of $j^{th}$ object for the $i^{th}$ group or cluster and its interval is defined as $(0\; < \; u_{ij} \leq 1)$, where in very special cases membership degrees are equal or very close to one $ (u_{ij} = 1 \; \; \& \;\;\;  u_{i^{'}j} =1)$. 
In general, critical objects are those that follow the data patterns of more than one data model. In other words, a critical object can fit into more than one data model. From the membership point of view, a critical object can obtain membership degrees from more than one category. Objects can be overlapped in some clusters, but they are not necessarily being considered as critical objects.\\
For brightening the functionalities and behaviour of critical objects in real life and applications, the Thesis provides several examples to show the necessity of considering this type of objects in better ways. The proposed examples are selected from different disciplines to show that critical objects exist in all domains and disciplines. The functionality of learning methods with regard to critical objects will be discussed in the following chapters. \\
%
%
\section{Mathematical and Geometrical Examples}
The main idea of this section is to show how and why learning methods should perform with respect to real applications as all applications and real life activities can be mathematically modelled. In other words, the sophisticated methods should be able to provide a relaxed environment for presenting all mathematical functions and equations. To present the idea, the Thesis selects major disciplines in algebra, set theory, and geometry to illustrate what type of methods are able to apply and implement different domains in math.\\
\subsection{Critical Objects in Geometry}
\begin{figure}[!ht]
\begin{center}
\leavevmode\fbox{\parbox[b][7cm][s]{120mm}{
\vfill\footnotesize {\includegraphics[width=12cm,height=7cm]{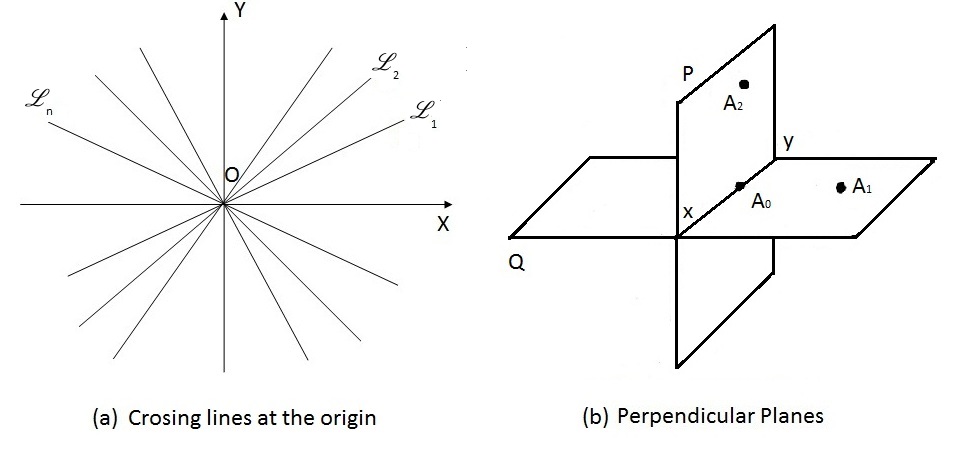}}\vfill}}
\caption{(a) Crossing lines at the origin contain the critical object (the origin), the member of all lines (clusters). (b) Two perpendicular planes contain critical objects (points) on the line $xy$ that can be assigned to another plane by small changes.}
\label{Geometry}
\end{center}
\end{figure}

Assume $ U= \{ u_{ij}(O)|O_j \in \mathscr{L}_i \}$ is a membership function that assigns a membership degree for each point $O_j$ to a line $\mathscr{L}_i$, where a line represents a cluster. Now consider Eq. (\ref{Trans-Eq}) which describes $n$ lines crossing at the origin shown by Fig. \ref{Geometry}(a).
\begin{equation}
AX =0
\label{Trans-Eq}
\end{equation}
where matrix $A$ is a $n \times d$ coefficient matrix, and $X$ is an $d \times 1$ matrix, in which $n$ is the number of lines and $d$ is the number of dimensions.
From a geometrical point of view, each line containing the origin is a subspace of $R^d$. Eq. ( \ref{Trans-Eq}) describes $n$ with its different lines as a subspace. Without the origin each one of those lines is not a subspace, since the definition of a subspace comprises the existence of the null vector as a condition in addition to other properties \cite{hundred-four}. Fig. \ref{Geometry}(b) demonstrates a similar situation for critical objects by illustrating two perpendicular planes with objects (points) on the line $xy$. The points on the line $xy$ are the full members of both planes (clusters). The point $A_0$ can be assigned to another plain by small changes. As figures explore, critical objects (the origin and all points on the line $xy$ ) have potential ability to participate in all groups (clusters) with full memberships and act such as normal objects in each cluster. On the other hand, they can move from one cluster to another by getting small changes in even one dimension. According to the figure for two and three dimensional search spaces, which can be visually illustrated, the Thesis emphasises that this type of object, critical, should be well considered. In higher dimensional search spaces we can find the same conditions, which presenting them as mathematical equations can be much easier. In linear algebra, which is the fundamental bases of kernel-based clustering methods, the null vector for $d$ dimensional search space, Eq. (\ref{null-vector}) or Eq. (\ref{null-vector-transpose}), is a fundamental key that cannot be removed from any spaces and subspaces. In fact, spaces and sub-spaces are meaningful if they contain null-vector, which this example presents the null-vector for two dimensional subspaces.
\begin{equation}
\label{null-vector}
N_v= [0,0,...,0] 
\end{equation} 
or for more convenient:
\begin{equation}
\label{null-vector-transpose}
N^{T}_v=
\begin{bmatrix}
0 \\
0\\
\vdots\\
0
\end{bmatrix}
\end{equation}\\
We should note that the kernel-based methods and other learning approaches that make use of linear algebra rely on the null vector that without the null vector the achievements cannot be obtained. The notation indicates that critical objects not only exist in all domains of geometry in different dimensions, but also losing or mistreating such objects causes irreparable consequences. The notation also emphasizes that providing a comprehensive method is mandatory in order to detect and treat critical objects.\\
\subsection{Critical Objects in Set Theory}
\begin{figure}[!ht]
\begin{center}
\leavevmode\fbox{\parbox[b][5.2cm][s]{80mm}{
\vfill\footnotesize {\includegraphics[width=8cm,height=5.2cm]{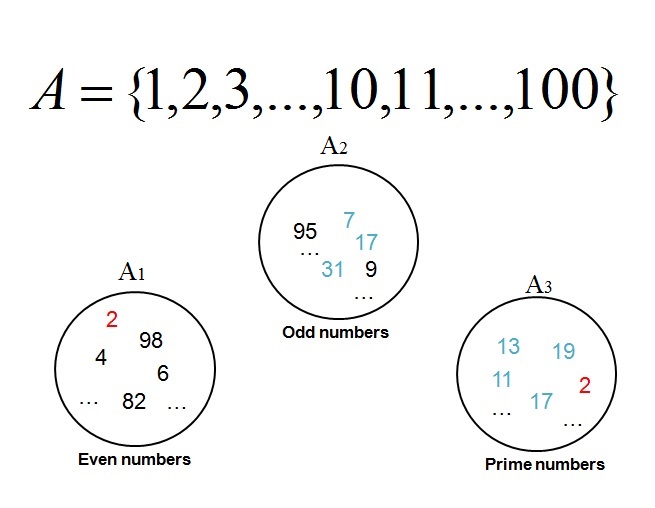}}\vfill}}
\caption{Categorizing a set of natural numbers into odd, even, or prime categories.}
\label{prime}
\end{center}
\end{figure}

\begin{figure}[!h]
\begin{center}
\leavevmode\fbox{\parbox[b][5.2cm][s]{80mm}{
\vfill\footnotesize {\includegraphics[width=8cm,height=5.2cm]{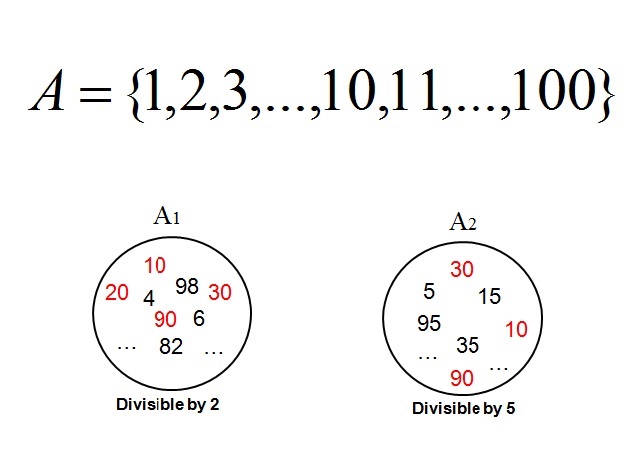}}\vfill}}
\caption{Categorizing a set of natural numbers into two sets: divisible by two or five.}
\label{divisible}
\end{center}
\end{figure}

In set theory, we can obtain some subsets from a superset according to the properties of similar members. For example, in natural numbers, we can categorize a set of numbers into different groups based on odd, even, and prime numbers that is presented by Fig. \ref{prime}. According to set theory, we can distinguish many cases that members can participate in other groups (sets or clusters) with full memberships, for instance, numbers that are divisible by two and five, Fig. \ref{divisible}. 

\hspace*{-7.5mm} By considering the examples from geometry and set theory, we can see some members that participate in more categories. These objects cannot be considered as outliers and cannot be removed from any of those categories. These objects are similar to normal objects with special characteristics. Removing them from any of the categories, that they participate in, leads to irreparable consequences. Lack of well consideration for critical objects by the methods leads to weak implementations of mathematical operations from different disciplines and domains. The challenges on learning methods to deal with critical objects will be covered in the following chapters by providing the mathematical presentations of membership assignments provided by different methods in comparison with the new method presented in this Thesis.  \\ 

%
%
\section{Examples of Critical Objects in Other Disciplines}
In recent years and due to the huge amount of data and connectivities in society and social networks, illustrating examples regarding critical objects is tangible. In social networks, each individual can be a member of different societies as a full or a partial member, e.g., one person can be a member of friends, family, and other groups. In the following section, the Thesis just explores two examples that are used in the following chapters, where different methods are compared with each other with regard to their functionalities according to critical objects.
\subsection{Example in Education and Society}
Another example can be a student from two or more departments that participate in those departments as a normal student. Assume we are planning to categorize a number of students (S) into some clusters based on their skills ($ U=\{ u_{ij}(S)|S_j \in \mathscr{D}_i \; , \; u_{ij} \rightarrow [0,1]\}$) with respect to each department \textit{$\mathscr{D}$} (Mathematics $\mathscr{D}_1$, Computer Science $\mathscr{D}_2$, Chemistry $\mathscr{D}_3$, and Physics $\mathscr{D}_4$). Again assume that a very good student (\textit{$S_1$}) with potential abilities to participate in more than one cluster ($\mathscr{D}_1$ and $\mathscr{D}_2$) as a full or normal member, as follow:
\begin{center}
 	$u_{\mathscr{D}_1} (S_1) \; = \; 1$   ,   $u_{\mathscr{D}_2}(S_2) \; = \; 1$\\ 
\end{center}
As the equations show, $S_1$ is a full member of two clusters (classes or departments) $\mathscr{D}_1$ and $\mathscr{D}_2$ out of four. As a conclusion, the critical members are capable of participating in different categories as a full or a partial member, which the learning methods need to treat them accordingly, otherwise the results are not desirable. Another example is related to economy and financial systems, where investments by customers, collaborators, and investors are very important for organizations. All of companies and organizations are willing to provide the most desirable environments for their valuable customers, employees, consumers, and investors, because they are assets and the bases of profits. Losing even a valuable member is not acceptable specially in the recent world, where customers and investors have more options in comparison with the old ages. Assume, we aim to assign memberships to boards' members of some companies as :
\begin{center}
  $ G= \{ u_{ij}(O)|O_j \in B_i \; ,\; u_{ij} \rightarrow [0,1] \}$ \\[0.4cm]
\end{center}
where ($B_i$) is the $i^{th}$ company's board, ($O_{j}$) is the $j^{th}$ member, and  $u_{ij}(O)$ is the membership degree of $O_{j}$ with respect to $B_i$, where each board or company is considered as a group (class or cluster). Again, there is a chance that a normal member participate in two or more boards. These critical objects have ability to play the main role in each board with the same ability of other normal members. The same scenario can happen for the valuable customers that might participate in two or more banks, financial organizations, and other companies. If the methods consider these types of objects as normal objects and treat them in the same ways, the organization that applied such a method will lose their customers and consequently their benefits and reputations very soon, because their customers will move to another groups. As a results, this type of example also proved that critical objects are the reasons that learning methods should be implemented accordingly, otherwise the desirable outcomes will not be reached. The mathematical presentation of membership assignments by the proposed method in the Thesis will be compared with other recent methods in the following chapters by the main consideration on critical objects. There are also some examples related to risk managements and security, which the Thesis will consider the risk managements' assessment in the experimental verification section by comparing the accuracy of the proposed method in the Thesis with the well-known classification methods such as decision trees, neural network, and nerofuzzy methods.
\subsection{Example in Security and Risk Management}
Security is the main issue for real time systems specially for financial and banking systems. Some of the customers who pay much attention to confidentiality and security on their network activities and transactions prefer to use the most secure channels, and for the others speed and the ease of services are more important. An optimized method should be a solution, but both strategies follow one common idea that any anomaly, abnormality, and intrusion should be handled in advance as the reputation of each organization is based on trust. Providing confidentiality (keeps customers' information in a safe box out of other's hands), integrity (guarantees the accuracy of information), and availability
(makes the information accessible any time and anywhere) at the same time is a big issue in all security systems. Risk managements are taken into consideration where the optimal systems have faced several risks on transactions and packets, reputation, information security, price, liquidity, foreign exchange, and compliance risks \cite{hundred-five}.  All security methods follow a common approach to detect any fault or abnormal transaction, but a comprehensive method needs to monitor and handle the behaviour of transactions (packets) in advance. Sophisticated systems need to be aware of the further movements of each transaction to prevent the system from any unwanted events beforehand.\\
Regarding the cyber threat taxonomy \cite{hundred-six}, different types of threat can be detected and prevented using machine learning approaches. Further more, the Thesis considers both classification and clustering approaches in its experimental verifications to evaluate and illustrate the ability of the proposed methods in intrusion detection and prevention systems. Cyber threats are mainly categorized in different categories: intrusion, fraud, malware, availability attacks, and abuse content, but some of the threats are commonly categorized into different groups. Cyber security is designed to protect computer systems, networks, programs, and data from any virus or attack using a set of technologies, processes, and components. Malware or virus is designed to get an unauthorized access to computer systems, which sometime replicates or spreads itself (worm), or prevents being detected (trojan horse). Malwares are also designed to collect information from the systems (spyware), to inject unsolicited advertising material (adware), to access and control the computer systems (rootkit), or to remotely take over and control computer systems (bot). Different types of malwares and viruses have been designed for different purposes and in all these cases detecting and preventing the viruses are the main goals of antiviruses, firewalls, intrusion detection (IDS), and intrusion prevention (IPS) systems.\\
Three main types of cyber analytics can be categorized as: misuse-based, anomaly-based, and hybrid approaches. Misuse-based approaches detect known attacks by using their signatures, where anomaly-based techniques model the normal network's behaviour and identify anomalies from normal behaviour. The former methods are effective without generating large number of false alarm but need frequent updates, while the latter approaches have high false alarm rates but are needed to detect new attacks, which cannot be detected by misuse-based methods. The hybrid methods make use of both strategies to reduce the false alarm rates in addition to raise the detection rates. Different methods have been used in anomaly detection techniques which this Thesis considers classification and clustering methods by applying the new method in their membership assignments. Consider a brief example related to received transactions on mobile banking or received by Internet Service Providers (ISP(s)). Again in such systems, where the main goal is to provide confidentiality, integrity, and availability at the same time, knowing the behaviour of objects or transactions is necessary. To keep the system safe and secure, the learning methods need to know in advance whether the objects are critical objects with ability to participate in both the normal and the abnormal groups, classes, or clusters. In other words, learning methods should consider the potential abilities of signals' (packets, files, or transactions) movements from the normal (or safe) to the abnormal (attack or risky) cluster or class.\\

%
%
\subsection{Example in Medicine}
Regarding the human body and human cells, we are willing to find the best treatment and also we are more interested to prevent human cells from being affected by diseases. To prevent human cells from participating in abnormal clusters, we need to check them in advance and see how far they are from abnormal (diseases) clusters. For example, we need to be aware of any movement of a normal cell to a cancer cell beforehand. The concept of critical objects helps to study the behaviour of human cells in advance to provide the most appropriate strategy (treatment) for upcoming events. In fact, we not only need to categorize human cells into healthy and unhealthy groups, but also we eager to find those critical objects that are about to move from a healthy to an unhealthy group or vice versa. In recent years and according to biology and bioinformatics, considering critical objects is very important, while scientists are looking for such objects in prediction, preventions, and sometimes for encouragement strategies. There are more examples in medicine, which will be fully discussed in the following chapters by using a dataset related to lung cancer from Harvard medical school for experimental verifications. The Thesis makes use of the dataset to propose a new learning approach to treat critical objects in addition to treat other objects in a comprehensive way. The Thesis also considers the issues with other methods with regard to all types of objects in different criteria in the following chapters.  
\section{New Data Type Taxonomy}
In learning methods, one of the most important factors is the type of data. Lack of well consideration for data types misleads the methods to recognize objects from their own categories and eventually results in losing needed information. The first and the most important factor in the methods is the ability to differentiate data objects in order to choose the most accurate approaches in learning procedures with respect to each particular object. Each type of object has its own properties and influences on final results, regardless of the type of the method. To differentiate data types and their impacts on learning approaches, analysing data types from different points of view is indispensable. This Thesis presents different types of objects with regard to their behaviour and their structures. The Thesis proposes a new data-type's taxonomy that categorizes data types in Structural-based and behavioural-based categories as two main data type categories, where each one of the categories includes different subcategories, briefly explored as follows.\\
\subsection{Structural-Based Category}
Data objects are categorized into different groups according to their structures listed as single or multivariable, complex, and advanced objects, presented as:
\begin{itemize}
\item single or multi-variable objects,
\item complex objects:
\begin{itemize}
\item structured data object: HTML files,
\item semi-structured data object: XML files,
\item unstructured data object: text files,
\item spatial data objects: maps/medical images,
\item hypertext: messages, reports, documents, ...,
\item multimedia: videos, audio, musics, and ... .
\end{itemize}
\item advanced objects:
\begin{itemize}
\item sequential patterns,
\item graph and sub-graph patterns,
\item objects in interconnected networks,
\item data stream, or stream data,
\item time series.
\end{itemize}
\end{itemize}
\subsection{Behavioural-Based Category}
In addition to the structure of data object, learning methods need to consider the behaviour of each object individually to evaluate the influences of each particular object. According to model-based approaches, objects are divided into different categories by attempting to optimize the fit between the data and some mathematical models, when the objects are generated by a mixture of underlying probability distributions. The data model can be extracted from a Gaussian mixture, a regression based, or a proximity-based model. This Thesis proposes a new data type taxonomy called "Behavioural-based". Data objects in this taxonomy are categorized into three main categories with respect to their behaviour known as Normal, Outlier, and Critical objects.
\begin{itemize}
\item \textit{Normal objects:} an object can be considered as a normal object, if the object follows one of the discovered data patterns.
\item \textit{Outliers:} data objects that do not fit the model of the data and do not obey the discovered patterns of the data are considered as outliers. Outliers are far from the rest of objects in datasets and may change the behaviour of the model if they are considered in learning procedures and measurements. Outliers are mostly known as noise or exceptions and are usually removed from datasets in most applications. There are however some applications that perform based on anomaly analysis, while some statistical distributions and probability models are used to check the occurrence of outliers.
\item \textit{Critical objects:} unlike outliers, a dataset may contain objects that follow the discovered patterns and do fit into more than one even all data models. The model and pattern of the data remain unchanged by considering critical objects as these objects obey the patterns of the data and they are in the discovered patterns. Critical objects are mostly known as partial or full members of several clusters (classes or partitions). Removing critical objects from datasets results in losing useful information, so these objects cannot be removed from any cluster or class that they participate in. The other important property of critical objects is that they have potential ability to move from one cluster or class to another (object's movement) by getting small changes even in one dimension in the feature spaces.
\end{itemize}
More details in critical objects and the methods proposed by the Thesis to deal with critical objects will be fully presented in the following chapters. The verifications and comparison of the methods proposed by the Thesis with other learning methods, besides illustrating some mathematical examples will be deeply discussed in the following chapters.
%
%
\chapter{Dominant Features and New Similarity Functions}
\label{WFD-Chap}
Due to the growth of data in recent years, sophisticated algorithms in data science have been introduced \cite{hundred-seven}. Most of the introduced algorithms make use of the well known methods such as sampling, data condensation, density-based approaches, grid-based approaches, divide and conquer, incremental learning, and distributed computing to deal with big data \cite{hundred-eight}, \cite{hundred-nine}. Although new processors are fast enough to help investigators to work on huge amount of data, extracting information from big data with advanced devices are costly and investigators should reduce the amount of cost in learning procedures. So, selecting the most suitable and accurate methods to prevent redoing the learning algorithms is the proper solution to obtain the accurate results without any extra cost. In learning methods, some important factors such as similarity functions, types of data object, and membership functions affect the final results.\\
Allowing data object to participate in more categories is the aim of the more sophisticated methodologies, but in such a scenario some methods were able to prepare a suitable environment and some were not. In order to provide the flexible environment, learning algorithms should work on both the vector and the feature spaces. As mentioned, similarity function is one the important factor that may or may not lead to obtaining desirable accuracy. In this chapter, the Thesis focuses on similarity functions and the parameters that affect the accuracy of similarity functions, consequently the accuracy of the selected methods. Most of the similarity functions work on the vector space without considering the feature spaces. This Thesis proposes a new type of feature called "Dominant" that easily affect the accuracy of almost all of similarity functions that perform in the vector space and the accuracy of some of the similarity functions that work on both the feature and the vector spaces. Dominant features have high impact on the final result of the methods. In the following sections, the Thesis demonstrates how dominant features can skew the final result, specially when we work on high dimensional search spaces. 
%
%
\section{Issues with Similarity Functions}
\label{Issues-similarity}
Assume in either clustering or classification problems, there are two objects in a four dimensional search space as $O_{1} = (2,2,2,5) $ and $O_{2} = (3,1,3,1)$ and a centroid (or prototype) $P =(2,2,2,2)$. Now assume that some methods and similarity functions are used to calculate the similarity between the objects and the prototype. If we use a distance function such as Euclidean distance, object $O_{2}$ seems overall more similar to the centroid, but from a features' perspective, $O_{1}$ is more similar to the centroid when compared with $O_{2}$ given that they share two out of three features. This example motivates the following distance functions. These functions can be applied in high dimensional search spaces $(d)$, typical of big data applications where $d$ is a very large number. Let us consider the following example to demonstrate the issues with Euclidean distance function.\\
\[O^{'}_{1} = (2,2,2,...,2, x) , \]
\[O^{'}_{2} = (3,1,3,1,...,3,1) , \]
\[P^{'}=(2,2,...,2)\]
where \\[0.2cm]
\[O^{'}_{1,1} = O^{'}_{1,2} = O^{'}_{1,3} = ... = O^{'}_{1, d-1} = 2\]
  \; \; \; \; \; and 
  \[ O^{'}_{1,d} = x > (\sqrt{d}+2) \] \\[0.2 cm]
\begin{equation}
\nonumber
P^{'}_{1} = P^{'}_{2} = P^{'}_{3} = ... = P^{'}_{d} = 2 
\end{equation}\\
According to all distance (or similarity) functions presented in Chapter \ref{State-Chap}, we see how these functions challenge the issues on the feature spaces. We almost have some features such as $(O^{'}_{1,d} = x)$ that influence the results of distance functions. The impact of such features will lead algorithms to improper accuracy. In most crucial systems such impacts are not repairable and similarity functions should evaluate the features of data objects from different perspectives. This is because each feature has its own effect on the similarity function and a single feature should not have a large impact on the final result. \\
Another issues with similarity functions in learning strategies are related to the norm of those functions. To get a better understanding of how different norms of similarity functions affect the final results, let us consider the following example. Regarding the centroid-based algorithms discussed in the previous chapters for clustering purposes, it should be noted that the algorithm runs for some iterations and in each iteration data objects are being compared with the centroids and the new centroid will be calculated based on the degree of objects in each cluster. Now if we aim to cluster objects into two clusters, then in any iteration we need to calculate the memberships for an object $O_j$ with respect to two clusters $C_1$ and $C_2$. The influences of objects on each cluster might continue till the end of procedure. But assume that the calculations are ($u_{1j} = (0.8)^{2}$ and $u_{2j}= (0.2)^{2}$) based on the Euclidean function. \\
 \begin{equation}
 \nonumber
  u_{1j} =(0.8)^2 = 0.64 \;\;\;\;\;\;  , \; \; \; \;\;\;\;\;\;\; u_{2j}=(0.2)^2=0.04 \thickapprox 0
   \end{equation}\\
where $u_{1j}$ presents the membership degree of object $O_j$ with respect to cluster $C_1$ and likewise $u_{2j}$ presents the membership of the object for $C_2$. The assignments indicate that the influences of the object with respect to the second cluster and eventually for the next iteration is almost zero. But if we change the norm of the similarity function, then we will obtain the values as follow, which the influences of the object on the second cluster and also for the next iterations still remain. \\
 \begin{equation}
 \nonumber
u_{1j} =(0.8)^{1.2} = 0.765 \;\;\;\;\;\;  , \; \; \; \;\;\;\;\;\;\; u_{2j}=(0.2)^{1.2}= 0.145 
   \end{equation}
These examples prove that why sophisticated similarity functions are needed in learning procedures. Restricting objects of participating in other clusters might categorize them into the wrong clusters (classes) in the early stages of learning procedures and this misleading continues until the end of the procedures, which consequently categorize the objects into wrong categories. The presented examples with respect to the issues with similarity functions regarding their functionalities in different norms and also their ability to handle the impact of features highlight the necessity of considering some similarity functions that can perform in both the feature and the vector spaces to cover diversity. In addition, using such a similarity function prevents the learning methods from any miss-assignments in learning procedures. The discussed issues are the reasons for the new sets of similarity functions in different norms proposed in this Thesis with the aim of covering diversity and overcoming the issues.   \\ 
\section{Dominant Features}
\label{dominant}
Consider the set of n objects $ O = \lbrace O_1, O_2, ..., O_n \rbrace$ presented by Eq. (\ref{Object-matrix}), where each object is described by numerical $feature-vector$ data that has a form $ O_j=\lbrace x_{j1}, x_{j2}, ..., x_{jd} \rbrace \subset \Re^d$ in d dimensional search space.\\

\begin{equation}
\label{Object-matrix}
{\Large O} =
\begin{bmatrix}
x_{11} \; \; , & ...\; \;  , & x_{1d}\\
x_{21} \;\; , & ... \; \;  , & x_{2d}\\
 \vdots &  \ddots & \vdots  \\
x_{n1} \;\; , & ... \;\; , & x_{nd}\\
\end{bmatrix}
\end{equation} \\

\hspace*{-7.5mm} The partition or membership matrix can be presented by Eq. (\ref{U-matrix}) that presents the membership degrees of n objects from X for c clusters. \\

\begin{equation}
\label{U-matrix}
{\Large U} = [u_{ij}]=
\begin{bmatrix}
u_{11} \; \; , & ...\; \;  , & u_{1n}\\
u_{21} \;\; , & ... \; \;  , & u_{2n}\\
 \vdots &  \ddots & \vdots  \\
u_{c1} \;\; , & ... \;\; , & u_{cn}\\
\end{bmatrix}
\end{equation} \\

\begin{figure}[!ht]
\begin{center}
\leavevmode\fbox{\parbox[b][6cm][s]{90mm}{
\vfill\footnotesize {\includegraphics[width=9cm,height=6cm]{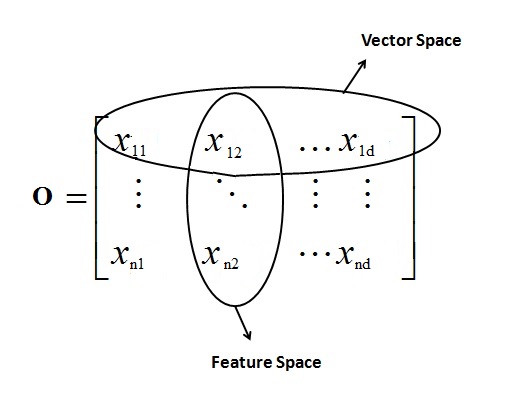}}\vfill}}
\caption{Feature and vector presentation of n data objects in d dimensional search space.}
\label{Feature-vector}
\end{center}
\end{figure}

\hspace*{-7.5mm} Most of the methods and similarity functions consider and evaluate data objects based on their total values on their vector spaces, presented by Fig. \ref{Feature-vector}. Most of the presented similarity functions compare data objects based on their vector spaces which do not cover some important factors that feature spaces provide. This kind of evaluation do not take into account any data pattern, either in the vector or the feature space. This comparison makes the functions more  vulnerable to some features that are called {\it "dominant"}. 
%
The presented vector space is used in most of similarity functions where the methods compare each data object with centroids. Regarding the figure, the feature space seems to present more information, but in realistic both the vector and the feature spaces should be considered precisely. Some may say that the vector space does not provide useful information except for comparison with centroid in similarity functions, but each space provides undeniable information. Assume a list of grades $G= \lbrace 0,1,2,3,4,5 \rbrace$ from students presented by Matrix (\ref{Grade-matrix}), where five is the highest grade. The $(n\times d)$ matrix presents the grades of $d$ courses for $n$ students, which $d=5$ in this example. \\
 \begin{equation}
\label{Grade-matrix}
{\Large G_{n\times d}} =
\begin{bmatrix}
g_{11} \; \; , & ...\; \;  , & g_{1d}\\
g_{21} \;\; , & ... \; \;  , & g_{2d}\\
 \vdots &  \ddots & \vdots  \\
g_{n1} \;\; , & ... \;\; , & g_{nd}\\
\end{bmatrix}
= 
\begin{bmatrix}
0, & 1, & 0, & 0, & 5\\
0, & 1, & 0, & 0, & 4\\
4, & 4, & 1, & 4, & 5\\
5, & 4, & 1, & 4, & 5\\
4  & \dots &    & \dots & 3\\
 \vdots &  \dots & \ddots & \dots & \vdots  \\
3, & 4, & 5, & 3, & 4\\
\end{bmatrix}
\end{equation} \\
According to the matrix, the first record $G_{1j}= \lbrace 0,1,0,0,5 \rbrace$ has a value $G_{15} = 5$ which is considered in the vector space but has given useful information. Based on other grades for the first student, we learn that either the grade is entered by mistake or the student is excellent in this particular subject. Also in the feature (column) space, the third column $G_{i3}= \lbrace0,0,1,1, \dots, 5\rbrace$ has provided useful information that the last value is significantly different from the rest of grades. These values of some features, which do not follow the pattern of the data either in the vector or the feature space, influence the result of the similarity functions. These features are defined and named {\it dominant} in this Thesis as they may influence the accuracy of learning methods. Dominant features are those that do not follow the data model, either in the feature or the vector space and sometimes any of them. As objects have some intervals (ranges, domains) for their upper and lower boundaries with respect to the feature or the vector space, the object that is considered far from the rest of values can be also selected as a dominant feature. Impact of these features affects the result of the methods, which can be handled and controlled in the way that the methods make benefit from. A dominant feature can be defined as a feature $x_f$ if the particular feature does not follow the pattern of data either in the vector or the feature space, which can be presented as follow: Eq. (\ref{dominant-1}) and Eq. (\ref{dominant-2}).\\ [0.2cm]
\begin{equation}
\label{dominant-1}
|x_f - \bar{x}| \gg  \delta^2 \;\;\;\; || \;\;\;\; |x_f| \gg |\bar{x}|
\end{equation}
\begin{center}
or\\[0.4cm]
\end{center}
\begin{equation}
\label{dominant-2}
|x_f - \bar{x}| > \lambda \delta^2 \;\;\;\; || \;\;\;\; |x_f| > \lambda |\bar{x}|
\end{equation}\\[0.2cm]
where $\lambda$ is a coefficient or threshold that can be chosen with respect to each problem or dataset, the mean value $(\bar{x})$, and variance $(\delta^2)$ can be calculated as Eq. (\ref{Mean}) and Eq. (\ref{Variance}) respectively, where $d$ is the number of features.\\
\begin{equation}
\label{Mean}
\bar{x} = \frac{1}{d} \sum_{f=1}^d (x_f) = \frac{x_1+x_2+ ... + x_d}{d}
\end{equation}\\[0.2cm]
\begin{equation}
\label{Variance}
Var(x)= \delta^2 = \frac{1}{d} \sum_{f=1}^d(x_f-\bar{x})^2= \frac{(x_1-\bar{x})^2 + (x_2-\bar{x})^2 + ... + (x_d-\bar{x})^2}{d}
\end{equation} \\[0.2cm]
Different values for $\lambda$ lead to different results for each problem, where defining the best value for $\lambda$ depends on several parameters such as type of features, ranges for feature's values, number of dimensions, and priority of each feature. In most applications, the most suitable value for $\lambda$ can be chosen by experiment. In order to see how dominant features can affect the final result of the similarity functions or the learning methods, the Thesis just considers the following example that shows how the similarity functions in different norms can be easily trapped by such features. Assume that we have two objects in five dimensional search space known as $O_1 = \lbrace 4,3,1,3,5\rbrace $ and $O_2=\lbrace 2,2,2,2,x \rbrace$ that are being compared with a centroid or prototype $P= \lbrace 2,2,2,2,2\rbrace$. $O_2$ contains a dominant feature which for convenience and also to illustrate the problem the Thesis called it $x$. Using $L_1$ norm similarity function and considering $(x=11)$ in this example, the similarity between the centroid and objects can be calculated as follow:\\
\begin{equation}
\nonumber
d_{L_1} (P,O_1) = \vert (4-2)\vert +  \vert (3-2)\vert +  \vert (1-2)\vert +  \vert (3-2)\vert +  \vert (5-2)\vert = 8 
\end{equation}\\[0.1cm]
\begin{equation}
\nonumber
d_{L_1} (P,O_2) =  \vert (2-2)\vert +  \vert (2-2)\vert +  \vert (2-2)\vert +  \vert (2-2)\vert +  \vert (2-11)\vert = 9  
\end{equation}\\
According to the $L_1$ norm distance functions, the results will be skewed if such a dominant feature occurs. Regarding the objects and the centroid, we see that four out of five features of $O_2$ are the same as the centroid's features, but the result tells $O_1$ is more similar to the centroid. This problem cannot be accepted in high dimensional search space. This type of problem can massively skew the result of similarity functions if the similarity functions are used from different norms such as the $L_2$ norm. Using the Euclidean distance function from the $L_2$ norm functions to evaluate the similarity between the objects and the centroid by considering $(x=7)$, which is calculated as follow, shows how similarity functions can be easily influenced if the methods do not pay attention to dominant features.\\
\begin{equation}
\nonumber
 d_{L_2} (P,O_1) = \sqrt{(4-2)^2 +  (3-2)^2 +  (1-2)^2 +  (3-2)^2 + (5-2)^2} = 4 
\end{equation}\\[0.2cm]
\begin{equation}
\nonumber
 d_{L_2} (P,O_2) = \sqrt{(2-2)^2 +  (2-2)^2 +  (2-2)^2 +  (2-2)^2 + (7-2)^2} = 5 
\end{equation}\\
Learning methods should consider the type of similarity functions used in their learning procedures in addition to check dominant features that might be available in datasets, otherwise the final results might not be the desirable results that we expected. In crucial systems, such as medicine and security any miscalculation will be very harmful. Using Weighted Feature Distance functions (WFD(s)) in different norms that are presented in the following sections helps to handle the impact of dominant features. In some applications, we need to make use of the impact of dominant features, e.g., in medicine we are looking for those samples that do not follow the pattern of data. Based on the values of these features, the decision will be better made. Finding dominant features can be done using different statistical measurements, which the Thesis just considers some of them in the following section. Density distribution and deviation approaches are considered as useful measurements to find and detect dominant features.
%
\section{New Weighted Feature Distance Functions \textendash{} WFD}
\label{WFD}
Let us recall the set of $n$ objects represented by $ O = \{ O_1, O_2, \;  ... \; , O_n \}$ in which each object is typically represented by numerical $feature-vector$ data, with the same priority in features, that has the form $X \;= \; \{ x_1,... \;, x_d\} \; \subset R^d $, where $d$ is the dimension of the search space or the number of features. Weighted Feature Distance $(WFD)$ is introduced to overcome the presented issues with similarity functions.
%
%
\begin{itemize}
\item { {$\bf WFD_{(L_{1})}$}}\\
Weighted feature distance $(WFD_{L_1})$ for $L_1$ norm is, Eq. (\ref{WFD $L_1$}):\\
\begin{equation}
\nonumber
 WFD_{(L_1)} \; = \; {\big{(}}|W_j O_j - W_l O_l| {\big{)}}  = 
\end{equation}
\begin{equation}
\label{WFD $L_1$}
  = \sum_{k=1}^d {\big{(}}|w_k x_{jk} - w_k^{'} x_{lk}| {\big{)}} 
\end{equation}\\[0.2cm]
%
%
\item { $\bf WFD_{(L_{2})}$}\\
 Weighted feature distance $(WFD_{L_2})$ for $L_2$ norm is, Eq. (\ref{WFD-$L_2$}):\\
 \begin{equation}
\nonumber
 WFD_{L_2} \; = \; \sqrt{\big{(} W_j O_j - W_l O_l \big{)} ^2} = 
\end{equation}
\begin{equation}
\label{WFD-$L_2$}
= \Big{(} \sum_{k=1}^d {\big{(}}|w_k x_{jk} - w_k^{'} x_{lk}|^2 {\big{)}} \Big{)} ^{(\frac{1}{2})}
\end{equation}\\
where $d$ is the number of variables or dimensions for numerical data objects. $w_k$ and $w_k^{'}$ are the weights assigned to features of the first and the second objects respectively. The Thesis introduces $(w_k =w_k^{'})$ if both objects are in the same scale. We can also obtain the Euclidean distance function from Eq. (\ref{WFD-$L_2$}) by assigning the same values to $w_k$ as: 
\[ w_1 = w_2 = ... = w_d = 1\]\\
By assigning the weights in Eq. (\ref{WFD-$L_2$}) we have:\\
\begin{equation}
\nonumber
\Big{(} \sum_{k=1}^d {\big{(}}|w_k x_{jk} - w_k^{'} x_{lk}|^2 {\big{)}} \Big{)} ^{(\frac{1}{2})} 
= \Big{(} \sum_{k=1}^d {\big{(}}|1 x_{jk} - 1 x_{lk}|^2 {\big{)}} \Big{)} ^{(\frac{1}{2})}
\end{equation}\\
which gives us the Euclidean function. That means $WFD_{L_2}$ norm is the superset of Euclidean distance function that we can obtain Euclidean function from WFD, but the opposite direction is not possible. The other $L_2$ norm distance functions discussed in this Thesis are the subsets of $WFD_{L_2}$.\\
%
%
\item { $\bf WFD_{(L_{P})}$}\\
Weighted feature distance $(WFD_{L_p})$ for $L_p$ norm is, Eq. (\ref{WFD $L_p$}):\\
\begin{equation}
\nonumber
 WFD_{(L_p)} \; = \; {\big{(}}|W_j O_j - W_l O_l|^p {\big{)}} ^{(\frac{1}{r})} = 
\end{equation}
\begin{equation}
\label{WFD $L_p$}
 = \Big{(} \sum_{k=1}^d {\big{(}}|w_k x_{jk} - w_k^{'} x_{lk}|^p {\big{)}} \Big{)}^{(\frac{1}{r})}
\end{equation}\\[0.2cm]
where $d$ is the number of variables or dimensions for numerical data objects. $p$ and $r$ are coefficients to explore different metrics, which $p$ and $r$ can be equal. $w_k$ and $w_k^{'}$ are the weights assigned to features of the first and the second objects respectively. $(w_k =w_k^{'})$ if both objects are in the same scale. It is necessary to note that weights cannot be formulated for all kind of problems, where each dataset has different characteristics and based on their characteristics different weights should be assigned to each feature or object. As same as optimization problems that we need to go through the whole search space to find the global optima, and for that we need to cover diversity to avoid being trapped by local optima. Covering diversity is a very important factor in weighted feature distance function.\\ 
\end{itemize}
%
%
\section{Prioritized Weighted Feature Distance Functions \textendash $\;$PWFD(s)}
\label{PWFD(s)}
\begin{figure}[!ht]
\begin{center}
\leavevmode\fbox{\parbox[b][5cm][s]{70mm}{
\vfill\footnotesize {\includegraphics[width=7cm,height=5cm]{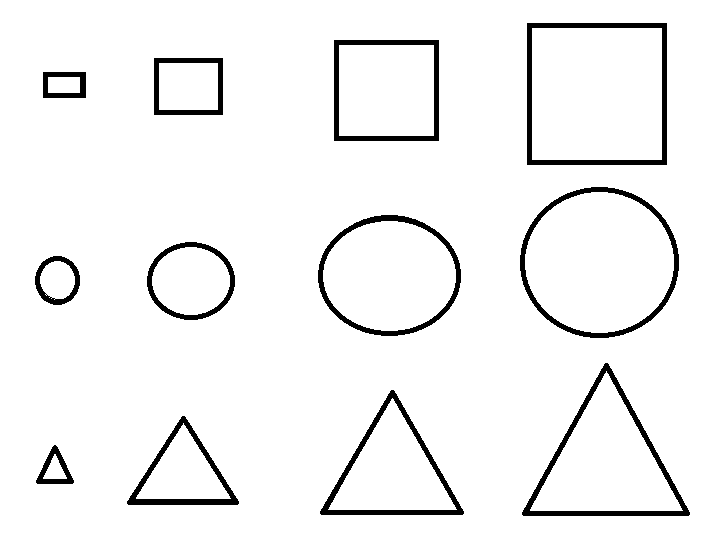}}\vfill}}
\end{center}
\caption{Presenting how similarity functions need to consider weights to cover scales in Image processing}
\label{Scale}
\end{figure}

In image processing and alike domains, we need to compare the similarity between objects that are similar, but they are different in size or scale. For instance, Fig. \ref{Scale} presents two circles with different radii that are similar, but they are in different scales. 
The assigned weights to features can be the same if the priority of features are the same, otherwise the features by higher priority can obtain higher weights. The procedure of weight assignment depends on the type of problem and the priority of the features. Besides the mentioned issues, the Thesis also adds new parameters to WFD functions to prioritize specific features to handle the impact of particular features.
%
%
\begin{itemize}
\item { $\bf PWFD_{(L_{1})}$} \\
Prioritized weighted feature distance $(PWFD_{L_1})$ for $L_1$ norm is, Eq. (\ref{PWFD $L_1$}):\\
\begin{equation}
\nonumber
 PWFD_{(L_1)} \; = \; {\big{(}} \frac{|W_j O_j - W_l O_l|}{W} {\big{)}}  = 
\end{equation}
\begin{equation}
\label{PWFD $L_1$}
  = \sum_{k=1}^d {\big{(}}\frac{|w_k x_{jk} - w_k^{'} x_{lk}|}{w_k^{\prime\prime}} {\big{)}} 
\end{equation} \\[0.2cm]

%
%
\item {$\bf PWFD_{(L_{2})}$}\\
 Prioritized weighted features distance $(PWFD_{L_2})$ for $L_2$ norm is, Eq. (\ref{PWFD $L_2$}):\\
 \begin{equation}
\nonumber
 PWFD_{(L_2)} \; = \; \sqrt{\big{(} \frac{W_j O_j - W_l O_l}{W} \big{)} ^2} =
\end{equation}
\begin{equation}
\label{PWFD $L_2$}
= \Big{(} \sum_{k=1}^d {\big{(}}\frac{|w_k x_{jk} - w_k^{'} x_{lk}|}{w_k^{\prime\prime}}{\big{)}}^2  \Big{)} ^{(\frac{1}{2})}
\end{equation}\\[0.2cm]

%
%
\item { $\bf PWFD_{(L_{p})}$}\\
Prioritized weighted feature distance $(PWFD_{L_p})$ for $L_p$ norm is, Eq. (\ref{PWFD $L_p$}):\\
\begin{equation}
\nonumber
 PWFD_{(L_p)} \; = \; {\big{(}}| \frac{W_j O_j - W_l O_l}{W}|^p {\big{)}} ^{(\frac{1}{r})} = 
\end{equation}
\begin{equation}
\label{PWFD $L_p$}
= \Big{(} \sum_{k=1}^d {\big{(}}\frac{|w_k x_{jk} - w_k^{'} x_{lk}|}{w_k^{\prime\prime}} {\big{)}}^p \Big{)}^{(\frac{1}{r})}
\end{equation}\\

where $d$ is the number of variables or dimensions for numerical data objects. $p$ and $r$ are coefficients to explore different metrics, which $p$ and $r$ can be equal. $w_k$ and $w_k^{'}$ are the weights assigned to features of the first and the second objects respectively. $(w_k =w_k^{'})$ if both objects are in the same scale. $w_k^{\prime\prime}$ is the weight coefficient to allow the method to get the priority on features with respect to other features. To get $L_p$ functions such as \textit{Minkowski}, just initialize the weights as $(w_k = w_k^{\prime} = w_k^{\prime\prime})$.
\end{itemize}
As discussed earlier, WFD(s) are the supersets of the other distance functions discussed in this Thesis with respect to each norm accordingly. WFD(s) are the subsets of PWFD(s) that we can obtain WFD(s) from PWFD(s). Note that, the proposed similarity functions are mostly designed for similarity proposes.
%
%
\vspace*{-3mm}
\chapter{\hspace*{-1mm}Bounded Fuzzy Possibilistic Method\textendash BFPM}
\label{BFPM-Chap}
%
%
%
\vspace*{-5mm}
As mentioned in the previous chapters, learning methods make use of some techniques in their learning steps such as similarity and membership functions. In the previous chapter, the Thesis discussed about normal, outliers, and critical objects which should be well considered by the methods in their learning procedures such as membership assignments and similarity assessments. In this chapter the Thesis introduces Bounded Fuzzy Possibilistic Method (BFPM) as a new learning method, as a superset of crisp, fuzzy, probability, and possibilistic methods, by providing the most flexible search space for data objects with the aim of overcoming the issues with crisp, fuzzy, probability, and possibilistic methods.
\vspace*{-2mm}
\section{Definition of BFPM}
\vspace*{-1mm}
According to the set of $n$ objects represented by $ O = \{ O_1, O_2, \;  ... \; , O_n \}$ that each object is typically described by numerical $feature-vector$ data that has the form $X \;= \; \{ x_1,... \;, x_d\} \; \subset R^d $, where $d$ is the dimension of the search space or the number of features, a cluster or a class is a set of values $\{ u_{ij}\}$, where $u$ represents a membership value, $j$ is the index to represent the $j^{th}$ object in the dataset, and $i$ is the index to present the $i^{th}$ cluster (set). A partition or membership matrix is often represented as a $c\times n$ matrix $U = [u_{ij}]$.
Bounded Fuzzy Possibilistic Method (BFPM) is defined as presented by Eq. (\ref{BFPM-Main}).\\
%
\begin{equation}
\nonumber
\label{BFPM-Main}
M_{bfpm} = \bigg\lbrace U \; \in \Re^{c\times n}| \; u_{ij} \; \in [0,1], \; \; \forall i,j; 
\end{equation}
\begin{equation}
{0 < \sum_{j=1}^n u_{ij} \leq n,  \; \;\; \forall i;  \;\;\;\;\; 0< \; 1/c \sum_{i=1}^c u_{ij} \leq 1, \;\; \forall j ; } 
  \bigg\rbrace
\end{equation}\\
where $c$ is the number of clusters and $u_{ij}$ is the membership of the object $O_j$ in cluster $i$. BFPM defines boundaries (constraints) to provide the most flexible search space for objects movement. The boundaries allow objects to participate in all clusters which overcomes the issues with possibilistic method that the learning algorithms might be trapped even if just one object obtains membership from just one cluster. BFPM, with its properties, provides the relaxed search space in comparison with other methods to evaluate the behaviour of each object with respect to even all classes or clusters. This flexibility allows objects to freely participate in other classes or clusters to show their potential abilities to participate in other groups. This idea have different advantages, where objects' movement analysis is one of them. In this chapter, the Thesis fully studies the properties of BFPM as a new learning method by giving some toy examples to demonstrate the issues with other methods in membership assignments. The chapter also presents how BFPM overcomes the issues with other methods, in addition to propose the object's movement analysis.\\
\section{Properties of the New Method}
\label{BFPCM Method}
\vspace*{-2mm}
Bounded Fuzzy Possibilistic Method (BFPM) makes it possible for data objects to have full memberships in several or even in all clusters in addition to inherit the ability of crisp, fuzzy, and possibilistic methods. This method also overcomes the issues on fuzzy and possibilistic clustering methods \cite{hundred-eleven}, \cite{hundred-twelve}. This is indicated in Eq. (\ref{BFPM-Main}) with the normalizing condition of $(1/c \sum_{i=1}^c u_{ij})$. Unlike Possibilistic method $(u_{ij} >0)$ that there is no boundary for membership functions, BFPM declares definite intervals $[0,1]$ for each data object with respect to each cluster. \\[0.2cm]
{\bf BFPM has the following properties:}\\[0.2cm]
{\bf Property 1:} \\
\hspace*{5mm}Each data object must be assigned to at least one cluster.\\[0.2cm]
{\bf Property 2:}\\
\hspace*{5mm}Each data object can potentially obtain a membership value of 1 in multiple, even in all clusters.\\[0.2cm]
{\bf Proof of Property 1:}\\
\hspace*{5mm} As the following inequality shows, each data object must participate in at least one cluster.
\begin{equation}
\nonumber
0< \; 1/c \sum_{i=1}^c u_{ij} \; \; \; \forall j;
\end{equation}
{\bf Proof of Property 2:}\\
\hspace*{5mm} With the aim of allowing objects to participate in more or even all clusters, and also avoiding any null values for memberships with respect to the clusters, in contrast to possibilistic methods, we can define the followings:
\begin{equation}
\nonumber
(0< \; u_{1j} \; \leq 1, \; \forall j \in C_1)
\end{equation}
\hspace*{70mm}$\vdots$
\vspace*{-2mm}
\begin{equation}
\nonumber
(0< \; u_{cj} \; \leq 1, \; \forall j \in C_c)
\end{equation}
Consequently, we obtain Eq. (\ref{proof2_1}) regarding the rules in fuzzy sets.\\
\begin{equation}
\label{proof2_1}
c \ast 0< \; \sum_{i=1}^{c} u_{cj} \; \leq c \ast 1, \; \forall j \in C_i
\end{equation}
By considering Eq. (\ref{proof2_1}), the above assumptions, and dividing all sides by c, we obtain the following equations:\\
\begin{equation}
\label{proof2_2}
 0< \; \frac{1}{c} \sum_{i=1}^{c} u_{cj} \; \leq 1, \; \forall j \in C_i
\end{equation}
According to the proof, each data object can potentially obtain a membership value of 1 in multiple, even in all clusters, and finally, we obtain the BFPM methodology, presented by Eq. (\ref{BFPM-Main}). \\[0.2cm]
%
The BFPM allows objects to participate in all clusters (classes) with no restrictions. Unlike crisp methods presented by Eq. (\ref{Crisp-M}), BFPM does not restrict objects to participate just in one cluster (class). BFPM can categorize objects as like as crisp methodologies when is needed, where objects do not obtain memberships from other categories. In comparison with fuzzy and probability methods, BFPM does not decrease the memberships assigned to objects with respect to all categories. Fuzzy and probability methods, because of their condition $(\sum_{i=1}^c u_{ij} =1 \;\; \forall j)$ presented by Eq. (\ref{fuzzy-Assign}), decrease the membership degrees assigned to objects when the objects participate in more clusters (classes).\\
Possibilistic methods categorize objects based on the condition $(\underset{1\leq i\leq c}{max}\; u_{ij} >0  \;\; \forall j)$, presented by Eq. (\ref{Possibilistic-M}). The condition tells that even obtaining a degree from one category satisfies the possibilistic condition, which means the implementation of possibilistic methods depends on initialization's parameters with respect to all clusters. As discussed in Chapter \ref{State-Chap}, possibilistic methods have difficulties with null solutions based on the condition that make the methods improper for processing critical objects in learning procedures. Possibilistic methods, discussed in Chapter \ref{State-Chap}, follow the same condition as fuzzy methods in each iteration, presented by the algorithm shown by Fig. \ref{centroid}. The presented algorithm prevents critical objects to present their potential abilities in their participations in other clusters. The possibilistic condition needs modifications as like as BFPM $(0< \; 1/c \sum_{i=1}^c u_{ij} \leq 1 \;\; \forall j)$, presented by Eq. (\ref{BFPM-Main}), to process and analyse critical objects and their movements from one category to other categories. As mentioned in the previous chapter, critical objects' movements should be precisely studied in advance to avoid further problems.\\[0.2cm]
Regarding the definition of critical objects ${\Big \lbrace}   \exists i, i^{'} \in K \Big{|} \;\; \big{|} f_{i}(O_j) \; - \; f_{i^{'}}(O_j) \big{|} < \varepsilon {\Big \rbrace}$ and $  {\Big \lbrace}  \exists i, .., i^{'} \in K \Big{|} \;\; 0\; < \; u_{ij} \leq 1 \;\;, .., \;\;\; \&  \;\;\; 0\; < \; u_{i^{'}j} \leq 1  {\Big \rbrace } $, presented by Eq. (\ref{Outstanding-data1}) and Eq. (\ref{Outstanding-data2}), and the second proof of BFPM methodology, it can be concluded that BFPM removes all the limitations in  partitioning problems while crisp methods limit objects to participate in just one cluster, fuzzy and probability methods decrease the memberships of objects with respect to each cluster, and the procedure of possibilistic algorithms will be terminated with the condition even if an object obtains a membership from just one cluster ($\underset{1\leq i\leq c}{max}\; u_{ij} >0  \;\; \forall j$). According to the second proof for BFPM properties, it is clear that objects can participate in other clusters to satisfy the necessity of critical objects' processing. The processing of objects' movements is very important in prediction and preventions strategies in both supervised and unsupervised methodologies that can be obtained using BFPM. \\
In conclusion, BFPM overcomes the issues with crisp, fuzzy, probability, and possibilitic methods in their partitioning procedures in addition to provide the relaxed search space to study the objects' movements  analysis in advance which is very important for crucial systems and cannot be obtained by other methods. Based on the Eq. (\ref{BFPM-Main}) and the proofs, we can conclude the following statement: \\
\begin{equation}
\label{subsets-of-BFPM}
M_{hcn}\subset M_{fcn} \subset M_{pcn} \subset M_{bfpm}.
\end{equation}
Eq. (\ref{subsets-of-BFPM}) indicates that the method BFPM differs from the crisp, fuzzy, probability, possibilistic methods which are subsets of the new proposed method. \\[0.2cm]
The new proposed method is called 'Bounded' for different reasons: 
 \vspace*{-2mm}
\begin{itemize}
\item BFPM provides precise boundaries (constraints) for upper and lower boundaries (to remove all the limitations) with its properties and its condition $(0< \; 1/c \sum_{i=1}^c u_{ij} \leq 1, \;\; \forall j)$, in comparison with possibilistic method to overcome the issues with PCM method and its condition $(\underset{1\leq i\leq c}{max}\; u_{ij} >0  \;\; \forall j)$.
 \vspace*{-2mm}
\item BFPM also bounds (ties) both fuzzy and possibilistic method in order to provide the most flexible search space for learning algorithms with the aim of presenting a new methodology (not an alternative or modified version of other methods) in addition to keep the abilities of both fuzzy and possibilistic methodologies in covering uncertainty in supervised and unsupervised learning methodologies.
 \vspace*{-2mm}
\item BFPM allows objects to bound (jump) from one partition to another to present their potential abilities to move from one partition to another, which allows the method to evaluate object movements (mutation) analysis. This capability of BFPM allows the method to extract useful information in advance for prevention and prediction strategies. 
\end{itemize} 
%
%
\section{Issues with Learning Methods}
\label{Issues}
Issues with crisp, fuzzy, and possibilistic methods have been discussed in several papers, explored in the previous chapters. The Thesis fully analyses the issues with the learning methods by illustrating some examples in wider perspectives. The necessity of considering a comprehensive method is being highlighted when intelligent systems are used to implement and express mathematical equations and operations such as the union and the intersection in set theory. In \cite{thirteen} some examples are shown to demonstrate how data objects should be clustered with less limitations and this Thesis deeply focuses on how to remove restrictions on data objects to participate in as much clusters as they can. Lack of this consideration by the methods weakens the accuracy and the ability of the proposed method. The following example from geometry explores that clustering methods should consider the properties of data objects, besides having ability to cluster them.
\subsection{Example in Geometry}
\label{Geo_BFPM}
Let's recall $ U= \{ u_{ij}(O)|O_j \in \mathscr{L}_i \}$, the function that assigns a membership degree for each point $O_j$ to a line $\mathscr{L}_i$, where a line represents a cluster. From a geometrical point of view, each line containing the origin is a subspace of $R^d$. Now consider the Eq. ( \ref{transversal-equ}) which describes $n$ with its different lines as a subspace, where the $c \times d$ matrix is a coefficient matrix for $n$ lines in two dimensional search space. Without the origin, each of those lines is not a subspace, since the definition of a subspace comprises the existence of the null vector as a condition in addition to other properties \cite{hundred-eleven}. It should be noted that removing or decreasing the origin's membership degree with respect to each cluster ruins the property of subspaces. This example shows that other methods cannot cover the subspaces for the origin. 
\begin{equation}
\label{transversal-equ}
\begin{bmatrix}
C_{1,1} & C_{1,2}\\
C_{2,1} & C_{2,2}\\
\vdots & \vdots  \\
C_{n1} & C_{n,2}\\
\end{bmatrix}
\times
\begin{bmatrix}
Y_1\\
Y_2\\
\end{bmatrix}
=
\begin{bmatrix}
0\\
0\\
\vdots \\
0\\
\end{bmatrix}
\end{equation}
Now assume, $Y_1=0 $ , $Y_2=0$ , $Y_1=Y_2$ , and $Y_1 = -Y_2$ as equations that represent some of those lines presented by (\ref{lines-equ}) with some data objects (points) on them as shown in Table \ref{points}. Note that point (0,0) is explored once, but all lines contain that point as well.
\begin{equation}
\label{lines-equ}
\begin{bmatrix}
1 & 0\\
0 & 1\\
1 & 1  \\
1 & -1\\
\end{bmatrix}
\times
\begin{bmatrix}
Y_1\\
Y_2\\
\end{bmatrix}
=
\begin{bmatrix}
0\\
0\\
0 \\
0\\
\end{bmatrix}
\end{equation}

\begin{figure}[!ht]
\begin{center}
\leavevmode\fbox{\parbox[b][4cm][s]{60mm}{
\vfill\footnotesize {\includegraphics[width=6cm,height=4cm]{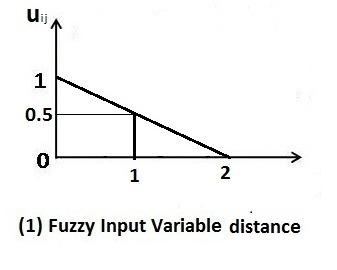}}\vfill}}
\caption{Membership function for distance input variable for the points on the crossing lines at the origin.}
\label{points-lines}
\end{center}
\end{figure}

\begin{table}[!ht]
\caption{Points on the presented crossing lines at the origin in two dimensional search space.}
\label{points}
\begin{center}
\begin{tabular}{ | c | c | c | c | }
\hline
 \underline {-1,0} &  \underline { -2,0} & -3,0 & -4,0 \\
\hline
 \underline { 1,0}  &   \underline { 2,0 }   & 3,0  & 4,0  \\
\hline
 \underline { 0,1 } &   \underline {0,2}  & 0,3  & 0,4   \\
\hline
 \underline {0,-1} &  \underline {0,-2} & 0,-3  & 0,-4  \\
 \hline
-1,-1	 & -2,-2 &  -3,-3 &  -4,-4 \\
\hline
1,-1	 & 2,-2 &  3,-3 &  4,-4 \\
\hline
-1,1	 & -2,2 &  -3,3 &  -4,4 \\
\hline
0,0	 & &  &   \\
\hline
\end{tabular}
\end{center}
\end{table}
\hspace*{-7.5mm} To clarify the idea, assume we have two of those lines $\mathscr{L}_1: \{ Y_2=0 \}$ and $\mathscr{L}_2 :\{Y_1=0\}$ with five points on each, including the origin, as shown in the following definitions:
%
\[	\mathscr{L}_1 = \lbrace p_{11}, p_{12}, p_{13}, p_{14}, p_{15} \rbrace  \]
\[= \lbrace (-1,0), (-2,0), (0,0), (1,0), (2,0) \rbrace \] \\
\[ \mathscr{L}_2 = \lbrace p_{21}, p_{22},p_{23}, p_{24}, p_{25}\rbrace \]
\[= \lbrace (0,-1), (0,-2), (0,0), (0,1), (0,2) \rbrace \] \\[0.2cm]
where $p_{ij}=(Y_1,Y_2)$. 
The origin is a member of all lines, but for convenience, the Thesis has given it different names such as $p_{13}$ and $p_{23}$ in each line above. The point distances with respect to each line and Euclidean $L_2$ norm $ \bigg{(} D_k(O_i,O_j)= \big(\sum_{j=1}^d \mid O_i - O_j\mid ^2 \big)^{(1/2)} \bigg{)} $ are shown in the $(2 \times 5)$ matrices below, where $2$ is the number of clusters and $5$ is the number of objects (points).
 \begin{equation}
\label{similarity Matrix}
\nonumber
D_1 =
\begin{bmatrix}
0.0 & , & 0.0 & , & 0.0 & , & 0.0 & , & 0.0 \\
2.0 & , & 1.0 & , & 0.0 & , & 1.0 & , & 2.0\\
\end{bmatrix}
\end{equation}
\begin{equation}
\label{similarity Matrix}
\nonumber
D_2=
\begin{bmatrix}
2.0 & , & 1.0  & , & 0.0 & , & 1.0 & , & 2.0 \\
0.0 & , & 0.0  & , & 0.0 & , & 0.0 & , & 0.0\\
\end{bmatrix}
\end{equation}
A zero value in the first matrix in the first row indicates that the object is on the first line. For example, in $D_1$, the first row shows that all the members of set $\mathscr{L}_1$ are on the first line. The second row shows how far each one of the points on the line are from the second line (cluster). Likewise the matrix $D2$ shows the data points on the second line. Each point has been assigned some membership degrees with regard to crisp and fuzzy sets as shown in the matrices below. The following membership function, presented by Eq. (\ref{Func_Matrix}) and Fig. \ref{points-lines}, has been used for crisp and fuzzy methods with respect to the conditions for these methods described in the previous chapters. 
\begin{equation}
\label{Func_Matrix}
U_{ij} = \left\{
\begin{array}{l l}
1 & \quad \mbox{if $D_{p_{ij}} = 0$ }\\
\mbox{$ 1- \frac{D_{p_{ij}}}{D_{\delta}} $ } & \quad \mbox{ if $ 0 < D_{p_{ij}} \leqslant D_{\delta}$ }\\
0 & \quad \mbox{if $D_{p_{ij}} > D_{\delta}$}\\
\end{array} \right.
\end{equation}\\[0.2cm]
where $D_{p_{ij}}$ is the Euclidean distance of object $O_{j}$ from cluster $i$, and $D_{\delta}$ is a constant that the Thesis uses to normalize the values. In this example $(D_{\delta}=2)$. 
By selecting crisp membership functions in membership assignments, the origin can be a member of just one cluster and must be removed from other clusters. Given the properties of fuzzy membership functions, if the number of clusters increases, the membership degree assigned to each object will decrease proportionally. For instance, in the case of two clusters, the membership of the origin $u_{13} =1/2$, but if the number of clusters increases to $n$ using Eq. (\ref{transversal-equ}) we obtain $u_{13} = \frac{1}{n}$. Hence, a point $p_{ij}$ will have a membership value $(u_{ij} ) $ that is smaller than the degree that we can expect to get intuitively $E(u_{ij})$ {(typicality \cite{thirteen} degree for $(u_{ij})$)}. For instance $( E(u_{11}) = E(u_{12}) = 1)$ given that they are points in the line. However, as the number of clusters increases, we will obtain smaller degrees for $(u_{ij} )$ since ($ 1 < c \;\; \longmapsto \; \; \; u_{ij} < E(u_{ij}) $ , $1 \ll c  \; \; \longmapsto  \;\; \; u_{ij} \mapsto 0;$). Crisp methods just consider the origin as a member of one cluster and do not allow such an object to participate in other clusters. Given the properties of the fuzzy membership functions, if the number of clusters increases, the membership degree assigned to each object will decrease proportionally. For instance in the case of two clusters, the membership of the origin $u_{13} =1/2$, but if the number of clusters increases to $n$ using Eq. (\ref{transversal-equ}) we obtain $u_{13} = \frac{1}{n}$. \\
In possibilistic method, as the method accepts any arbitrary object with a membership value grater than zero with respect to even one cluster or class, then the implementations are not restricted and consequently obtaining different results with not reasonable accuracy. And according to this example, it is shown that possibilistic methods need more restrictions to allow the origin to participate in all lines (clusters).\\

\begin{picture}(10.3,5.0)
\linethickness{0.3mm}
\hspace*{10mm}\put(0,0){\line(1,0){400}}
\end{picture}\\
\begin{equation}
\label{Mem_Matrix_L1}
\nonumber
{ U_{crisp}(\mathscr{L}_1) =
\begin{bmatrix}
1.0 &,& 1.0 &, & {\bf 1.0} & , & 1.0 & , & 1.0 \\
0.0 &, & 0.0 & , & {\bf 0.0} & , & 0.0 & ,& 0.0\\
\end{bmatrix}}
\end{equation}
\begin{equation}
\label{Mem_Matrix_L2}
\nonumber
{ U_{crisp}(\mathscr{L}_2) =
\begin{bmatrix}
0.0 & , & 0.0 & , & {\bf 0.0} & , & 0.0 & , & 0.0 \\
1.0 & , & 1.0 & , & {\bf 0.0} & , & 1.0 & , & 1.0\\
\end{bmatrix}}
\end{equation}
\begin{center}
or
\end{center}
\begin{equation}
\label{Mem_Matrix_L1}
\nonumber
{ U_{crisp}(\mathscr{L}_1) =
\begin{bmatrix}
1.0 &,& 1.0 &, & {\bf 0.0} & , & 1.0 & , & 1.0 \\
0.0 &, & 0.0 & , & {\bf 0.0} & , & 0.0 & ,& 0.0\\
\end{bmatrix}}
\end{equation}
\begin{equation}
\label{Mem_Matrix_L2}
\nonumber
{ U_{crisp}(\mathscr{L}_2) =
\begin{bmatrix}
0.0 & , & 0.0 & , & {\bf 0.0} & , & 0.0 & , & 0.0 \\
1.0 & , & 1.0 & , & {\bf 1.0} & , & 1.0 & , & 1.0\\
\end{bmatrix}}
\end{equation}
\begin{picture}(10.3,5.0)
\hspace*{10mm}  - - - - - - - - - - - - - - - - - - - - - - - - - - - - - - - - - - - - - - - - - - - - - - - - - - - - - - - 
 \end{picture}
%
%
\begin{equation}
\label{Mem_Matrix_L1}
\nonumber
{ U_{fuzzy}(\mathscr{L}_1) =
\begin{bmatrix}
1.0 & , & 0.5 & , & {\bf 0.5} & , & 0.5 & , & 1.0 \\
0.0 & , & 0.5 & , & {\bf 0.5} & , & 0.5 & , & 0.0\\
\end{bmatrix}}
\end{equation}
\begin{equation}
\label{Mem_Matrix_L2}
\nonumber
{ U_{fuzzy}(\mathscr{L}_2) =
\begin{bmatrix}
0.0 & , & 0.5 & , & {\bf 0.5} & , & 0.5 & , & 0.0 \\
1.0 & , & 0.5 & , & {\bf 0.5} & , & 0.5 & , & 1.0\\
\end{bmatrix}}
\end{equation}
\begin{picture}(10.3,5.0)
\hspace*{10mm}  - - - - - - - - - - - - - - - - - - - - - - - - - - - - - - - - - - - - - - - - - - - - - - - - - - - - - - - 
 \end{picture}\\
%
can be accepted by PCM based on its condition $(\underset{1\leq i\leq c}{max}\; u_{ij} >0  \;\; \forall j)$:
\begin{equation}
\label{Mem_Matrix_L1}
\nonumber
 U_{pcm}(\mathscr{L}_1) =
\begin{bmatrix}
1.0 & , & 1.0 & , & {\bf 1.0} & , & 0.5 & , & 1.0 \\
0.0 & , & 0.0 & , & {\bf 0.0} & , & 0.5 & , & 0.0\\
\end{bmatrix}
\end{equation}
\begin{equation}
\label{Mem_Matrix_L2}
\nonumber
 U_{pcm}(\mathscr{L}_2) =
\begin{bmatrix}
0.0 & , & 0.5 & , & {\bf 0.0} & , & 0.0 & , & 0.0 \\
1.0 & , & 0.5 & , & {\bf 0.0} & , & 1.0 & , & 1.0\\
\end{bmatrix}
\end{equation}
\begin{picture}(10.3,5.0)
\hspace*{10mm}  - - - - - - - - - - - - - - - - - - - - - - - - - - - - - - - - - - - - - - - - - - - - - - - - - - - - - - - 
 \end{picture}\\
The issues with Fuzzy and PCM are not accepted by BFPM, which allows BFPM to obtain:
\begin{equation}
\label{Mem_Matrix_L1}
\nonumber
{ U_{bfpm}(\mathscr{L}_1) =
\begin{bmatrix}
1.0 & , & 1.0 & , & \textbf{1.0} & , & 1.0 & , & 1.0 \\
0.0 & , & 0.5 & , & \textbf{1.0} & , & 0.5 & , & 0.0\\
\end{bmatrix}}
\end{equation}
\begin{equation}
\label{Mem_Matrix_L2}
\nonumber
{ U_{bfpm}(\mathscr{L}_2) =
\begin{bmatrix}
0.0 & , & 0.5 & , & \textbf{1.0} & , & 0.5 & , & 0.0 \\
1.0 & , & 1.0 & , & \textbf{1.0} & , & 1.0 & , & 1.0\\
\end{bmatrix}}
\end{equation}\\
\begin{picture}(10.3,5.0)
\linethickness{0.3mm}
\hspace*{10mm} \put(0,0){\line(1,0){400}}
\end{picture}\\
The presented comparison was aimed to show the properties of critical objects. The results show that which learning methods can provide the proper search space for objects, specially critical objects, in order to evaluate their functionalities. In follow we look at the  $c \times n$ partition or membership matrix provided by crisp, fuzzy (probability), possibilistic, and BFPM to analyse and compare the functionality of these methods, where $c$ is the number of clusters and $n$ is the number of objects. To present the membership matrix, the objects $p_{11}, \; p_{12}, \; p_{14}, \; p_{15}$ from the first line $\mathscr{L}_1$ (known as the first cluster) and the objects $p_{21}, \; p_{22}, \; p_{24}, \; p_{25}$ from the second line $\mathscr{L}_2$ (known as the second cluster) are depicted, but the origin which was named as $p_{13}$ and $p_{23}$ is depicted just once as $p_{13}$. The following membership matrices are obtained by crisp, fuzzy, possibilistic, and BFPM methods respectively.\\

\begin{equation}
\label{Crisp1_Matrix_L1_2}
\nonumber
{ U(Crisp) =
\begin{bmatrix}
1.0 , & 1.0 , & {\bf 1.0} , & 1.0 , & 1.0 , & 0.0 , & 0.0  , & 0.0  , & 0.0 \\
0.0 , & 0.0 , & {\bf 0.0} , & 0.0 , & 0.0 , & 1.0 , & 1.0  , & 1.0  , & 1.0 \\
\end{bmatrix}}
\end{equation}
\begin{center}
or
\end{center}
\begin{equation}
\label{Mem_Matrix_L1}
\nonumber
{ U(Crisp) =
\begin{bmatrix}
1.0 , & 1.0 , & {\bf 0.0} , & 1.0 , & 1.0 , & 0.0 , & 0.0 , & 0.0 , & 0.0 \\
0.0 , & 0.0 , & {\bf 1.0} , & 0.0 , & 0.0 , & 1.0 , & 1.0 , & 1.0 , & 1.0 \\
\end{bmatrix}}
\end{equation}\\

\begin{equation}
\label{Fuzzy_Matrix_L1_2}
\nonumber
{ U(Fuzzy) =
\begin{bmatrix}
1.0  , & 0.5 , & {\bf 0.5}  , & 0.5  , & 1.0 , & 0.0  , & 0.5  , & 0.5 , & 0.0\\
0.0  , & 0.5 , & {\bf 0.5}  , & 0.5  , & 0.0 , & 1.0  , & 0.5  , & 0.5 , & 1.0\\
\end{bmatrix}}
\end{equation}\\

It can be accepted by PCM based on its condition:\\
\begin{equation}
\label{PCM_Matrix_L1_2}
\nonumber
U(PCM) =
\begin{bmatrix}
1.0 , & 1.0 , & {\bf 1.0} , & 0.5 , & 1.0 , & 0.0 , & 0.0 ,  & 0.5 , & 0.0\\
0.0 , & 0.5 , & {\bf 0.0} , & 0.0 , & 0.0 , & 1.0 , & 0.5 ,  & 1.0 , & 1.0\\
\end{bmatrix}
\end{equation}\\

\begin{equation}
\label{BFPM_Matrix_L1_2}
\nonumber
{ U(BFPM) =
\begin{bmatrix}
1.0  , &  1.0  ,  & \textbf{1.0}  , & 1.0  , & 1.0 , & 0.0  ,  & 0.5  , &  0.5  ,  & 0.0\\
0.0  , &  0.5  ,  &  \textbf{1.0}  , & 0.5  , & 0.0 , & 1.0  ,  & 1.0  , &  1.0  , & 1.0\\
\end{bmatrix}}
\end{equation}\\
Regarding the results from membership matrices from different methods and BFPM, it can be concluded that crisp method cannot allow the origin to participate in more than one cluster which cannot cover subspaces for the other lines (except one). On the other hand objects cannot get any membership from other clusters which is completely prevent the analysis of critical objects' movements in crucial systems. Moving to fuzzy methods, it is concluded that fuzzy methods are able to allow objects to participate in other clusters, but their memberships will be decreased by the number of clusters that the objects participate in. In fact, it is not possible to show the precise properties of each object. As a result, fuzzy methods cannot precisely cover the objects' movement analysis. Possibilistic methods are able to allow objects to participate in more clusters, but its condition ($\underset{1\leq i\leq c}{max}\; u_{ij} >0  \;\; \forall j$) allows the learning algorithms being trapped by assigning a membership to each object from just one cluster, which that might make the algorithm to perform improperly for the first iterations, which consequently leads to improper accuracy. In the related work in this Thesis, some of the methods that are introduced to cover the issues with PCM methods are studied. The issue with PCM makes it difficult to analyse the objects' movement from one cluster to another. BFPM makes the search space very flexible for the objects to participate in all clusters to show their potential abilities for objects' movement analysis. BFPM allows investigators to evaluate critical objects in crucial systems based on their ability to move from one cluster to another. \\
According to this example, BFPM allows the origin to participate in all clusters (lines) with full membership without decreasing its ability, and on the other side, it shows which points have the potential ability to move from one cluster to another by mutation. In conclusion, crisp membership functions are not able to assign membership degrees to objects participated in more than one cluster. This kind of membership function is just useful for some special clustering methods such as hierarchical clustering. Fuzzy membership function reduces the memberships assigned to objects with respect to each cluster. Possibilistic membership functions can obtain higher degrees, but its condition can lead to different unsatisfactory results.\\
\vspace*{-4mm}
\subsection{Example in Set Theory}
\label{Set_T_BFPM}
\begin{figure}[!h]
\begin{center}
\leavevmode\fbox{\parbox[b][6.0cm][s]{90mm}{
\vfill\footnotesize {\includegraphics[width=9cm,height=6.cm]{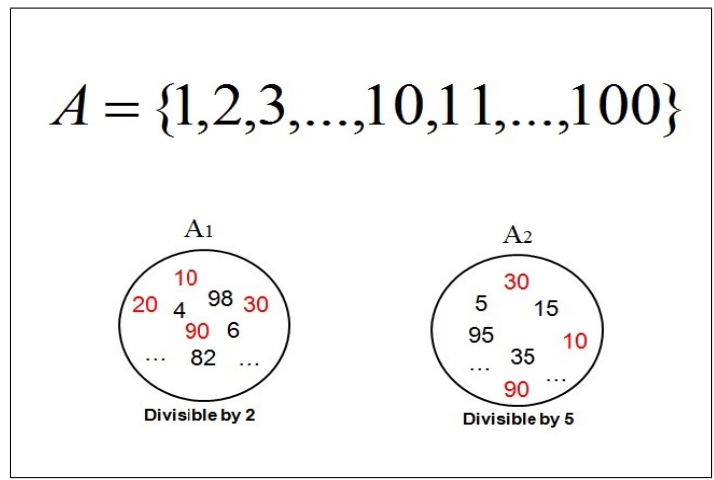}}\vfill}}
\caption{Categorizing numbers in divisible by two and divisible by five categories.}
\label{prime-mem}
\end{center}
\end{figure}

Regarding the other examples for set theory presented in the previous chapters and Fig. \ref{prime-mem}, we see that none of the discussed membership functions can provide the mathematical operation (the intersection $A_1 \cap A_2$), where $A_1$ and $A_2$ are two sets. According to the example, some members are divisible by two and five as full members of both sets such as $(A_1 \cap A_2 = \lbrace 10, 20, 30, ... \rbrace $). 

\hspace*{-7mm} Removing the objects or decreasing their memberships from any of those sets, or formally from the intersection, will lead to losing very important information. In other words, we are not allowed to consider the member $30$ as a full member of just the category "divisible by two", but removing this member from another category "divisible by five". It is not also acceptable if we consider this member as a partial member of both categories. As a result, the intersection cannot be implemented by crisps methods as crisp methods separate all data objects into different groups. The intersection operation cannot be also precisely implemented by fuzzy, probability, and possibilistic methods.\\
In conclusion and according to the issues with other methods, necessity of using a comprehensive method to cover mathematical operations in addition to provide the most suitable search space is undeniable. In other words, the proposed method should be able to accurately categorize data objects without removing or decreasing the important proprieties of each individual object.\\
\subsection{Example in Society and Economy}
\label{Soc_Eco_BFPM}
Regarding the example related to board's members, assume that we want to give a promotion or bonus to each member of each board based on their activities, the proportion of their contributions, and so on. In convenience, let's calculate bonus for three different members $\Big{(} A= \lbrace a_1,a_2,a_3 \rbrace \Big{)}$ participated in two boards $\Big{(}B= \lbrace b_1,b_2 \rbrace \Big{)}$, based on the amount of contribution presented in Table \ref{Benefit-proportion}. 
\begin{table}[!ht]
\caption{Amount of bonuses based on the amount of contributions.}
\label{Benefit-proportion}
\begin{center}
\begin{tabular}{ | c | c | c |}
\hline
{\bf Contribution} & {\bf Membership} & {\bf Benefit}  \\
\hline
\hline
15\% & 0.5 &  10K \$ \\
\hline
20 \% & 0.8 & 20k \$ \\
\hline
25 \% & 0.9 &  25K \$  \\
\hline
30 \% & 1.0 &  30K \$  \\
\hline
\end{tabular}
\end{center}
\end{table}
In Table \ref{Benefit-proportion}, membership degrees assigned to contribution are demonstrated. Table \ref{Modified Membership Degrees} explores the membership degrees obtained by $(a_1,a_2,a_3)$ with contribution presented in Table \ref{Member-proportion for Board}, based on different methods.
\begin{table}[!h]
\caption{Member-proportion to boards.}
\label{Member-proportion for Board}
\begin{center}
\begin{tabular}{ | c | c | c | c | }
\hline
{\bf Board} & {\bf Member} & {\bf Proportion}  &  {\bf Membership } \\
\hline
\hline
$b_1$ & $a_1$  & 25 \% & 0.9  \\
\hline
$b_1$ & $a_2$ &  25  \%& 0.9   \\
\hline
\hline
$b_2$	& $a_2$ & 25  \%& 0.9   \\
\hline
$b_2$ & $a_3$	& 15 \% &  0.5  \\
\hline
\end{tabular}
\end{center}
\end{table}
\begin{table*}
\caption{Modified membership degree for boards' members with respect to clustering methods.}           
\label{Modified Membership Degrees}
\begin{center}
\begin{tabular}{ | c | >{\small}c | >{\small}c ||  >{\small}c | >{\small}c ||  >{\small}c | >{\small}c |}
\hline
& \multicolumn{2}{| c |}{\bf Crisp} & \multicolumn{2}{| c |}{\bf Fuzzy/Probabilistic} & \multicolumn{2}{| c |}{\bf BFPM} \\
\cline{2-7}
{\bf Object} & {\footnotesize \bf Cluster 1} & {\footnotesize \bf Cluster 2}  & {\footnotesize \bf Cluster 1} & {\footnotesize \bf Cluster 2} & {\footnotesize \bf Cluster 1} & {\footnotesize \bf Cluster 2}\\
\hline
\hline
$a_1$	& 1 & 0 &   0.9 & 0 & 0.9 & 0 \\
\hline
$a_2$ & 1/0 & 0/1 & 0.5	& 0.5  & 0.9 & 0.9 \\
\hline
$a_3$ & 0& 1& 0  &  0.5  & 0 & 0.5\\
\hline
\end{tabular} 
\end{center}
\end{table*}
\begin{table*}[!t]
\caption{Bonus assigned to members by crisp, fuzzy, and BFPM method.}
\label{Members-bonus by others}
\begin{center}
\begin{tabular}{| c | c | >{\small}c  | >{\small}c  |  >{\small}c  |}
\hline
{\bf Member} & {\bf Board} & $\bf B_{crisp}$ & $\bf B_{fuzyy}$ & $\bf B_{bfpm}$ \\
\hline
\hline
$a_1$ & $b_1$ & 30k \$ & 25k \$  & 25k \$ \\
\hline
$a_2$ & $b_1$ &  30k \$ & 10k \$ & 25k \$\\
\hline
$a_2$ & $b_2$ &  0k \$ & 10k \$  & 25k \$ \\
\hline
$a_3$ & $b_2$ &  30k \$ & 10k \$  & 10k \$ \\
\hline
\end{tabular}
\end{center}
\end{table*}
%
%

\hspace*{-7mm} Consequently, the amount of bonuses obtained by each member with respect to clusters (classes) and their contributions are presented in Table \ref{Members-bonus by others}. This example demonstrates $a_2$ as a critical object that can participate in more than one cluster with partial memberships. According to the type of mining algorithms, $a_2$ is obtained right bonuses just by BFPM method $(50k)$. In crisp method $a_2$ can obtain a bonus from just one board $(30k)$ and in fuzzy or probability methods the amount of bonuses assigned to $a_2$ is reduced to $(20k)$. Critical objects are normal objects with special ability on participating in more than one cluster. These objects cannot be removed from datasets as like as the origin from crossing lines. In some cases, critical objects can move from one cluster to another by small changes or mutation. Mutation is very important in crucial systems in the way that the system must not lose its members. For example, according to bioinformatics \cite{hundred-twelve}, a normal cell in human body should not get mutation to move to disease or cancer category.\\
In financial systems, losing any customer specially investors is irreparable. Knowing and considering critical objects in advance will brighten the direction of mining and learning strategies. In almost all domains and disciplines, there are some critical objects that should be treated accordingly. Almost all of the methods neglect considering critical objects and they treat critical objects as normal objects which leads to low accuracy and irreparable consequences. BFPM presents a new learning methodology to treat all types of objects with respect to their properties and functionalities in addition to provide the most suitable search space to remove unnecessary restrictions to study and track the objects behaviour besides obtaining the desirable accuracy. The proposed methodology not only analyses the current status of the systems, but also evaluates the upcoming events in the near future. In other words, BFPM simultaneously offers prediction, prevention, and partitioning methodologies \cite{hundred-thirteen}. BFPM avoids the problem of reducing the objects' memberships as the number of clusters increases. BFPM allows the origin to attend all clusters (lines) with full memberships without decreasing its ability, and on the other side, it shows which points have the potential ability to move from one cluster to another by mutation \cite{hundred-fourteen}.\\
In crucial systems, addressing critical objects is very important, as investigators need to encourage or prevent objects to/from contributing to other clusters \cite{hundred-fifteen}. Critical objects are important even in multi objective optimization problems, as we are interested to find the solution that fits in all the objective functions \cite{hundred-sixteen}. In all disciplines and problems we can find such a situation that we need to allow members to obtain membership degrees from more than one cluster. In fuzzy, probability, and possibilistic methods, this requirement has been handled by offering partial memberships and in this Thesis we offer partial and also full membership degrees for more than one even all clusters. Another example can be a teacher or a student from two or more departments that needs to be assigned full memberships for more than one department. Assume we are planning to categorize a number of students or teachers (S) into some clusters (classes), based on their skills with respect to each department \textit{$\mathscr{D}$} (mathematics, physics, chemistry and so on) shown as $\{  u_{ij}(S)|S_j \in \mathscr{D}_i \; , \; u_{ij} \rightarrow [0,1]\}$. Again assume that a very good student (\textit{$S_j$}) with potential ability to participate in more than one cluster ($\mathscr{D}_i$) with the full membership degree (1):\\
\begin{center}
 	$u_{(\mathscr{D}_1)} (S_j) \; = \; 1$   ,   $u_{(\mathscr{D}_2)} (S_j) \; = \; 1$\\ , ... ,   and  $u_{(\mathscr{D}_n)} (S_j) \; = \; 1$\\
\end{center}
As the equations show, $S_j$ is a member with the full membership degree of some clusters (classes or departments) $\mathscr{D}_1, \mathscr{D}_2, ..., \mathscr{D}_n$. Following membership assignments are calculated based on crisp, fuzzy, and BFPM methods for membership values for the student $(S_1)$ who is a member of two departments $\mathscr{D}_1$ and $\mathscr{D}_2$ out of four:    
\begin{equation}
\label{Mem_Matrix_T1}
\nonumber
U_{ij}(S_j) =
\begin{bmatrix}
u_{\mathscr{D}_1} & , & u_{\mathscr{D}_2} & , & u_{\mathscr{D}_3} & , & u_{\mathscr{D}_4} &   \\
\end{bmatrix}
\end{equation}

\begin{picture}(10.3,5.0)
\linethickness{0.3mm}
\hspace*{10mm}\put(0,0){\line(1,0){400}}
\end{picture}
\begin{equation}
\label{Mem_Matrix_T1}
\nonumber
U_{Crisp}(S_1) =
\begin{bmatrix}
{\bf 1.0} & , & 0.0 & , & 0.0 & , & 0.0 &   \\
\end{bmatrix}
\end{equation}
or
\begin{equation}
\label{Mem_Matrix_T2}
\nonumber
{ U_{Crisp}(S_1) =
\begin{bmatrix}
0.0 & , & { \bf 1.0} & , & 0.0 & , & 0.0 & \\
\end{bmatrix}}
\end{equation}
\begin{center}
\hspace*{10mm} - - - - - - - - - - - - - - - - - - - - - - - - - - - - - - - - - - - - - - - - - - - - - - - - 
\end{center}
\begin{equation}
\label{Mem_Matrix_L1}
\nonumber
{ U_{fuzzy}(S_1) =
\begin{bmatrix}
{\bf 0.5} & , & {\bf 0.5} & , & 0.0 & , & 0.0 &  \\
\end{bmatrix}}
\end{equation}
\begin{center}
\hspace*{10mm} - - - - - - - - - - - - - - - - - - - - - - - - - - - - - - - - - - - - - - - - - - - - - - - - 
\end{center}
\begin{equation}
\label{Mem_Matrix_L1}
\nonumber
{ U_{pcm}(S_1) =
\begin{bmatrix}
{\bf 0.5} & , & {\bf 0.0} & , & 0.0 & , & 0.0 &  \\
\end{bmatrix}}
\end{equation}
\begin{center}
\hspace*{10mm} - - - - - - - - - - - - - - - - - - - - - - - - - - - - - - - - - - - - - - - - - - - - - - - - 
\end{center}
\begin{equation}
\label{Mem_Matrix_T2}
\nonumber
{ U_{bfpm}(S_1) =
\begin{bmatrix}
{\bf 1.0} & , & {\bf 1.0} & , & 0.0 & , & 0.0 & \\
\end{bmatrix}}
\end{equation}
\begin{picture}(10.3,5.0)
\linethickness{0.3mm}
\hspace*{10mm} \put(0,0){\line(1,0){400}}
\end{picture}\\

\hspace*{-7.5mm} As results show, students can obtain credit from just one department in crisp method. In fuzzy and probability methods, the students memberships and credits are divided into the number of departments. In possibilistic methods, it is acceptable even if students obtain credits more than zero from any arbitrary department. Only the students' credits are precisely assigned by BFPM. In all the presented examples, obtaining $(u_{ij}>0) $ even for one cluster satisfies the possibilistic condition $(\underset{1\leq i\leq c}{max}\; u_{ij} >0)$ which cannot cover the concept of critical objects. In fact, PCM with its condition sometimes behaves like as crisp methods in uncertain condition, as categorizing an arbitrary object in one cluster satisfies the condition. \\
In conclusion, it can be concluded that BFPM provides not only removes the limitations for objects in their participations in other clusters in partitioning strategies in comparison with other partitioning methods, but also it offers the methodology of analysing objects movement from one partition to another. \\
\section{Generations of BFPM}
The Thesis introduces Algorithms \ref{FPM}\textendash\ref{BFPM-WFD} for clustering purposes and Algorithm \ref{BFPCM} for classification problems in order to compare the accuracy of different methods on different search spaces. Both types of classification and clustering algorithms proposed in this Thesis make use of BFPM membership assignments. The only difference between these types is the training step which is used for classification problems. The first goal of the Thesis is to check how different membership and similarity functions affect the final outcomes. The Thesis also aims to check which algorithm can be able to cover diversity in both the feature and the vector spaces. It should be noted that the domain for memberships are defined precisely in an interval $[0,1]$, which means lower values than zero will be considered as zero and higher values than one are considered as one.\\
%
%
%
\subsection{FPM Algorithm} 
This algorithm was inspired by fuzzy and possibilistic algorithms to assign the membership value to objects. Eq. (\ref{Fuzzy CM}) and Eq. (\ref{prototype-U}) show how the algorithm calculates $(u_{ij})$, and how the centroids $(v_i)$ will be updated in each iteration. The algorithm runs until reaching the condition:
\begin{equation}
\nonumber
 \underset{1 \leq  k  \leq c }{max} { \Big{ \lbrace} || V_{k,new} - V_{k,old} ||^2 \Big{\rbrace}}  < \varepsilon 
\end{equation}
 The value assigned to $\varepsilon $ is a pre-determined constant that varies based on the type of objects and clustering problems to convince that there is no more update on centroids.
\begin{algorithmic}
\begin{algorithm}
\caption{FPM Algorithm}
\label{FPM}
\scriptsize
\textbf{Input:\; O,} c, m	\\
\textbf{Output:\; U, V} 	
\STATE  \textbf{Initialize V;}
\WHILE {$ \underset{1 \leq  k  \leq c }{max} { \lbrace || V_{k,new} - V_{k,old} ||^2 \rbrace}  > \varepsilon$ }
    \STATE
    \begin{equation}
	\label{Fuzzy CM}
		u_{ij} = [\sum_{k=1}^c (\frac{||O_j - v_i||}{||O_j - v_k||} )^{\frac{2}{m-1}}]^{\frac{1}{m}} , \;\; \forall i,j; \; \; u_{ij} \in [0,1]
	\end{equation}
	 \STATE
 	 \begin{equation}
	\label{prototype-U}
	V_i = \frac{\sum_{j=1}^n (u_{ij})^m O_j}{ \sum_{j=1}^n (u_{ij})^m} ,\; \; \forall i
	\end{equation}
\ENDWHILE
\end{algorithm}
\end{algorithmic}
$c$ is the number of clusters, $O_j$ is the $j^{th}$ object, $V_i$ is the $i^{th}$ centroid or vector center, and $m$ is the fuzzification constant. As the basic FPM algorithm assigns $(u_{ij})$ based on the total distance, Eq. (\ref{Euclidean}), the Thesis implements Algorithms \ref{BFPCM-I} and \ref{BFPCM-II} not only to compare the objects based on their similarity using the distance function, but also to check the similarity between features of objects and similar features of centroids individually.
\begin{equation} 
\nonumber
D_E = \sqrt{ |O_j - O_l|^2} 
\end{equation}
\begin{equation} 
\label{Euclidean} 
 = \sqrt{ (x_{j1} - x_{l1})^2 + (x_{j2} - x_{l2})^2  + ... + (x_{jd} - x_{ld})^2} 
\end{equation}\\

\hspace*{-7mm} $d$ is the number of dimentions, $O_j$ and $O_l$ are two different objects in $d$ dimensional search space. $D_E$ is the Euclidean distance as it works based on the total distance between features of data objects.\\
%
%
\subsection{FPM-I Algorithm}
 The FPM-I, presented by Algorithm \ref{BFPCM-I}, works in three different steps. 
 \begin{algorithmic}
\begin{algorithm}
\caption{FPM-I Algorithm}
\label{BFPCM-I}
\scriptsize
\textbf{Input:\; O,} c, m	\\
\textbf{Output:\; U, V} 	
\STATE  \textbf{Initialize V;}
\WHILE {$ \underset{1 \leq  k  \leq c }{max} { \lbrace || V_{k,new} - V_{k,old} ||^2 \rbrace}  > \varepsilon$ }
    \STATE
    \begin{equation}
	\label{U1}
		u_{ij}^{'} = [\sum_{k=1}^c (\frac{||O_j - v_i||}{||O_j - v_k||} )^{\frac{2}{m-1}}]^{\frac{1}{m}} , \;\; \forall i,j; \; \; u^{'}_{ij} \in [0,1]
	\end{equation}
    \STATE
    \vspace*{-2mm}
    \begin{equation}
    \nonumber
		Compute (u_{ij}^{''}) ;
	\end{equation}
	\STATE
	    \vspace*{-3mm}
    \begin{equation}
	\label{U-max}
		u_{ij} = max(u_{ij}^{'}, u_{ij}^{''}), \;\; \forall i,j
	\end{equation}
 	 \STATE
 	     \vspace*{-2mm}
 	 \begin{equation}
 \nonumber
	V_i = \frac{\sum_{j=1}^n (u_{ij})^m O_j}{ \sum_{j=1}^n (u_{ij})^m} ,\; \; \forall i
	\end{equation}
\ENDWHILE
\end{algorithm}
\end{algorithmic}
\begin{itemize}
\item Step one: \\
 The first step uses the FPM to assign the membership degree $(u_{ij}^{'})$ to each object, as shown by Eq. (\ref{U1}). As mentioned earlier, the first step makes use of the total distances between objects and centroids.
\item  Step two: \\
 The second step assigns the fuzzy membership $(u_{ij}^{''})$ degree as a weight to each feature of each object, which consequently allows us to assign another membership degree to each object with respect to all of the features' distances. In order to assign a weight to each feature, the algorithm uses the distance between each object's feature and the same feature from each centroid. By considering this distance, each feature obtains a weight from zero to one. This weight is assigned to the object using Eq. (\ref{Weight-Assignment}).\\
\begin{equation}
\label{Weight-Assignment}
W_{if} = 1-(|V_{if} - O_{jf}|)
\end{equation}
where $V_{if}$ is the value for the $f^{th}$ feature of the $i^{th}$ cluster's centroid, and $O_{jf}$ is the value of the $f^{th}$ feature of the $j^{th}$ object. The negative weight is considered to be zero. To assign the final membership degree to each object, the algorithm checks the number of dominant features in each object with respect to each centroid. Based on this number, the algorithm will decide whether or not we need to update $(u_{ij}^{'})$. If the FPM step appropriately assigns the membership degree $(u_{ij}^{'})$ to an object, then the second step will not get involved. To check the accuracy of the first step assignments, the algorithm considers the number of features that their assigned weight is greater than $0.5$. If more than half of the features obtained the weights greater than $0.5$ and the first step did not consider this particular object as a member of this cluster or centroid, then the second step will assign a new membership degree $(u_{ij}^{''})$ to that particular object. Finally the new membership degree will be generated for the object based on the average value of weights. In fact, the second step is designed to cover the whole search space. The number of dominant features that affect the result are considered to assign the membership degree to each object.
\item Step three: \\
In the third step, the final membership degree for each object obtained from the two previous steps is extracted. The maximum value of membership degrees $ \Big{(} u_{ij} = max(u_{ij}^{'},u_{ij}^{''}) \Big{)}$ from the previous steps is considered as the final membership degree $(u_{ij})$ for the object $j$ in order to update the centroid for the next iteration, as shown by Eq. (\ref{U-max}).
\end{itemize}
In brief, the first step involves assigning the membership degree to each object based on the distance between the object and the central location of each cluster, while the second step considers the distances between the features of each object and the features of each centroid to assign the membership degree to an object. The algorithm updates $u_{ij}$ and $V_i$ for several iterations to reach the condition $ \underset{1 \leq  k  \leq c}{max} { \Big{\lbrace} || V_{k,new} - V_{k,old} ||^2 \Big{\rbrace}}  < \varepsilon $.\\

%
%
\subsection{FPM-II Algorithm}
The Thesis introduces FPM-II, shown by Algorithm \ref{BFPCM-II}, to minimize the computational complexity and thus to reduce the cost of calculating weights and assigning the membership degrees to each feature individually. \\
%
\begin{algorithmic}
\begin{algorithm}
\caption{FPM-II Algorithm}
\label{BFPCM-II}
\scriptsize
\textbf{Input:\; O,} c, m	\\
\textbf{Output:\; U, V} 	
\STATE  \textbf{Initialize V;}
\WHILE {$ \underset{1 \leq  k  \leq c }{max} { \lbrace || V_{k,new} - V_{k,old} ||^2 \rbrace}  > \varepsilon$ }
    \vspace*{-3mm}
    \STATE
    \begin{equation}
    \nonumber
		u_{ij}^{'} = [\sum_{k=1}^c (\frac{||O_j - v_i||}{||O_j - v_k||} )^{\frac{2}{m-1}}]^{\frac{1}{m}} , \;\; \forall i,j; \; \; u^{'}_{ij} \in [0,1]
	\end{equation}
 	 \STATE
 	     \vspace*{-3mm}
 	 \begin{equation}
 	 \nonumber
	V_i = \frac{\sum_{j=1}^n (u_{ij}^{'})^m O_j}{ \sum_{j=1}^n (u_{ij}^{'})^m} ,\; \; \forall i
	\end{equation}
\ENDWHILE
\STATE
    \vspace*{-3mm}
\begin{equation}
    \nonumber
		Compute (u_{ij}^{''}) ;
	\end{equation}
\STATE	
\begin{equation}
    \nonumber
		u_{ij} = max(u_{ij}^{'},u_{ij}^{''}) , \;\; \forall i,j
	\end{equation}
\end{algorithm}
\end{algorithmic}

\hspace*{-7.5mm} The FPM-II algorithm (Algorithm \ref{BFPCM-II}) assigns the membership degree $(u_{ij}^{'})$ to each object without being affected by the second step of the membership assignment. This procedure ends upon reaching the condition $ \underset{1 \leq  k  \leq c }{max} {\Big{\lbrace} || V_{k,new} - V_{k,old} ||^{2} \Big{\rbrace}} < \varepsilon $. Then, in the second step, the algorithm assigns membership degrees $(u_{ij}^{''})$ to each object based on its features with respect to each individual cluster (centroid). Finally, the third step extracts the maximum membership degree from the two previous steps as the final membership degree of objects $\Big{(} u_{ij} = max(u_{ij}^{'},u_{ij}^{''}) \Big{)}$. Algorithms \ref{BFPCM-I} and \ref{BFPCM-II} proved that the proposed algorithms have some drawbacks in membership assignments that in each iteration this error misleads the algorithms for the next iterations, which consequently results in improper accuracy. To avoid such an improper membership assignments, the new BFPM methodology has been applied on previous algorithms. In other words, BFPM has been applied on data objects by providing the most flexible search space to allow objects to participate in as much clusters as they can to show their potential abilities in moving from one cluster to another. The following algorithm inspired by the achievements from the previous algorithms (Algorithms \ref{BFPCM-I} and \ref{BFPCM-II}).\\

%
%
\subsection{BFPM Algorithm}
BFPM is implemented to prove that the method is able to provide a flexible search space for data objects besides clustering them as accurate as possible \cite{hundred-fifteen, hundred-sixteen, hundred-seventeen}. The procedure of this algorithm is also similar to the previous algorithms without checking the miss-assignments in each iteration. This algorithm makes use of the proposed BFPM method in membership assignments to allow objects to participate in more clusters by providing the facility to analyse the objects' movement analysis. The algorithm makes also use of the Euclidean distance function in its similarity measurements, but the procedure of membership assignments has been handled by BFPM.  
\vspace*{-3mm}
\begin{algorithmic}
\begin{algorithm}
\caption{BFPM Algorithm}
\label{BFPM}
\scriptsize
\textbf{Input:\; O,} c, m	\\
\textbf{Output:\; U, V} 	
\STATE  \textbf{Initialize V;}
\WHILE {$ \underset{1 \leq  k  \leq c }{max} { \lbrace || V_{k,new} - V_{k,old} ||^2 \rbrace}  > \varepsilon$ }
    \STATE
        \vspace*{-4mm}
    \begin{equation}
	\label{BFPM-u}
		u_{ij} = \Big{[}\sum_{k=1}^c \big{(}\frac{||O_j - v_i||}{||O_j - v_k||} \big{)}^{\frac{2}{m-1}}\Big{]}^{\frac{1}{m}} , \;\; \forall i,j; \; \; u_{ij} \in [0,1]
	\end{equation}
	 \STATE
	     \vspace*{-2mm}
 	 \begin{equation}
	\label{BFPM-P}
	V_i = \frac{\sum_{j=1}^n (u_{ij})^m O_j}{ \sum_{j=1}^n (u_{ij})^m} ,\; \; \forall i \; ; \; \; \; \; \;  { (0 < \frac{1}{c} \sum_{i=1}^c u_{ij} \leq 1)}.
	\end{equation}
\ENDWHILE
\end{algorithm}
\end{algorithmic}
Eq. (\ref{BFPM-u}) and Eq. (\ref{BFPM-P}) show how the algorithm calculates $(u_{ij})$ and how the centroids $(v_i)$ will be updated in each iteration. The algorithm runs until reaching the condition:
\begin{equation}
\nonumber
 max_{1 \leq  k  \leq c } { \lbrace || V_{k,new} - V_{k,old} ||^2 \rbrace}  < \varepsilon
 \end{equation}
 The value assigned to $\varepsilon $ is a predetermined constant that varies based on the type of objects and clustering problems. U is the $(c \times n)$ partition matrix, $ V= \lbrace v_1, v_2,..., v_c \rbrace $ is the vector of $c$ cluster centers in $ \Re^d , \; m  $ is the fuzzification constant, and $ ||.||_A $ is any inner product A-induced norm, and Euclidean distance function presented by Eq. (\ref{Euclidean}). So by using Euclidean distance function, diversity in both the vector and the feature spaces has not been covered by this algorithm. In fact, some other parameters specially in the feature spaces that might affect the accuracy of the algorithm have not been considered by this algorithm, which encouraged to apply the new similarity functions in learning procedure. The concept of diversity and the issues with most of the similarity functions encouraged to make use of Weighted Feature Distance (WFD) functions in the following algorithm with the aim of covering diversity besides handling the impact of dominant features on the final results. 
%
%
\vspace{-2mm}
\subsection{BFPM-WFD}
\label{BFPM-W}
Again, as BFPM algorithm assigns $(u_{ij})$ based only on the total distance shown by Eq. (\ref{Euclidean}), the algorithm BFPM-WFD (BFPM Weighted Feature Distance) is implemented to allow each individual feature to have its own affect on distance functions by overcoming the drawbacks of dominant features on the final result. In some of the distance functions, the result is being affected by dominant features \cite{hundred-thirteen}. Weighted Feature Distance (WFD) for $L_2$ norm is being described as follow, where $d$ is the number of variables or dimensions for numerical data objects. $w_k$ and $w_k^{'}$ are the weights assigned to features of the first and the second objects respectively. The Thesis introduces $(w_k =w_k^{'})$ if both objects are in the same scale.
\begin{equation}
\label{WFD $L_2$}
 WFD_{L_2} \; = \; \sqrt{\big{(} W_j O_j - W_l O_l \big{)} ^2} =  \Big{(} \sum_{k=1}^d {\big{(}}|w_k x_{jk} - w_k^{'} x_{lk}|^2 {\big{)}} \Big{)} ^{(\frac{1}{2})}
\end{equation}
The memberships and centroids are updated in each iteration based on Eq. (\ref{Fuzzy C-Means membership update}) and Eq. (\ref{prototype-update}). 
\begin{algorithmic}
\begin{algorithm}
\caption{BFPM-WFD}
\label{BFPM-WFD}
\scriptsize
\textbf{Input:\; O,} c, m	\\
\textbf{Output:\; U, V} 	
\STATE  \textbf{Initialize V;}
\WHILE {$ \underset{1 \leq  k  \leq c }{max} { \lbrace || V_{k,new} - V_{k,old} ||^2 \rbrace}  > \varepsilon$ }
    \STATE
    \begin{equation}
    \nonumber
		 \Big{\lbrace} u_{ij} = \Big{[}\sum_{k=1}^c \big{(}\frac{||O_j - v_i||}{||O_j - v_k||} \big{)}^{\frac{2}{m-1}}\Big{]}^{\frac{1}{m}} , \;\; \forall i,j \; ; \;\; u_{ij} \in [0,1]
	\end{equation}
	\begin{equation}
	\label{Fuzzy C-Means membership update}
		||O_j - O_l|| \; =  \Big{(} \sum_{f=1}^d {\big{(}}|w_f.x_{jf} - w_f^{'}. x_{lf}|^2 {\big{)}} \Big{)} ^{(\frac{1}{2})} \Big{\rbrace}
	\end{equation}
	 \STATE
 	 \begin{equation}
	\label{prototype-update}
	V_i = \frac{\sum_{j=1}^n (u_{ij})^m O_j}{ \sum_{j=1}^n (u_{ij})^m} ,\; \; \forall i \; ; \; \; \; \; \;  { (0 < \frac{1}{c} \sum_{i=1}^c u_{ij} \leq 1)}.
	\end{equation}
\ENDWHILE
\end{algorithm}
\end{algorithmic}
\subsection{BFPCM Algorithm}
\label{BFPCM}
The algorithm performs on selecting the training and the testing datasets using random sampling, in which those sets can be chosen using different techniques such as \textit{holdout}, \textit{random sampling}, \textit{k-fold cross validation}, and \textit{bootstrap}. The mentioned techniques for selecting the test and train datasets will be discussed in more details in section \ref{Evaluating the classifier}. The main idea of the algorithm is to cover both the feature and the vector spaces by assigning membership degrees and weights to features and objects during the learning procedures, presented by Algorithm \ref{BFPCM} \cite{hundred-seventeen}. Eq. (\ref{object-U}) shows how $u_{ij}^{'}$ will be generated for objects in the test dataset with respect to classes based on the $L_2$ norm of distance function. The algorithm assigns a membership degree $u_{ij}^{''}$ to each object in the testing dataset based on Eq. (\ref{feature-w}). This assignment assigns a weight to each feature of each object with regard to the distance between the features of the object and the feature of other objects in training dataset. Consequently, the final membership matrix for objects will be calculated using Eq. (\ref{Mem-U}).   
\begin{algorithmic}
\begin{algorithm}
\caption{BFPCM Algorithm}
\label{BFPCM}
\scriptsize
\textbf{Input:\; O,} c,	\\
\textbf{Output:\; U} 	
\FOR { $(\forall \; O_j$ in $DA_T)$ }
    \STATE
        \vspace*{-2mm}
 	 \begin{equation}
	\label{object-U}
	u_{ij}^{'} = \min_{1\leq i\leq c}\Big{[}\sum_{k=1}^d (||O_j - O_l||^{2})\Big{]}, \;\; \forall i, \; \forall \; j \in  DA_T, \;  \forall \; l\in DA_{Train}; \; \; u^{'}_{ij} \in [0,1]
	\end{equation}
    \vspace*{-2mm}
	 \STATE
    \begin{equation}
	\label{feature-w}
		u_{ij}^{''} =  \min_{1\leq i\leq c}  \Big{[}\frac{\sum_{k=1}^d ({||O_{jd} - O_{ld} ||^{2}* w_d})}{d} \Big{]} , \;\; \forall i, \; \forall \; j \in  DA_T, \;  \forall \; l\in DA_{Train}; \; \; u^{''}_{ij} \in [0,1]
	\end{equation}
	 	 \begin{equation}
	\label{Mem-U}
	u_{ij} = \frac{u_{ij}^{'} + u_{ij}^{''}}{2} , \;\; \forall i, j; \; \; u_{ij} \in [0,1]
	\end{equation}
\ENDFOR
\end{algorithm}
\end{algorithmic}
where $c$ is the number of classes or class labels, $DA_T$ is the testing dataset, $DA_{Train}$ is the training dataset, $U$ is the membership matrix, $u_{ij}$ is the membership degrees for the $j^{th}$ object for the $i^{th}$ class, and $||.||$ is the $L_2$ norm distance function.  
%
%
\chapter{Accuracy Measures for Evaluating BFPM}
\label{Measure-Chap}
In this chapter, the Thesis aims to study some statistical measurements that help us check and evaluate the accuracy of the proposed and discussed methods \cite{hundred-eighteen}, \cite{hundred-nineteen}. Different methods work well for specific datasets, where their results can be evaluated using different measurements \cite{hundred-twenty}. The accuracy of supervised methods can be mostly measured based on the percentage of the correct labelled objects, but in the following sections the Thesis explores other techniques to evaluate the accuracy in classification problems. As mentioned, clustering is a form of unsupervised learning techniques, where we cannot observe the real number of clusters, but it is common to use the notation of \textit{accuracy} (applicable to supervised learning) with \textit{distance} \cite{hundred-nineteen}. Validity indices are proposed to check the distance (compactness and separation) between data objects and the centroids. It is also very common to run unsupervised methods on labelled class datasets which leads to precise evaluation of the results of unsupervised methods. Running unsupervised methods on labelled class datasets presents how the proposed methods can be reliable in the situations that we have no information in advance. The Thesis first studies about some clusters' validity indices and evaluating measurements on classifiers, and then the Thesis discusses some statistical measurements. Using benchmark datasets for classification problems for unsupervised methods is very common by researchers, as labelled objects can lighten the functionality of their methods. In this Thesis also some benchmark datasets from classification problems have been chosen for evaluating the accuracy of clustering methods proposed in this Thesis.\\
%
%
\section{Validity Indices}
\label{Validity indices}
Validity indices are like fitness functions to evaluate the quality of the obtained clusters \cite{hundred-twenty-one}. On the other hand, these indices may help to find the optimal number of clusters whenever the number of clusters is unknown \cite{hundred-twenty-two}. Most of the validity indices are data dependent, so selecting  different validity indices may affect the accuracy of the clustering method. Based on this reason, running different validity indices on clustering methods seems necessary to evaluate the accuracy of the presented methods. There are two ways to check the performance of clustering methods. One is based on using the cluster validity functions and the other is the comparison indices methods $(s(U,V))$. There are number of functions to calculate the clustering performances, but the Thesis mentions just some of them and the Thesis uses them to evaluate the accuracy of the proposed method (BFPM). It should be noted that some of the validity indices discussed in this Thesis are introduced for crisp clustering methods. In order to use them for the proposed method, the Thesis first hardens the assigned values generated by BFPM algorithms.
%
%
\begin{itemize}
\item {\bf Partition Coefficient}\\
Partition coefficient index attempts to evaluate the accuracy of clustering methods by using Eq. (\ref{VPC(u)}) \cite{hundred-twenty-three}. The maximum value for this index shows the better performance for the clustering method.
\begin{equation}
\label{VPC(u)}
 V_{pc}(U) = \frac{\sum_{j=1}^n \sum_{i=1}^c(u_{ij}^2)}{n}
\end{equation}
$u_{ij}$ is the membership degree assigned to the $j^{th}$ object with respect to the $i^{th}$ cluster and $n$ is the number of data objects.
%
%
\item {\bf Partition Entropy}\\
This index evaluates the performance of the clustering method using Eq. (\ref{VPE(u)}). The minimum value for this index shows the better performance for the clustering method \cite{hundred-twenty-three}.
\begin{equation}
\label{VPE(u)}
 V_{pe}(U) =\frac{-1}{n} \bigg\lbrace \sum_{j=1}^n \sum_{i=1}^c(u_{ij} \;  log \;u_{ij})\bigg\rbrace
\end{equation}
$u_{ij}$ is the membership degree assigned to the $j^{th}$ object with respect to the $i^{th}$ cluster and $n$ is the number of data objects.
%
%
%
\item {\bf Xie-Beni Function}\\
Xie-Beni attempts to minimize the value of the validity function shown by Eq. (\ref{Xie Fun}) \cite{hundred-twenty-three}.
\begin{equation}
\label{Xie Fun}
V_{xb}(U,v_1,L,V_c;O) = \frac{\sum_{i=1}^c \sum_{j=1}^n u_{ij}^2 \bigg(||O_j-v_i ||^2 \bigg)}{ n * \bigg( min_{i\neq k } || v_i - v_k||^2\bigg)}
\end{equation}
$O$ is the data object, $v_i$ is the $ i^{th}$ centroid, and $u_{ij}$ is the membership degree.
%
%
%
\item {\bf DB Function}\\
DB index evaluates the performance of the clustering method by maximizing the distance between centroids distances in one side and minimizing the distance between the centroid and the objects belong to that cluster, shown by Eq. (\ref{DB Index}) \cite{hundred-twenty-four}.
\begin{equation}
\label{DB Index}
R_k = max_{k,k\neq i} \left( \frac{ e_i + e_k}{D_{ik}} \right)
\end{equation}
$D_{ik} $ is the distance between the $i^{th}$ and the $k^{th}$ centroids, $e_i$, and $e_k$ are the average errors for clusters $C_i$ and $C_k$. \\
\begin{equation}
e_i = \frac{1}{N_i} \sum_{x\in C_i} || O-P_i||^2.
\end{equation}
$P_i$ is the $i^{th}$ centroid (or prototype). By using the above equations, Eq. (\ref{DB Index}) can be rewritten as follow :
\begin{equation}
DB(K) = \frac{1}{K} \sum_{i=1}^K R_i
\end{equation}
The minimum value of DB index shows the better performance of the clustering method.
%
\item {\bf CS Index}\\
This index has the same concept as DB index, but it deals with the clusters with different densities, shown by Eq. (\ref{CS_Index}) \cite{hundred-twenty-five}.
\begin{equation}
\label{CS_Index}
CS(K) = \frac{\sum_{i=1}^K \bigg(\frac{1}{N_i} \sum_{O_j \in C_i} (max_{O_l \in C_i} D(O_l,O_j))\bigg)}{\sum_{i=1}^K \bigg( max_{k\in K, k \neq i} D(P_i,P_k) \bigg)}
\end{equation}
The minimum value for CS(K) Index shows the better performance.
%
%
%
\item {\bf G and $I_{G}$ Fuzzy Validity Measure}\\
The compactness and separation fuzzy validity function is shown by Eq. (\ref{G}) \cite{hundred-twenty-six}.
\begin{equation}
\label{G}
G = \frac{DS_s}{CP}
\end{equation}
where :
\begin{equation}
DS_s = \frac{1}{n^2} \sum_{j_1=1}^n \sum_{j_2=1}^n D^2 \big(O_{j_1},O_{j_2} \big)w_2
\end{equation}
$w_2 = min \lbrace max_{i_1} u_{i_1j_1} , max_{i_2 \neq i_1} u_{i_2j_2} \rbrace $ and $D$ is the distance between two objects. $DS_s$ is called the separation of the fuzzy c-partition.
\begin{equation}
CP= \frac{2}{n(n-1)} \sum_{j_1=1}^{n-1} \sum_{j_2=j_1+1}^n \sum_{i=1}^c D^2 \big(O_{j_1},O_{j_2}\big) w_1
\end{equation}
$w_1 = min \lbrace u_{ij_1},u_{ij_2} \rbrace$ and $n$ is the number of data objects which should be clustered in $c$ clusters. $CP$ is called the compactness of the fuzzy c-partition. According to the compactness and separation, the larger value of $G$ means the clusters are clustered more compacted and separated in c-clusters.\\
$I_G $ validity index is defined as Eq. (\ref{I_G}).
\begin{equation}
\label{I_G}
I_G = \frac{G}{\hslash(Y)}
\end{equation}
where $\hslash$ is any function such as $\hslash : Y \longmapsto Y^{y}$, where $y = \lbrace 0.5,1,2,... \rbrace$. 
\end{itemize}
%
%
\section{Evaluating the Accuracy of a Classifier or Predictor}
We need to know how accurate the model is. In order to evaluate the accuracy of the classifier or predictor, error measures and accuracy measures are used. Error or accuracy is mostly measured by calculating the  percentage of the correct labelled objects in the test dataset. Generally speaking, the accuracy and error rates can be calculated by achieving one of them from the other as: $ Err(M) = 1 - Acc(M)$, where $Err(M)$ is the error ratio and $Acc(M)$ is the accuracy ratio of the classifier $M$. The other way of calculating these measures is using the confusion matrix. The matrix contains the percentages or numbers of objects from the test dataset that are classified in true positive, true negative, false positive and false negative shown by Table \ref{Confusion_Matrix}. 
%
\begin{table}[!ht]
\caption{Confusion matrix for positive and negative objects.}
\label{Confusion_Matrix}
\begin{center}
\begin{tabular}{ | c | c | c | c |}
\cline{1-4}
\multicolumn{2}{ | c |}{} &\multicolumn{2}{  c |}{\multirow{2}{*}{Predicted Class}}\\
\multicolumn{2}{ | c |}{} &\multicolumn{2}{  c |}{}\\
\cline {3-4}
\multicolumn{2}{ | c |}{}  & \multirow{2}{*}{$C_1$} & \multirow{2}{*}{$C_2$}\\
\multicolumn{2}{ | c |}{}  & & \\
\cline{1-4}
\multirow{2}{*}{Actual} & \multirow{2}{*}{$C_1$} & \multirow{2}{*}{true positives} &  \multirow{2}{*}{false negatives} \\
 & & & \\
\cline{2-4}
Class & \multirow{2}{*}{$C_2$} & \multirow{2}{*}{false positives} & \multirow{2}{*}{true negatives} \\
 & & & \\
\cline{1-4}
\end{tabular}
\\[0.2cm]
\end{center}
\end{table}
True positives and true negatives are those objects that are correctly labelled as positive and negative for the specific classes and false positives and false negatives objects are incorrectly labelled. From the matrix, we can get the \textit{sensitivity} and \textit{specificity}, where tells us how classifier can recognize the positive objects and how well it recognizes the negative objects respectively.  \\
\begin{equation}
sensitivity = \frac{t_{pos}}{pos}
\end{equation}
$t_{pos}$ is the number of true positives that were correctly labelled and $pos$ is the number of positives.\\[0.2cm]
%
\begin{equation}
specificity = \frac{t_{neg}}{neg}
\end{equation}
$t_{neg}$ is the number of true negatives that were correctly labelled and $neg$ is the number of negatives.\\[0.2cm]
\begin{equation}
precision = \frac{t_{pos}}{(t_{pos} + f_{pos})}
\end{equation}
$f_{pos}$ is the number of false positives.\\[0.2cm]
According to the confusion matrix two types of accuracy can be obtained into two ways: one using sensitivity and specificity as:
\begin{equation}
sensitivity \frac{pos}{(pos + neg)} + specificity \frac{neg}{(pos + neg)}.
\end{equation}
and also using the summation of correctly selected objects from the true positive and the true negative as $Acc= t_{pos} + t_{neg}$. Error ratio can be also calculated by measuring the differences between the predicted value $y_j^{'}$ and the exact value $y_j$.\\
\begin{equation}
Absolute \; error : \; \; \; \; |y_j - y_j^{'}|
\end{equation}
\begin{equation}
Squared \; error : \; \; \; \; (y_j - y_j^{'})^2
\end{equation}
\begin{equation}
Mean \; absolute \; error : \; \; \; \;  \frac{\sum_{j=1}^d |y_j - y_j^{'}|}{d}
\end{equation}\\
\begin{equation}
Mean \; squared \; error : \; \; \; \;  \frac{\sum_{j=1}^d (y_j - y_j^{'})^2}{d}
\end{equation}\\
\begin{equation}
Relative \; absolute \; error : \; \; \; \;  \frac{\sum_{j=1}^d |y_j - y_j^{'}|}{\sum_{j=1}^d |y_j - \bar{ y_j}|}
\end{equation}\\
\begin{equation}
Relative \; squared \; error : \; \; \; \;  \frac{\sum_{j=1}^d (y_j - y_j^{'})^2}{\sum_{j=1}^d (y_j - \bar{y_j})^2}
\end{equation}\\[0.1cm]
where $\bar{y_j}$ is the mean of the objects in the training dataset shown as: \\ [0.2cm] 
\hspace*{7cm}$ \bar{y_j} = \frac{\sum_{j=1}^t (y_j)}{d}$
\section{Classifier Evaluation}
\label{Evaluating the classifier}
In general, selecting different training datasets might lead to different results, therefore, the selection of training and testing datasets should be precise for measuring the accuracy related to each comparison. There are several techniques to select or divide datasets into the training and the testing datasets in order to evaluate the accuracy, which some of the well-known techniques can be named as the holdout, random subsampling, k-fold cross-validation, and bootstrap methods.
\begin{itemize}
\item {\bf Holdout}\\
In this method the given dataset in randomly selected or divided into two independent sets, the training and the testing datasets. Mostly, two-thirds of the dataset are selected for the training dataset and the rest for the testing dataset. 
\item {\bf Random Subsampling}\\
This strategy is similar to holdout method, in which the same strategy as the holdout method is repeated for $t$ times as iterations. The average accuracy from iterations can be considered as the final result. This method is more reliable, because in the holdout method only a portion of the initial data is used to derive
the model.
\item {\bf K-Fold Cross Validation}\\
In k-fold cross-validation, the datasets are being randomly selected into k separated subsets (mutually exclusive) in almost similar size as $DA= \lbrace DA_1, DA_2, ..., DA_k \rbrace $. The algorithm will be run for $t$ number of iterations to select the training and the testing datasets. In each iteration, one of the partitions from $DA$ is being selected as the testing dataset ($DA_{Test}$) and the rest of the data is collectively considered as the training dataset. The accuracy is calculated on the overall number of correct classified objects from the k iterations divided by the total number of objects in the initial data. For the error, the total loss from the k iterations divided by the total number of the initial objects is considered.
\item {\bf Bootstrap}\\
Unlike other methods, this method allows the selecting algorithm to choose samples uniformly with replacement. That means one data object might be selected in the train dataset more than once. 
\end{itemize}
%
\vspace*{-5mm}
\chapter{Experimental Verification}
\label{Experiment-Chap}
\vspace*{-9mm}
The proposed method has been verified in different steps. The Thesis made use of the proposed method and the proposed similarity functions in both supervised and unsupervised forms. Some datasets related to real problems from different disciplines in medicine (diagnoses diseases for lung cancer), risk management (banking loan), and security (anomaly detection) have been used in experimental verifications. The dataset for lung cancer has been received from Harvard University, which were precisely collected for many years from healthy individuals and people diagnosed as cancer patients. The dataset for risk management and anomaly detection have been received from an international bank, which were confidentially selected  for many years and used by the proposed methodology and algorithms. The BPFM methodology and algorithms have been applied on these datasets to analyse the functionality of the proposed methods in addition to study objects' movements in real problems. The Thesis also makes use of the well-known benchmark datasets from UCI repository in different domains such as medicine, life, images, videos, and physical, presented by Table \ref{datasets}. The idea is to have different datasets from different domains with different features to check all different strategies on membership and similarity functions.\\
Datasets are available for researchers and provided by the University of California \cite{hundred-twenty-seven}. Objects' movements analysis and critical objects' analysis have been evaluated in this Thesis. Thus, the novelty of the proposed idea led the Thesis to providing some measurements to check and evaluate the functionality of the proposed method and functions. In conclusion, variety of datasets from different domains have been selected in the evaluation stage. The main goal of this chapter is to demonstrate new achievements that have been covered by the method and the functions proposed by this Thesis. The software was implemented using Java to provide the needed infrastructure for the experimental verifications.
%
\begin{table}[!ht]
\caption{Multi dimensional datasets.}
\vspace*{-5mm}
\label{datasets}
\begin{center}
\begin{tabular}{ | c | c | c | c | c | c |}
\hline
 {\bf \small Dataset} & {\bf \small  Features} & {\bf \small  Objects} & {\bf \small Clusters}  &{\bf  \small Area} & {\bf \small Data type}\\
 \hline
\hline
{\bf { \footnotesize Lung Cancer (Harvard University)}}  & 71   & 231  & 2  & Medicine & {\footnotesize Multivariate}\\
\hline
{\bf { \footnotesize Risk (International Bank)}}  & 14   & 360  & 2  & Finance & {\footnotesize Multivariate}\\
\hline
{\bf { \footnotesize Iris}}  & 4   & 150  & 3  & Life & {\footnotesize Multivariate}\\
\hline
{\bf { \footnotesize Pima Indians}} &  8  & 768  & 2  & {\footnotesize Life/Medicine}  & { \footnotesize Multivariate} \\
\hline
{\bf { \footnotesize Yeast }}		& 8 & 1299  & 4 & {\footnotesize Life/Medicine} & { \footnotesize Multivariate} \\
\hline
{\bf { \footnotesize MAGIC	}}	& 11 &  19200 &  2 & {\footnotesize Physical} & { \footnotesize Multivariate} \\
\hline
{\bf { \footnotesize Dermatology}}		& 34 &  358 &  6 & {\footnotesize Medicine} & { \footnotesize Multivariate} \\
\hline
\multirow{2}{*}{\bf {\footnotesize Libras}}		& \multirow{2}{*}{90} &  \multirow{2}{*}{360} &  \multirow{2}{*}{15} & {\footnotesize Image} & { \footnotesize Multivariate},\\
 & & & & {\footnotesize Video} & {\footnotesize sequential} \\
\hline
\end{tabular}
\end{center}
\end{table}
%
\section{Critical Objects and Object's Movement Analysis (Mutation)}
\begin{figure}[!h]
\begin{center}
\leavevmode\fbox{\parbox[b][6cm][s]{160mm}{
\vfill\footnotesize {\includegraphics[width=16cm,height=6cm]{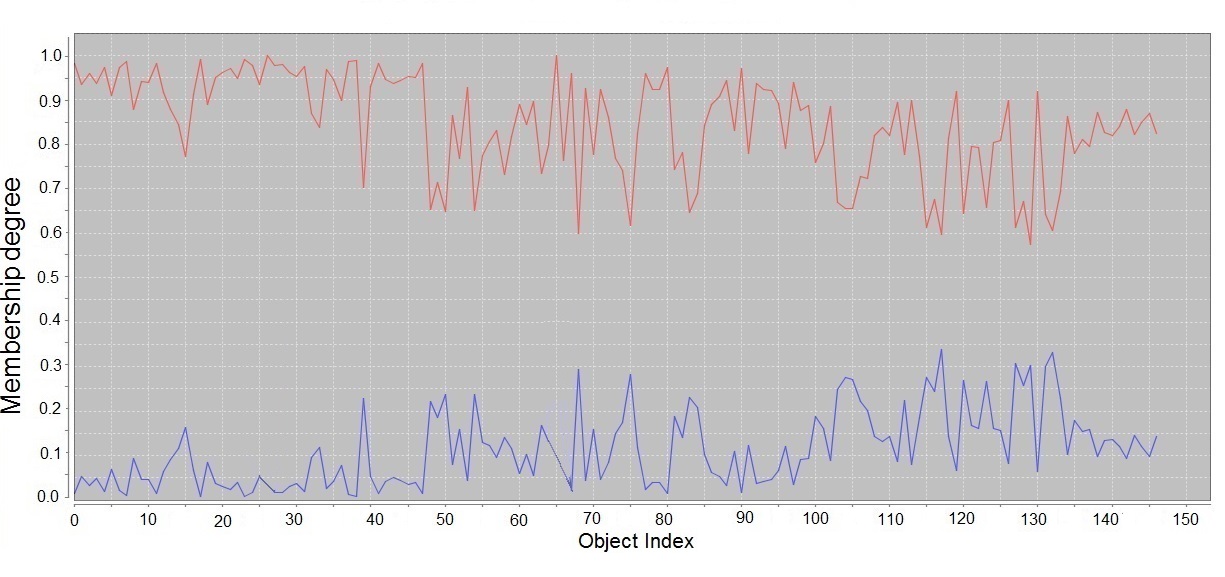}}\vfill}}
\end{center}
\caption{Mutation plot for Iris data objects for their own and the closest clusters, obtained by Fuzzy methods.}
\label{Iris-F}
\end{figure}

\begin{figure}[!ht]
\begin{center}
\leavevmode\fbox{\parbox[b][6cm][s]{160mm}{
\vfill\footnotesize {\includegraphics[width=16cm,height=6cm]{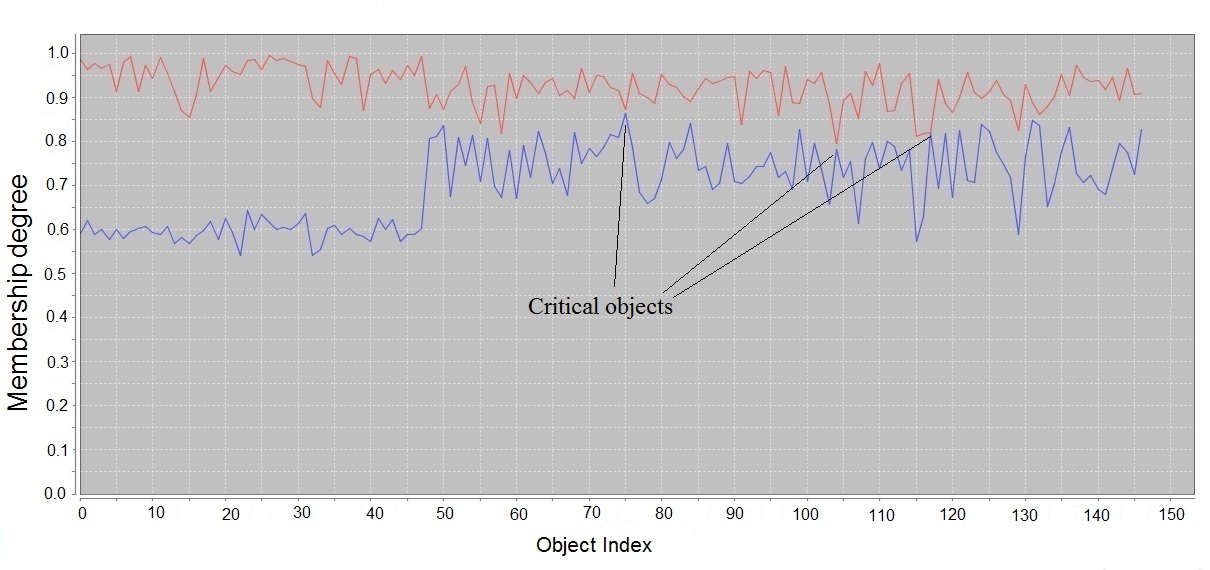}}\vfill}}
\end{center}
\caption{Mutation plot for Iris data objects from their own to the closest cluster, obtained by BFPM.}
\label{Iris-BFPM}
\end{figure}

According to the presented problems in medicine, security, and risk management, partitioning data objects into normal and attack, healthy and unhealthy, or normal and risky categories is useful but is not sufficient. Advanced systems claim more sophisticated analysis on data objects, because even small negligence leads to irreparable consequences. To reduce the amount of cost for further investigations and treatments, analysing the behaviour of data objects and studying their potential abilities to move from their own cluster to another are the reasonable solutions. To get a better understanding, let us look at the well-known benchmark datasets "Iris", which is selected from UCI repository. Iris dataset is commonly used by researchers and the properties of this dataset is clearly presented and explained, which is the main reason that Iris dataset is selected for this analysis. For the analysis, fuzzy and BFPM clustering methods have been applied on Iris dataset, where the results are depicted by Fig. \ref{Iris-F} and Fig. \ref{Iris-BFPM} respectively. 

\begin{figure}[!ht]
\begin{center}
\leavevmode\fbox{\parbox[b][5.8cm][s]{160mm}{
\vfill\footnotesize {\includegraphics[width=16cm,height=5.8cm]{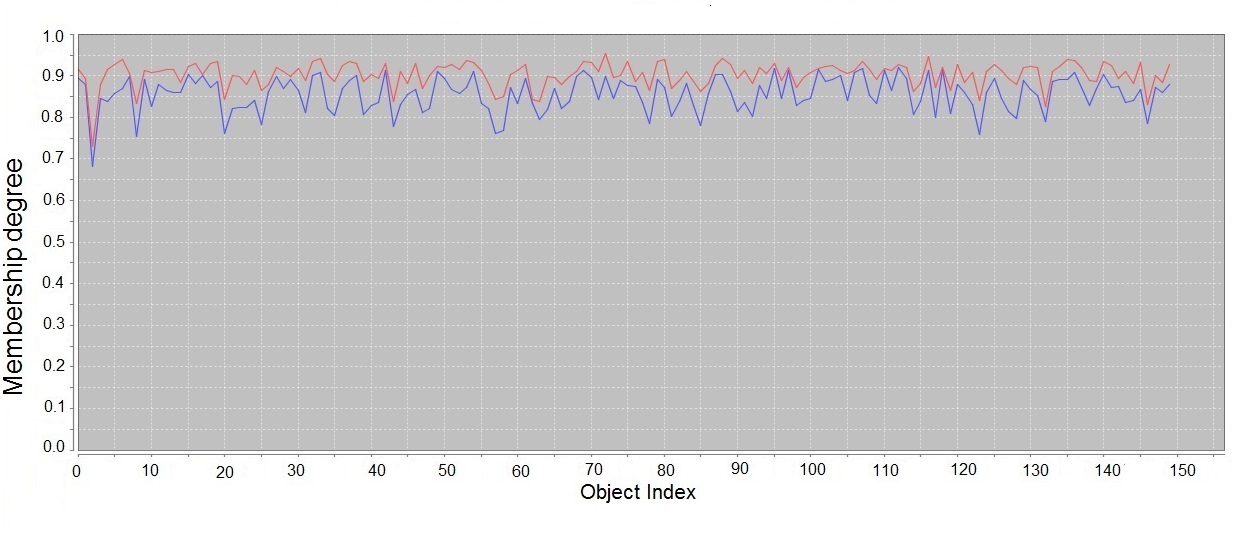}}\vfill}}
\end{center}
\vspace*{-2mm}
\caption{Mutation plot for MAGIC data objects for their own and the closest clusters, obtained by BFPM.}
\label{Magic-BFPM}
\end{figure}

\begin{figure}[!h]
\begin{center}
\leavevmode\fbox{\parbox[b][5.8cm][s]{160mm}{
\vfill\footnotesize {\includegraphics[width=16cm,height=5.8cm]{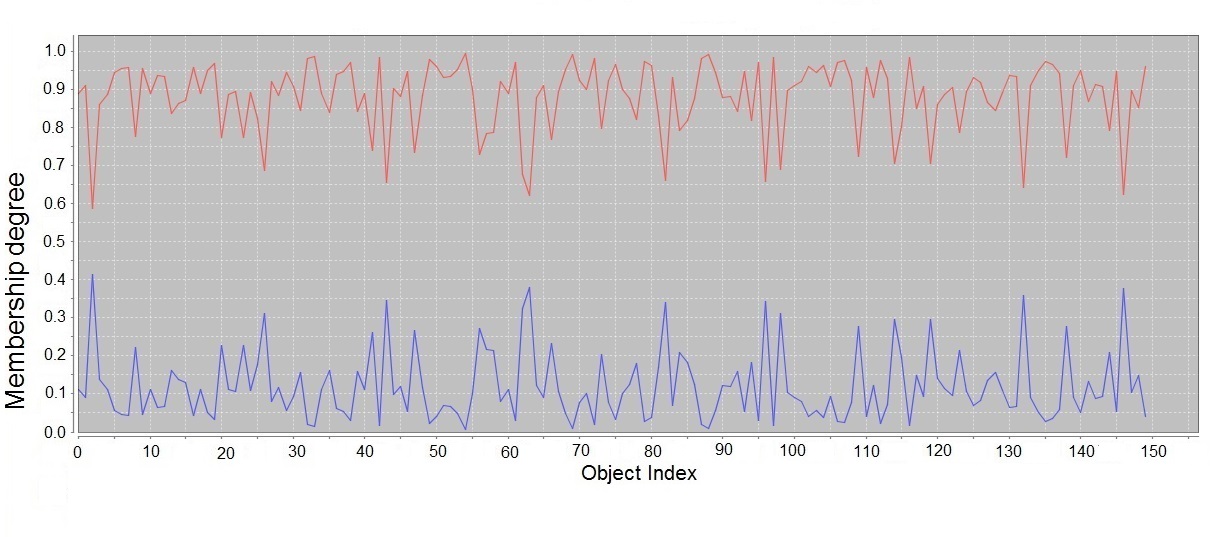}}\vfill}}
\end{center}
\vspace*{-2mm}
\caption{Mutation plot for MAGIC data objects for their own and the closest clusters, obtained by Fuzzy methods.}
\label{Magic-F}
\end{figure}

\hspace*{-7.5mm} Plots for Iris data objects are compared with regard to mutation analysis, where X axis presents data objects and Y axis shows the memberships obtained by data objects with respect to their own cluster and the closest neighbour cluster. Upper points with respect to each data object illustrate the memberships obtained by objects from their own cluster, and the lower points demonstrate the memberships that each data object obtained from the closest cluster. Based on the plots, it can be concluded that fuzzy methods cluster data objects with the aim of completely separating data objects from each other using the fuzzy condition $(\sum_{i=1}^c u_{ij} =1)$. This methodology is good for partitioning purposes, but for tracking and studying the behaviour of data objects for the near future and further analysis is not useful as objects are completely separated from each other. The fuzzy (probability) condition for data objects with respect to clusters leads to higher memberships for the current cluster and very low memberships for the closest cluster, which does not provide a suitable search space to track the potential abilities of data objects in their future movements. That is the reason why Fig. \ref{Iris-F} does not show any critical object or critical area from Iris dataset. On the other side, Fig. \ref{Iris-BFPM} that makes use of BFPM methodology, presents critical objects and critical areas for Iris dataset.\\
BFPM allows to study and track the behaviour of data objects with respect to all clusters and also to evaluate their movements in the near future in addition to well functionalities in partitioning concepts. This facility is provided by BFPM based on its condition in learning procedures $(0< \; 1/c \sum_{i=1}^c u_{ij} \leq 1)$, which allows objects to obtain full memberships from more, even all clusters without any restrictions. The potential abilities of data objects for moving from one cluster to another can be better analysed when data objects are allowed to obtain higher memberships with respect to all clusters, which this condition is clearly depicted by Fig. \ref{Iris-BFPM}. According to the figure, the critical objects and areas that might trap the methods in their learning procedure by miss-assignments are clearly presented. These objects can be found in almost all datasets, which have special properties that should be considered by the methods, otherwise the accuracy might not be desirable. Lack of this consideration might lead the systems to facing the serious consequences specially crucial systems in medicine, finance, security, and so on. In conclusion, type of data objects is another key factor that should be considered by learning methods.\\
Critical objects are very important in the way that learning methods might give them the wrong memberships with respect to all clusters which influence the final results. BFPM checks how far data objects are from the other clusters and how critical objects are close to neighbour clusters. This perspective allows investigators not only to find the crucial areas and objects that learning methods might be trapped by, but also gives more opportunities to evaluate and study the future behaviour of such objects. Table \ref{Move} shows the potential ability of data objects, from different benchmark datasets, to move from one cluster to another by getting even small changes in their feature spaces (mutation). 
%
\begin{table}[!ht]
\caption{Critical objects with ability to move from one cluster to another (Mutation).}
\vspace*{-5mm}
\label{Move}
\begin{center}
\begin{tabular}{ | c | c | c | c | c |  }
\hline
{\bf Dataset} & {\bf No. Objects} & {\footnotesize \bf Objects} $\bf > 85 \% $ & {\footnotesize \bf Objects}$\bf > 75 \% $ & {\footnotesize \bf Objects}$\bf > 70 \% $     \\
\hline
\hline
Iris & 150 & 25   & 99 & 99     \\
\hline
Pima &  768  & 677  &  751 & 751  \\
\hline
Yeast 	&1299 & 868 &  1135 & 1264  \\
\hline
MAGIC & 19200 & 18742 &  18776 & 19199  \\
\hline
Dermatology  & 358   & 286 & 331  & 355   \\
\hline
Libras &  360  & 238  & 317  & 336  \\
\hline
\end{tabular}
\end{center}
\end{table}
%
The table shows the number of data objects with their ability to get partial memberships from the closest cluster. For example, the first row of the table is about data objects from "Iris" dataset, where 25 data objects have over 85 percent as potential ability to move to the closest cluster, and 99 data objects have over 75 percent ability. In other words, 25 objects have obtained the memberships of $0.85$ from the closest cluster. This experiment is very helpful in crucial systems such as medicine and security where we need to know what transactions might be risky in the near future by getting small changes. In medicine, we need to prevent normal cells to move to cancer or diseases clusters and also we aim to encourage those cancer cells to move to normal clusters. Fig. \ref{Magic-BFPM} and Fig. \ref{Magic-F} plot 150 data objects from  MAGIC datasets and their membership degrees obtained by BFPM and fuzzy methods respectively. By considering the vertical axes, we see that data objects can show their abilities with respect to each cluster by using BFPM. Critical areas and objects are completely presented in plots obtained by BFPM. BFPM is able to show the critical objects in case if the system needs to check the critical areas for further analysis, but fuzzy methods mostly keep the objects separated and the system has difficulties for mutation analysis. BFPM can be applied in multi-objective optimization problems to find the optimized and global solutions, where a comprehensive method is needed to provide the critical solutions \cite{hundred-sixteen}.

\section{BFPM Evaluation Using Different Similarity Functions}
So far, the Thesis evaluated the functionality of BFPM in different domains in addition to objects movement analysis. The presented experiments proved the hypothesis that critical objects might lead the systems to improper accuracy if the methods treat all types of objects in the same way or do not pay enough attention to critical objects or do not provide the flexible search space. In this section, the Thesis aims to study the impact of features on final results by comparing the proposed methods when using different similarity functions. The Thesis considers other parameters that might increase the accuracy of BFPM by considering the key components such as covering both the feature and the vector spaces (diversity) and handling the impact of dominant features that fully discussed in the previous chapters. In this experiment, the Thesis compares the accuracy of BFPM when it makes use of the general $L_2$ norm distance function in comparison with Weighted Feature Distance (WFD) function. The results are presented by Table \ref{Accuracy WFD}, which are obtained by assigning different weights to features $(w=\frac{1}{2}, w=\frac{1}{3}, w=\frac{1}{4}, \; and \; w=\frac{1}{d})$, where $d$ is the number of dimensions or the number of features. The table presents the results of BFPM and BFPM-WFD algorithms for different datasets.\\

\begin{table}[!ht]
\caption{Accuracy rates of BFPM based on different distance functions and different assigned weights $(w)$ to objects' features.}
\label{Accuracy WFD}
\begin{center}
\begin{tabular}{ | c | c | c | c | c | c | c |}
\hline
{\bf Distance Function} & {\footnotesize \bf Iris} & {\footnotesize \bf Pima} & { \footnotesize \bf Yeast} & {\footnotesize \bf MAGIC} & {\footnotesize \bf Dermatology} &   {\footnotesize \bf Libras} \\
\hline
\hline
{\bf Euclidean}  & 97.33 & 99.90 & 67.71 & { 100.00} & 77.40 & 57.00 \\
\hline
$\bf WFD_{w=\frac{1}{2}}$   & { 100.00} & { 100.00} & 77.20 & { 100.00} & 89.50 &  69.00 \\
\hline
$\bf WFD_{w=\frac{1}{3}}$   & { 100.00} & { 100.00} & 77.30 & { 100.00} & 83.00 &  62.50 \\
\hline
$\bf WFD_{w=\frac{1}{4}}$   & { 100.00} & { 100.00} & 82.03 & { 100.00} & 86.00 &  61.38 \\
\hline
$\bf WFD_{ w=\frac{1}{d}}$  & { 100.00} & { 100.00} &  82.03 & { 100.00} &  92.40 & 61.40 \\
\hline
\end{tabular}
\end{center}
\end{table}

\begin{figure}[!ht]
\begin{center}
\leavevmode\fbox{\parbox[b][5.5cm][s]{160mm}{
\vfill\footnotesize {\includegraphics[width=16cm,height=5.5cm]{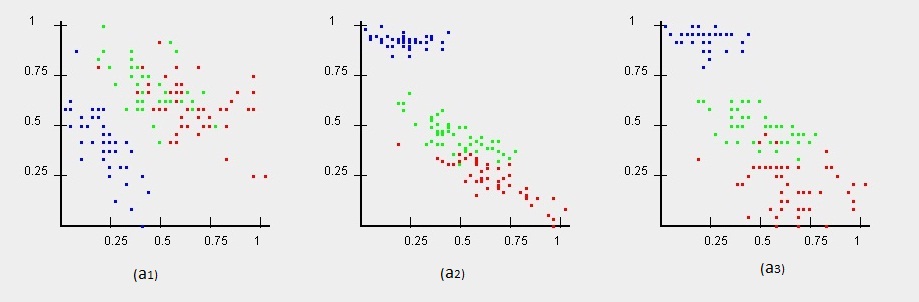}}\vfill}}
\end{center}
\caption{Iris dataset's plot with respect to assigned weights to features clustered into three clusters. ($a_1$) illustrated based on the feature weights (1,1,0,0), ($a_2$) shown based on the feature weights (1,0,1,0), and ($a_3$) depicted based on the feature weights (1,0,0,1).}
\label{3cluster}
\end{figure}

\begin{figure}[!ht]
\begin{center}
\leavevmode\fbox{\parbox[b][5.5cm][s]{160mm}{
\vfill\footnotesize {\includegraphics[width=16cm,height=5.5cm]{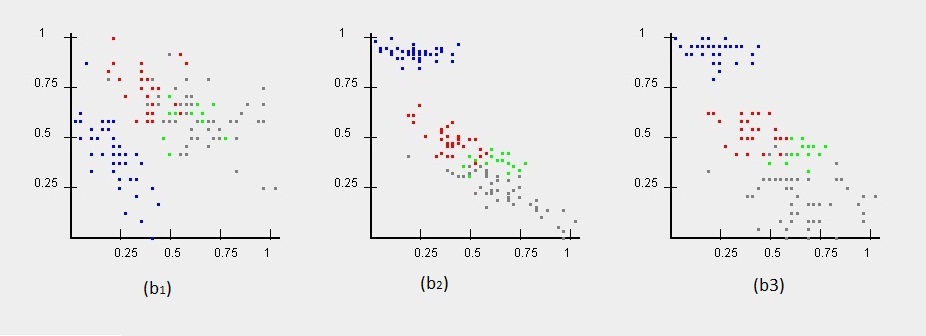}}\vfill}}
\end{center}
\caption{Iris dataset's plot with respect to assigned weights to features clustered into four clusters. ($b_1$) illustrated based on the feature weights (1,1,0,0), ($b_2$) shown based on the feature weights (1,0,1,0), and ($b_3$) depicted based on the feature weights (1,0,0,1).}
\label{4cluster}
\end{figure}

\hspace*{-7.5mm} By comparing the results of the table, we see that BFPM-WFD performs better than BFPM in all datasets. The interesting point is that in BFPM-WFD, all the assigned weights which are less than one lead to obtaining better results and in some datasets such as "Libras" the larger weights result in the better performance of the algorithm, but in some other datasets such as "Yeast" the better results are obtained by the smaller weights. Surprisingly, we can see from the results obtained for "Dermatology" dataset that the better results cannot be obtained in a descending or ascending order. Larger weights sometimes lead to better outcomes, although the best result for Dermatology has been obtained by $(w=\frac{1}{d})$. This experiment reminds us the main concept in optimizations problems, which the methods might be trapped by local optima. This experiment encourages the author of the Thesis to apply more experiments to select the proper weights in similarity measurements to reach the global optima. This experiment also indicates that using the flexible search space results in obtaining better results, although there are still some other parameters that might affect the final results if the methods leave them untouched. In other words, covering diversity cannot be handled just by providing the relaxed searching space and the methods need to make use of other sophisticated approaches in their learning procedures such as similarity functions and considering the type of objects.\\
This experiment was a good motivation to move forward to consider different types of objects and also consider different types of features to cover all the necessities of providing a comprehensive learning approach. The Thesis aims to consider some more experiments on similarity functions by considering different norms of similarity functions in addition to assign weights to the features. This experiment is designed to find an answer for the hypothesis that whether different norms of similarity functions have different impacts on the results of the methods or not. But before applying different similarity functions to evaluate the functionality of the proposed methods, getting a visualized view on the results obtained by clustering methods that make use of Weighted Feature Distance (WFD) functions directs to better understanding. Fig. \ref{3cluster} and Fig. \ref{4cluster} show the effects of assigning weights to features on the final partitioning results obtained by BFPM for Iris dataset in order to cluster objects for three and four number of clusters respectively. \\

\hspace*{-7.5mm} The figures are depicted based on different sets of weights assigned to features of each individual object as: ($a_1$) feature weights (1,1,0,0), ($a_2$) feature weights (1,0,1,0), and ($a_3$) feature weights (1,0,0,1). Each set of assigned weights leads to obtaining different partitioning results. Evaluating the results from assigned weights prevents the system of being trapped by dominant features, specially in high dimensional search spaces. Fig. \ref{3cluster} indicates that the second and the third set of weights are well selected, while the first set of weights are not proper for this problem as in $(a_1)$ objects are not well clustered. This experiment indicated the necessity of considering weighted similarity functions in learning procedures to not only obtain the proper results but also to prevent the systems of being trapped by wrong similarity measurements. The same conditions applied on Iris dataset to cluster objects into four clusters depicted by Fig. \ref{4cluster}. The same weights assigned to features presented by ($b_1$) feature weights (1,1,0,0), ($b_2$) feature weights (1,0,1,0), and ($b_3$) feature weights (1,0,0,1). Figures emphasize the importance of considering the impact of features on the final results of partitioning problems. Table \ref{Accuracy WFD-m} shows the results of BFPM and BFPM-WFD on different distance functions, different assigned weights, and different fuzzification constants $m$. The results prove the hypothesis that covering diversity in both the feature and the vector spaces are needed in order to obtain the desirable results. \\

\begin{table}[!ht]
\caption{Accuracy rates of BFPM based on different distance functions by assigning different weights $(w)$ and fuzzification constant $(m)$.}
\label{Accuracy WFD-m}
\vspace{-4mm}
\begin{center}
\begin{tabular}{ | c | c | c | c | c | c | }
\hline
\multirow{2}{*}{\bf  Dataset} & \multirow{2}{*}{\bf Distance Function} & $\bf m =1.2  $  & $\bf m =1.6   $ & $\bf m =1.8  $  & $\bf m =2.0  $\\
\cline{3-6}
 & & \multicolumn{4}{  c |}{$\bf w =\frac{1}{d}$}  \\
\hline
\multirow{2}{*}{\footnotesize \textbf{Iris}} 
	& BFPM & 97.33 &   100.00  & 100.00 & 100.00 \\ 
	\cline{2-6}
	& BFPM-WFD  &  100.00 &  100.00  & 100.00 & 100.00   \\
\hline
\hline
\multirow{2}{*}{\footnotesize \textbf{Pima}} 
	& BFPM &  99.00  &  100.00  & 100.00 & 100.00 \\
	\cline{2-6}
	& BFPM-WFD  &  100.00 &  100.00  & 100.00 & 100.00    \\
\hline
\hline
\multirow{2}{*}{\footnotesize \textbf{Yeast}} 
	& BFPM &   84.82  & 81.29 & 84.99 & 67.71 \\
	\cline{2-6}
	& BFPM-WFD &  93.51 & 95.84  & 88.47 & 82.30   \\
\hline
\hline
\multirow{2}{*}{\footnotesize \textbf{MAGIC}} 
	& BFPM &  99.90  &  100.00  & 100.00 & 100.00 \\
	\cline{2-6}
	& BFPM-WFD  &  100.00 &  100.00  & 100.00 & 100.00   \\
\hline
\hline
\multirow{2}{*}{\footnotesize \textbf{Dermatology}} 
	& BFPM &  96.08  &  69.58 & 65.89 &77.40 \\
	\cline{2-6}
	& BFPM-WFD &  98.83 &  97.44 & 89.93 & 92.40   \\
\hline
\hline
\multirow{2}{*}{\footnotesize \textbf{Libras}} 
	& BFPM &  95.11  & 83.44 & 62.77 &  57.00 \\
	\cline{2-6}
	& BFPM-WFD  &  99.72 & 90.27 & 86.11 & 61.40   \\
\hline
\end{tabular}
\end{center}
\end{table}

\hspace*{-7.5mm} According to the results, we can see that in some datasets different norms have different effects on the final results. Despite the fact that the accuracy of the algorithm is very similar in almost all cases when the algorithm is applied for the low number of clusters (Iris, Pima, and MAGIC datasets), for other datasets that are supposed to be clustered in the higher number of clusters the best results should be evaluated based on different norms. It means that the methods might perform well for those datasets that are supposed to be binary clustered. Based on the results, we can see for some datasets such as Dermatology, the lower norms lead to better results, while for some other datasets such as Iris and MAGIC the higher norms results in better outcomes. It should be noted that alike the optimization approaches, the methods can be trapped by local optima and covering more experiments and proper selection of weights and similarity functions in different norms leads to obtaining the global optima.  \\
In conclusion, different weights and different similarity functions result in different outcomes. Similarity functions that do not perform in both the vector and the feature spaces might lead to improper accuracy for some datasets. Handling the impact of features, specially dominant features cannot be covered by all types of similarity functions. \\
\vspace*{-5mm}
\section{Experiments on Clustering Strategy}
In this experiment, the Thesis compares the accuracy of the proposed methods by applying the presented algorithms FPM-I, FPM-II, BFPM, and BFPM-WFD on "Iris", "Pima Indian", "Yeast", and "MAGIC". The fuzzification constant for this experiment is the same for all algorithms ($m=2$). The idea is whether the proposed algorithms have made any improvement, where some of the algorithms such as FPM-I, FPM-II, and BFPM-WFD consider the impact of features (dominant features) on the final results and some of them such as BFPM and BFPM-WFD consider critical objects and areas in their learning procedures. The results are depicted on Table \ref{Accuracy-BFPM-FCM} with regard to datasets . This experiment is designed to verify whether different membership assignments have different influences on the final results, or membership functions lead to the same results.\\

\begin{table}[!ht]
\caption{Accuracy rates of different versions of BFPM algorithms: FPM-I, FPM-II, BFPM, and BFPM-WFD using Euclidean distance function ($L_2$ norm) and WFD function applying fuzzification constant ($m=2$) is compared.}
\label{Accuracy-BFPM-FCM}
\vspace{-4mm}
\begin{center}
\begin{tabular}{ | c | c | c | c | c | c | }
\hline
{ \bf Method $\downarrow$ $\Big{|}$ $\underset{\longrightarrow}{\bf Dataset}$} & {\bf Iris} & {\bf Pima Indian} & {\bf Yeast} &  {\bf MAGIC}  \\
\hline
\hline
{\footnotesize {\bf FPM-I}} 		& 94.11 & 67.00  & 67.40 & 60.00 \\
\hline
{\footnotesize {\bf FPM-II}} 		& 93.19 & 76.00  & 68.00 & 63.30  \\
\hline
{\footnotesize {\bf BFPM}} 	 	&  97.33  &  99.90  &  67.71 & 100.00  \\
\hline
{\footnotesize {\bf BFPM-WFD}} 	 	&  100.00  &  100.00  &  82.30 & 100.00  \\
\hline
\end{tabular}
\end{center}
\end{table}

\hspace*{-7.5mm} According to the results from Table \ref{Accuracy-BFPM-FCM}, it can be concluded that considering the feature spaces leads to better results. More interesting point is that providing the more suitable search space leads to better outcomes. By considering the results from FPM-I and FPM-II, we can see that considering the feature spaces during the learning procedure led to just better results in Iris dataset. This evaluation indicates that the effects of dominant features in lower dimensional search spaces should be handled during the learning procedures. By comparing the results obtained by FPM-I, FPM-II, and BFPM, we can conclude that considering critical objects leads to better outcome as BFPM provides the most flexible search space for data objects to obtain memberships from all clusters. This facility allows the algorithm to evaluate objects with respect to all clusters for all iterations, which consequently results in better outcomes. According to this evaluation, it is clear that effects of critical objects on the final results are much higher than the effects of features (dominant).  
By comparing the results from BFPM and BFPM-WFD, we can see that BFPM-WFD performs better than BFPM as BFPM-WFD considers more parameters in its learning procedures by evaluating critical objects and the impact of dominant features, while BFPM just considers the effects of critical objects on the final result.  \\
In conclusion, this experiment with its results proved that there are some crucial parameters in membership assignments and similarity evaluations that should be covered by any membership and similarity functions that are used in learning procedures in order to obtain the desirable results. The results show that FPM-I and FPM-II perform better for  some datasets as they aim to consider the impact of features on their learning procedures in different strategies. BFPM membership assignment, proposed in this Thesis, led to better outcomes as this methodology considers the effects of critical objects in learning procedures. BFPM-WFD performs better than other algorithms as this algorithm gives more opportunities to critical objects to participate in all clusters and also considers the effects of features, which results in better outcomes. \\

\section{Comparison between BFPM and Other Centroid-Based Methods}
Iris dataset, as a well-known dataset in machine learning, has been always considered for the experiment to check the functionality of a new proposed method. Iris dataset is also considered for the evaluation step in this Thesis. the Thesis aims to compare the results from BFPM with other well-known centroid-based clustering methods (FCM, K-means , and PCM) with regard to membership assignments. This experiment compares different versions of BFPM (FPM-I, FPM-II, and BFPM) with modified Fuzzy C-Means (FCM), modified Possibilistic C-Means (PCM), and fuzzy K-menas, presented by Table \ref{Accuracy-Iris-Unsupervised}. The modified methods are LAC (Locally Adaptive Clustering) \cite{hundred-thirty-one}, WLAC (Weighted Locally Adaptive Clustering) \cite{hundred-thirty-one}, FWLAC (Fuzzy Weighted Locally Adaptive Clustering) \cite{hundred-thirty-one}, CD-FCM (Collaborative Distributed Fuzzy C-Means) \cite{seventeen}, KCD-FCM (Kernel-based Collaborative Distributed Fuzzy C-Means) \cite{seventeen}, KFCM-F (Kernel-based Fuzzy C-Means and Fuzzy clustering) \cite{hundred-thirty-two}, WEFCM (Weighted Entropy-regularized Fuzzy C-Means) \cite{seventeen}, and APCM (Adaptive Possibilistic C-Means) \cite{eighty-two}.\\

\begin{table}[!ht]
\caption{Accuracy rates of modified methods: k-means, FCM, and PCM in comparison with the proposed FPM-I, FPM-II, and BFPM for Iris dataset. }
\label{Accuracy-Iris-Unsupervised}
\begin{center}
\setlength\tabcolsep{4.5pt}
\begin{tabular}{ | c | c | c | c | c | c | c | c | c | c | c | c |}
\hline
\multicolumn{12}{|c|}{{\color{white} \rot{.....}} Iris Dataset}\\[1ex]
\hline
{\color{white} \rot{.....}}& \multicolumn{7}{c|}{Modified FCM} & {PCM} & \multicolumn{3}{c|}{BFPM-Versions} \\[1ex]
\hline
 {\small \rot{Fuzzy} \rot{k-means}}
 & {\small \rot{LAC}} & { \small \rot{WLAC}} & {\small \rot{FWLAC}} & {\small \rot{CD-FCM}}  & {\small \rot{KCD-FCM }} &  {\small \rot{KFCM-F}} & {\small \rot{WE-FCM}} &   {\small \rot{APCM}} & {\small \rot{FPM-I}} &  {\small \rot{FPM-II}} & {\small \rot{BFPM}} \\
\hline
\hline
{\color{white} \rot{.....}}{ 74.96} & { 90.21}  & { 90.57} & { 94.37} & { 95.90}  & { 96.18}  & { 92.06} & {96.66} &  {96.74} & {94.11}  & {93.19} & {{97.33}}  \\[1ex]
\hline
\end{tabular}
\end{center}
\end{table}
%
\hspace*{-7.5mm} The results were achieved by assigning the fuzzification constant as ($m=2$). According to the results presented by the table, FPM-I and FPM-II perform better than fuzzy k-means method and some of the modified versions of FCM methods, but some of the modified FCM and modified PCM methods perform better than FPM-I and FPM-II algorithms. According to the results, considering the feature spaces leads to better outcomes as FPM-I and FPM-II consider the feature spaces to handle the impact of dominant features. Fuzzy weighted and Weighted Entropy perform better than FPM-I and FPM-II by tuning the weights assigned to features. APCM performs better than other presented methods by using more relaxed search space in comparison with other modified fuzzy methods. According to the results, BFPM performs better than all modified versions of k-means, FCM, and PCM methods by providing the most flexible search space for data analysis. According to the interest of obtaining   more achievements in medicine, several approaches have been applied on medicine-related datasets. Regarding that, a dataset related to diabetes has been chosen for this experiment. Table \ref{Accuracy-Pima-part1} presents the accuracy rates of modified FCM and k-means methods in comparison with the proposed FPM-I, FPM-II, and BFPM for Pima dataset. The modified methods compared on Pima dataset are k-means++ \cite{hundred-thirty-three}, ik-means (intelligent K-Means) \cite{hundred-thirty-three}, $H \& W$ (Hartigan and Wong) \cite{hundred-thirty-three}, PAM (Partition Around Medoids) \cite{hundred-thirty-three}, MoG (Mixtures of Gaussian) \cite{hundred-thirty-four}, PMoG (Possibilistic Mixtures of Gaussian) \cite{hundred-thirty-four}, and PFCM (Possibilistic Fuzzy C-Means) \cite{hundred-thirty-four}. According to the results, BFPM performs better than other methods for Pima dataset, where the  number of clusters is two. \\

\begin{table}[!ht]
\caption{Accuracy rates of modified FCM and k-means methods in comparison with the proposed FPM-I, FPM-II, and BFPM for Pima dataset.               
}
\label{Accuracy-Pima-part1}
\begin{center}
\setlength\tabcolsep{4.5pt}
\begin{tabular}{ | c | c | c | c | c | c | c | c | c | c | c |}
\hline
\multicolumn{10}{|c|}{{\color{white} \rot{.....}} Pima Dataset}\\[1ex]
\hline
\multicolumn{4}{|c|}{{\color{white} \rot{.....}} Modified K-means} & \multicolumn{3}{c|}{{\color{white} \rot{.....}} Modified C-means} & \multicolumn{3}{c|}{BFPM-Versions}\\[1ex]
\hline
{\color{white} \rot{.....}}  {k-means++} &  {ik-means} & {PAM} & {$H \& W$} & MoG  &  PMoG &  PFCM &  {FPM-I} & {FPM-II} & {BFPM}  \\
\hline
\hline
{\color{white} \rot{.......}} 67.00 & 67.00  & 67.00 & 67.00 & 72.88 & 74.49 & 71.05 & 67.00  & 76.00 & {99.90}  \\[1ex]
\hline
\end{tabular}
\end{center}
\end{table}
\hspace*{-7.5mm} Biology and bioinformatics also make use of learning methods either supervised or unsupervised for more achievements. To apply the proposed methods in these domains, Yeast dataset that is related to molecular and cellular biology for proteins has been used in this experiment. Table \ref{Accuracy-Yeast-part1} presents the accuracy rates of modified FCM methods in comparison with the proposed FPM-I, FPM-II, and BFPM for Yeast dataset. The modified methods are LAC (Locally Adaptive Clustering) \cite{hundred-thirty-one}, WLAC (Weighted Locally Adaptive Clustering) \cite{hundred-thirty-one}, and FWLAC (Fuzzy Weighted Locally Adaptive Clustering) \cite{hundred-thirty-one}. According to the results, BFPM perform better than other methods, but the more interesting points is that unlike the results for Iris dataset, assigned weights by WLAC and FWLAC did not lead to better results. One of the reasons for this scenario might be the reason, that the Thesis discussed in the previous experiment, is the affect of dominant features in higher dimensional search space. As mentioned, the influences of each individual feature on the final results may decline due to higher dimensionality.  \\

\begin{table}[!ht]
\caption{Accuracy rates of modified FCM methods in compare with the proposed FPM-I, FPM-II, and BFPM for Yeast dataset.               
}
\label{Accuracy-Yeast-part1}
\vspace*{-3mm}
\begin{center}
\begin{tabular}{ | c | c | c | c | c | c |}
\hline
\multicolumn{6}{|c|}{{\color{white} \rot{.....}} Yeast Dataset}\\[1ex]
\hline
\multicolumn{3}{|c|}{{\color{white} \rot{.....}} Unsupervised learning methods} & \multicolumn{3}{c|}{BFPM-Versions}\\[1ex]
\hline
{\color{white} \rot{.....}}  { LAC} &  { WLAC} & { FWLAC}  & { FPM-I} & { FPM-II} & { BFPM}   \\[1ex]
\hline
\hline
{\color{white} \rot{.....}} 44.28 &  45.33 & 45.38  & 67.40 & { 68.00}   &  67.71 \\[1ex]
\hline
\end{tabular}
\end{center}
\end{table}
\hspace*{-7.5mm} As discussed earlier, machine learning has widely used in diagnosing diseases and this Thesis aims to apply the proposed methods on different datasets for diagnosis purposes using clustering and classification techniques. In follow, the proposed methods has been applied on Dermatology dataset for diagnosis of Eryhemato-Squamous diseases. The accuracy rates of centroid-based methods in comparison with BFPM and BFPM-WFD for Dermatology dataset have been presented by Table \ref{Accuracy-Dermatology-part1}. The modified methods are Mediod \cite{hundred-thirty-five}, Centroid \cite{hundred-thirty-five}, Dataset Seeded Centroid (DS-Centroid) \cite{hundred-thirty-five}, K-Means Seeded Centroid (KMSCentroid) \cite{hundred-thirty-five}, and k-means \cite{hundred-thirty-five}. According to the results, we can conclude that BFPM performs better than some other methods by providing the BFPM methodology in searching space, but in order to obtain the best performance other parameters such as considering different similarity functions is needed.\\

\begin{table}[!ht]
\caption{Accuracy rates of modified FCM method and different unsupervised centroid-based methods in comparison with BFPM for Dermatology dataset.               
}
\vspace*{-4mm}
\label{Accuracy-Dermatology-part1}
\begin{center}
\begin{tabular}{ | c | c | c | c | c | c | c | c | }
\hline
\multicolumn{8}{|c|}{{\color{white} \rot{.....}} Dermatology Dataset}\\[1ex]
\hline
\multicolumn{6}{|c|}{{\color{white} \rot{.....}} \multirow{2}{*}{Unsupervised centroid-based methods}} & \multicolumn{2}{c|}{BFPM}  \\
 \multicolumn{6}{|c|}{{\color{white} \rot{.....}}} & \multicolumn{2}{c|}{Versions}\\[1ex]
\hline
 {\color{white} \rot{.....}}  \multirow{2}{*}{Mediod} & \multirow{2}{*}{D-Mediod} & \multirow{2}{*}{Centroid}  & DS & KMS & \multirow{2}{*}{K-means} & \multirow{2}{*}{ BFPM }   & { BFPM- } \\
   &   &    & Centroid & Centroid & &  & WFD  \\[1ex]
\hline
\hline
\multirow{2}{*}{83.18}  & \multirow{2}{*}{73.44} &  \multirow{2}{*}{72.46} & \multirow{2}{*}{80.44}   &  \multirow{2}{*}{80.24}  & \multirow{2}{*}{80.99} & \multirow{2}{*}{77.40} & \multirow{2}{*}{92.40}\\[1ex]
\hline
\end{tabular}
\end{center}
\end{table}
%
\hspace*{-7.5mm} Representing images and sub-images of an image in image processing concept is a main concern in vision systems. According to difficulty of the task of image processing for machines, the main divide and conquer algorithm has been utilized to divide an image into different non-overlapped regions. The procedure of division is called segmentation. Several approaches have been introduced in segmentation clustering with respect to memberships or belongingness of image's pixels. Each pixel (object) obtains
different memberships based on the type of membership function that has been used in learning procedures \cite{hundred-thirty-six}. In this experiment, the accuracy of fuzzy methods with recent clustering methods have been compared with BFPM, where Table \ref{Accuracy-Libras-part1} presents the accuracy rates of several centroid-based clustering methods in comparison with BFPM for Libras dataset. The "Movement-libras" dataset contains 360 objects in 90 dimensional search space (attributes) with respect to 15 clusters. The modified methods are: Mediod, Dynamic Medoid (D-Medoid), Centroid, Dataset Seeded Centroid (DS-Centroid), K-Means Seeded Centroid (KMSCentroid), and k-means \cite{hundred-thirty-five}. According to the results, BFPM performs better than other prototype-based methods.\\
\begin{table}[!ht]
\caption{Accuracy rates of several unsupervised centroid-based methods in compare with the proposed BFPM for Libras dataset.               
}
\vspace*{-3mm}
\label{Accuracy-Libras-part1}
\begin{center}
\begin{tabular}{ | c | c | c | c | c | c | c | }
\hline
\multicolumn{7}{|c|}{{\color{white} \rot{.....}} Libras Dataset}\\[1ex]
\hline
 \multicolumn{6}{|c|}{{\color{white} \rot{.....}} Unsupervised centroid-based methods} & 
 {BFPM Versions}\\[1ex]
\hline
{\color{white} \rot{.....}}  \multirow{2}{*}{Mediod} &  \multirow{2}{*}{D-Mediod } & \multirow{2}{*}{Centroid}  & DS & KMS & \multirow{2}{*}{K-means} & \multirow{2}{*}{BFPM}  \\
   &  &   & Centroid & Centroid & &  \\[1ex]

\hline
\hline
{\color{white} \rot{.....}} 46.80 &  27.99 & 24.89  & 41.48 & 45.06 & 46.40 & 57.00   \\[1ex]
\hline
\end{tabular}
\end{center}
\end{table}
\section{BFPM Functionality}
\subsection{BFPM Reveals Information in Medicine (Lung Cancer Diagnosis)}
\begin{figure*}[!ht]
\begin{center}
\leavevmode\fbox{\parbox[b][11cm][s]{160mm}{
\vfill\footnotesize {\includegraphics[width=16cm,height=11cm]{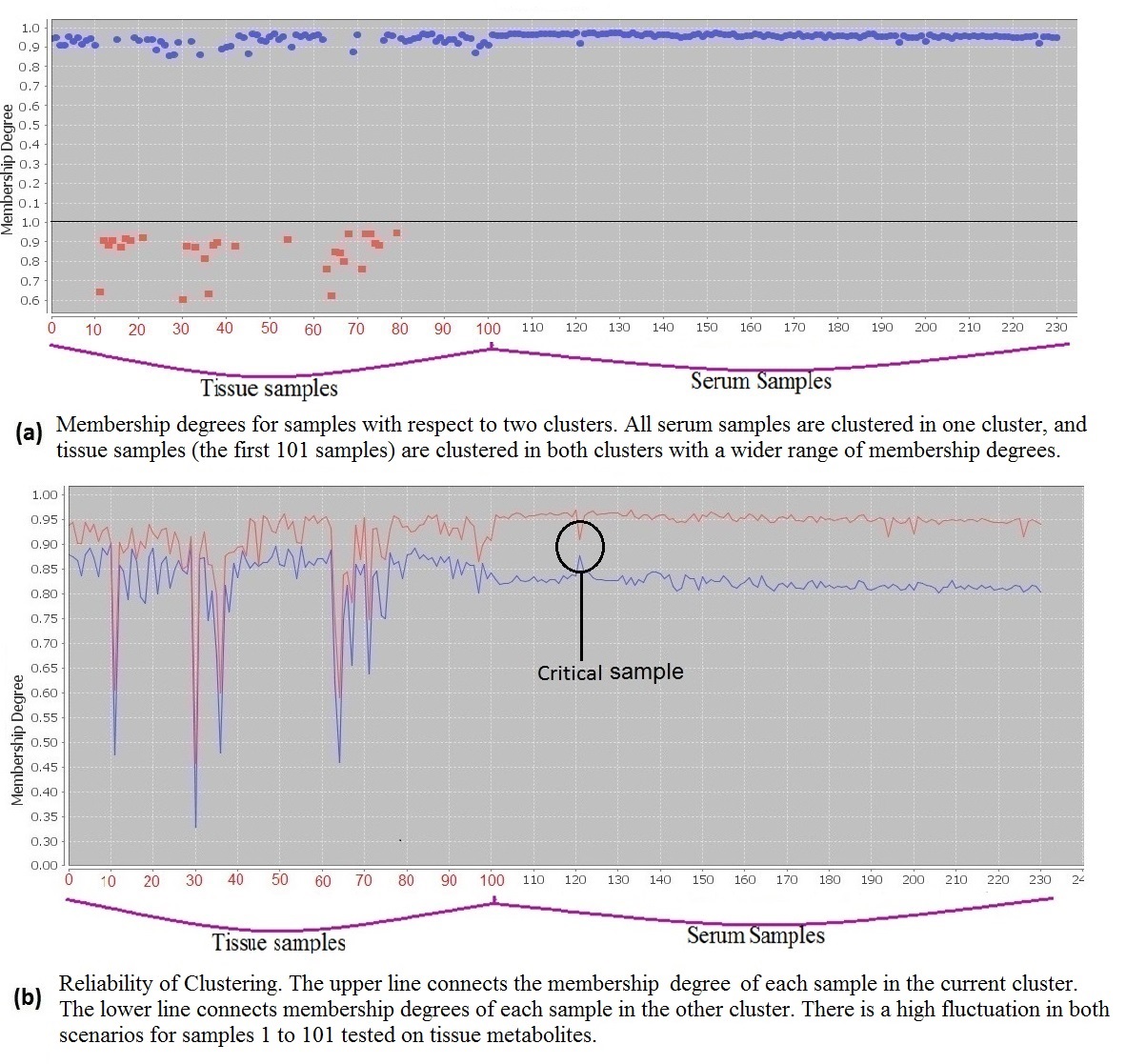}}\vfill}}
\caption{Tissue and serum samples tested by 61 metabolites using BFPM. The first 101 samples on X axis are from tissue of individuals with lung cancer, the other 101 samples are from serum of individuals with lung cancer, and the last 29 samples are from healthy individuals. The Y axis represents the memberships.}
\label{BFPM-F1}
\end{center}
\end{figure*}

\begin{figure*}[!ht]
\begin{center}
\leavevmode\fbox{\parbox[b][10cm][s]{160mm}{
\vfill\footnotesize {\includegraphics[width=16cm,height=10cm]{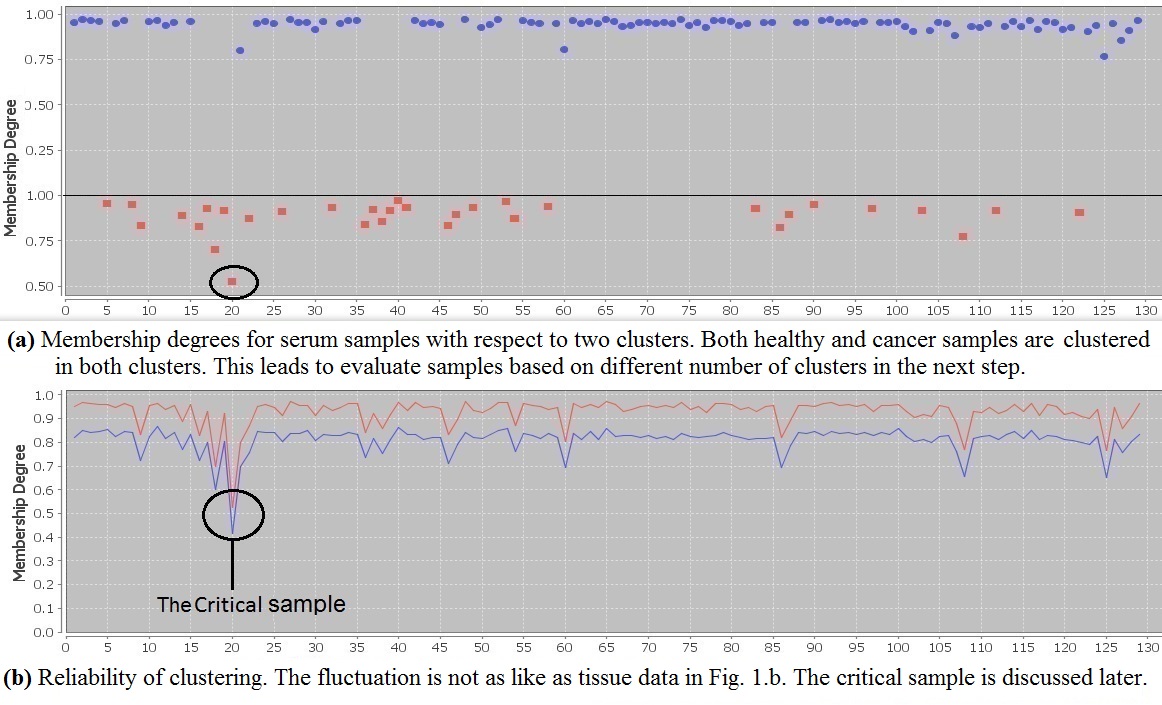}}\vfill}}
\caption{Serum samples tested by 61 metabolites using BFPM. The first 101 samples on X axis are from serum of individuals with lung cancer and the last 29 samples are from healthy individuals. The Y axis represents the memberships.}
\label{BFPM-F2}
\end{center}
\end{figure*}

\begin{figure*}[!ht]
\begin{center}
\leavevmode\fbox{\parbox[b][6.5cm][s]{160mm}{
\vfill\footnotesize {\includegraphics[width=16cm,height=6.5cm]{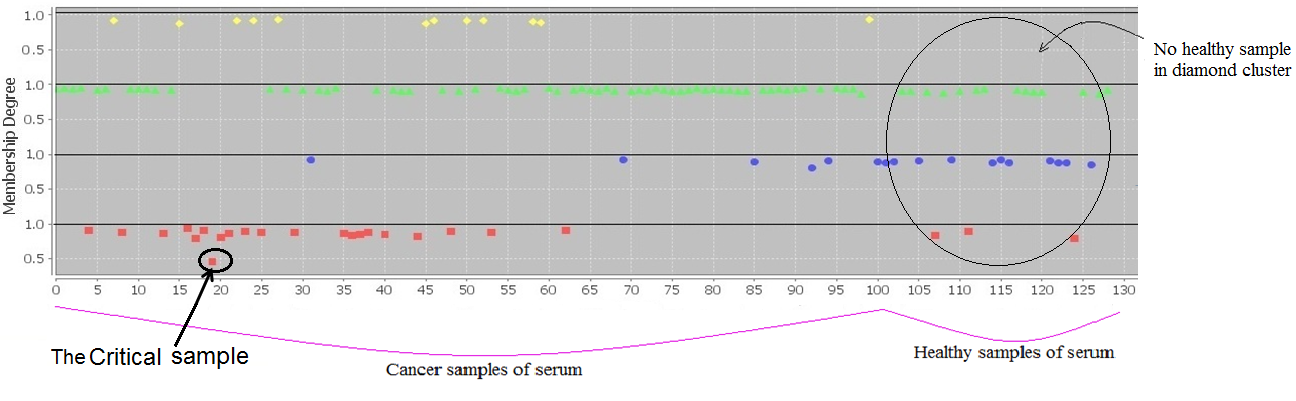}}\vfill}}
\caption{Serum samples in 4 clusters. No healthy sample in the diamond cluster generates a hypothesis that diamond samples are different from the other cancer samples. This hypothesis is strengthened through the covariate assessment.}
\label{BFPM-F3}
\end{center}
\end{figure*}

The dataset that has been used in this experiment is about serum and tissue samples (objects) where obtained from the Harvard/MGH lung cancer susceptibility study repository \cite{hundred-twelve}. Informed consent was obtained from lung cancer patients and healthy controls prior to banking samples and after the nature and possible consequences of the study were explained. There are 231 samples, 101 are tissue samples and 101 are serum samples of the same individuals with lung cancer and 29 are serum samples of healthy individuals. In total, 61 metabolites (spectral regions) were measured in the following process. Researchers were blinded to the status of the samples during all measurement and experimental steps. Samples were stored at $-80^{\circ C}$ until to analysis. High resolution magic angle spinning magnetic resonance spectroscopy (HRMAS MRS) measurements were performed using our previously developed method on a Bruker Avance (Billerica, MA) $600 MHz$ spectrometer. Measurements were conducted at $4^{\circ C}$ with a spin-rate of $3.6 Hz$ and a Carr-Purcell-Meiboom-Gill sequence with and without water suppression. Ten $\mu L$ of serum or ten $\mu g$ of tissue were placed in a $4 mm$ Kel-F zirconia rotor with ten $\mu L$ of $D2O$ added for field locking. HRMAS MRS spectra were processed using AcornNMR-Nuts (Livermore, CA), and peak intensities from $4.5 - 0.5$ ppm were curve fit.\\
Relative intensity values were obtained by normalizing peak intensities by the total intensity of the water unsuppressed file. The resulting values, which were less than 1 percent of the median of the entire set of curve fit values were considered as noise and eliminated. Spectral regions were defined by regions, where 90 percent or more of samples had a detectable value with 32 regions resulting. Following MRS measurement, tissues were formalin-fixed and paraffin-embedded. Serial sectioning was performed by cutting $5 \mu m$-thick slices at $100 \mu m$ intervals throughout the tissue, resulting in $10-15$ slides per piece. After haemotoxylin and eosin $(H \& E)$ staining, a pathologist with $>25$ years experience read the slides to the closest 10 percent for percentages of the following pathological features: cancer, inflammation/fibrosis, necrosis, and cartilage/normal. The available dataset is normalized in an interval $[0 , 1]$, which includes 71 features in total. Features contain five categorical or qualitative variables and all of them are nominal, 61 independent numerical or quantitative variables (metabolites), which all of them are ratio with discrete values, and five numerical variables as ratio type with discrete values. Feature selection techniques also assist the learner to obtain better accuracy in addition to reduce the complexity of the procedure \cite{hundred-twenty-eight}, but in here, the features are considered with the same priority and all the features are evaluated in the clustering procedure.\\
Extracting knowledge from datasets can be obtained by running different methods, either supervised (classification) or unsupervised (clustering) methods. The accuracy of classification techniques is mostly measured through the percentage of the correct labelled samples and strongly depends on the correct selection of samples and the number of samples in each class. A large number of samples in one class dominates the accuracy and using some techniques such as t-test are needed to control the quality for the small sizes \cite{one}. Therefore, in this study, the aim is to utilize clustering techniques due to the different number of case and control individuals in the dataset. Furthermore, clustering techniques can lead to presenting the degree that cases are different from controls. In here, BFPM clustering method is employed to illustrate the similarity between samples. Moreover, covering sample movements and having a flexible search space encouraged to use BFPM Algorithm in clustering methods based on BFPM membership function. The former concept leads to computing the potential ability of each sample to participate in another cluster, and the latter one is to cluster samples based on a flexible search space (diversity). Using BFPM, facilitates finding critical samples those that are about to move from healthy to cancer category or vice versa.\\
Here, BFPM for clustering is utilized to see behavior of samples tested based on metabolites. The BFPM employs membership degrees for clustering samples and provides information about each sample individually in addition to insights into reliability of the clusters. In general, the number of clusters can be estimated using different techniques such as visualization of the dataset, optimization under probabilistic mixture-model framework, using certain validity indices to evaluate the intra-cluster and inter-cluster similarities, and other heuristic approaches \cite{hundred-twenty-nine}. However, each of these approaches suffers some sort of bias. For example, validity indices can be biased through using some parameters in the evaluation's procedures such as dominant feature in similarity functions \cite{hundred-ten}. Visualization techniques on the other hand are extremely sensitive to the chosen dimensions \cite{seven}, \cite{hundred-thirty}. In this study however, the analysis on the results obtained from different number of clusters to find patterns in the data associated with clinic-pathological behavior of lung cancer to differentiate healthy from lung cancer samples is presented. Some results of clustering are depicted in Fig. \ref{BFPM-F1}, Fig. \ref{BFPM-F2}, and Fig. \ref{BFPM-F3}, where on the X axes, samples are located. Fig. \ref{BFPM-F1}(a) depicts the results of clustering for all samples including cancer (both tissue and serum samples) and healthy samples. The first 101 samples of Fig. \ref{BFPM-F1} are tissue metabolites of individuals with lung cancer, the other 101 samples are serum metabolites of individuals with lung cancer, and the last 29 samples are serum metabolites of the healthy individuals. On the Y axes membership degrees for each sample is provided, varying from 0 to 1. 
Fig. \ref{BFPM-F1}(b) shows two lines. The upper one connects the membership degrees of each sample in the current cluster. The lower line connects membership degrees of each sample in the second cluster. According to the figure, all serum samples are clustered in one cluster and tissue samples (the first 101 samples) are clustered in two clusters. This achievement showed that serum samples of healthy and cancer samples have some common properties with some of the tissues samples. \\
To obtain more results, the serum samples are clustered separately. Fig. \ref{BFPM-F2} and Fig. \ref{BFPM-F3} illustrate the results for serum samples of cancer and healthy individuals. The first 101 samples of figures are serum metabolites of individuals with lung cancer and the last 29 samples are serum metabolites of the healthy individuals. Here, we analyze the serum samples excluding the tissue samples with the aim of highlighting similarities and differences among the serum samples. By analysing serum samples with regard to both clusters, Fig. \ref{BFPM-F2}, we see healthy and cancer samples spread in both clusters. This is not surprising due to sharing common parameters between cancer and healthy samples. Extracting similarity between healthy and unhealthy individuals through serum samples was a lead to increase the number of clusters to check in what extend samples show their properties. Therefore, different number of clusters have been chosen and interesting information has been obtained accordingly. The most significant achievements were related to the results of clustering samples into four clusters, presented by Fig. \ref{BFPM-F3}. Digging in Fig. \ref{BFPM-F3}, we could observe differences between healthy and cancer samples, reviewed below. Very interestingly, it is noticed that no healthy sample is categorized in the diamond cluster. We can also see samples in the diamond cluster have high membership degrees that represent the reliability of the cluster. Having no healthy sample in the diamond cluster generates a hypothesis that these samples must differ from the other cancer samples. 
\begin{table}[!ht]
\caption{Comparison of the diamond samples and the rest of cancer samples with regard to the covariant variables: five year survival (short/long), cancer stage (low/high), average age, and cancer type (squamous (s) / adenocarcinoma (a)).}
\label{table1}
\begin{center}
\setlength\tabcolsep{2.8pt}
\begin{tabular}{ | c | c | c | c | c | c | c | c | c |}
\hline
{\small \bf Serum metabolites} &  {\footnotesize \bf Samples} & {\footnotesize \bf
 Survival(s)} & {\footnotesize \bf Survival(l)} & {\footnotesize \bf Stage(l)} & {\footnotesize \bf Stage(h)} &  {\footnotesize \bf Age} &  {\footnotesize \bf Type(s)} & {\footnotesize \bf Type(a)}  \\
\hline
\hline 
 {\small \bf Diamond samples } & {\small 12} & {\small 63.30 \%} & {\small 36.70 \%} & {\small 83.33 \%}  & {\small 16.67 \%} & {\small 64.82} &  {\small 57.33 \% }& {\small 42.67 \%}  \\
\hline
 {\small \bf Cancer samples} & \multirow{2}{*}{\small 89} &  \multirow{2}{*}{\small 59.09 \%} &  \multirow{2}{*}{\small 40.91 \%} &  \multirow{2}{*}{\small 58.43 \%}  & \multirow{2}{*}{\small 41.57 \%} & \multirow{2}{*}{\small 66.08} &  \multirow{2}{*}{\small 44.94 \%} &  \multirow{2}{*}{\small 55.06 \%}  \\
 {\small \bf excluding diamond}  & & & & &  & & & \\
  \hline
   {\small \bf Normal samples} & {\small 29} &  - &  - &  -  &  - & {\small 66.83} &  - &  -  \\
 \hline
\end{tabular}
\end{center}
\end{table}
Results from analyzing covariates of individuals in the diamond cluster and comparing with the rest of cancer individuals are summarized in Table \ref{table1}. These individuals have shorter survival time and nearly 15 percent more with squamous cell carcinoma than adenocarcinoma. Very similar to the diamond cluster that includes no healthy individuals, the square cluster includes only three healthy individuals and the rest of 23 individuals have cancer. 
\begin{table}[!ht]
\caption{Analysing the square samples with regard to the covariant variables: five year survival (short/long), cancer stage (low/high), average age, smocked cigarette, and cancer type (squamous (s) / adenocarcinoma (a)).}
\label{table2}
\setlength\tabcolsep{1.4pt}
\begin{center}
\begin{tabular}{ | c | c | c | c | c | c | c | c | c | c |   }
\hline
{\small \bf Serum metabolites} &  {\footnotesize \bf Samples} & {\footnotesize \bf
 Survival(s)} & {\footnotesize \bf Survival(l)} & {\footnotesize \bf Stage(l)} & {\footnotesize \bf Stage(h)} &  {\footnotesize \bf Age} & {\footnotesize \bf Cigarette} & {\footnotesize \bf Type(s)} & {\footnotesize \bf Type(a)}  \\
\hline
\hline 
 {\small \bf Cancer samples in} & {\small 21} & {\small 40.91 \%} & {\small 59.09 \%} & {\small 60.87 \%}  & {\small 39.13 \%} & {\small 65.31} & {\small 65.72} & {\small 39.13 \%} & {\small 60.87 \%}  \\
 {\small \bf square cluster} & &  &  &   &  &  &  &  &   \\
\hline
 {\small \bf Cancer samples} & {\small 80} &  {\small 61.25 \%} &  {\small 38.75 \%} &  {v 59.75 \%}  & {\small 41.25 \%} & {\small 65.48} & {\small 56.44} & {\small 46.25 \%} &  {\small 53.75 \%}  \\
 {\small \bf excluding squares} &  &  &   &   &  &  &  & &   \\

  \hline
 {\small \bf All cancer samples} & {\small 101} &  {\small 60.40 \%} &  {\small 39.60 \%} &  {\small 61.39 \%}  & {\small 38.61 \%} & {\small 65.93} & {\small 60.43} & {\small 46.53 \%} &  {\small 53.47 \%}  \\
  \hline
   {\small \bf Normal samples in} & {\small 3} &  - &  - &  -  &  - & {\small 74.60} & {\small 48.71} & - &  -  \\
      {\small \bf square cluster} &  &   &   &   &  &  &  &  &    \\
 \hline
\end{tabular}
\end{center}
\end{table}
\begin{table}[!ht]
\caption{Circle cluster. Analysis of samples in the circle cluster shows that cancer individuals in this cluster are obviously younger that the healthy individuals in this cluster.}
\label{table3}
\setlength\tabcolsep{2.5pt}
\begin{center}
\begin{tabular}{ | c | c | c | c | c | c | c | c | c | }
\hline
{\small \bf{Serum metabolites}}&  {\footnotesize \bf{Samples}} & {\footnotesize \bf{Survival(s)}} & {\footnotesize \bf{Survival(l)}} & {\footnotesize \bf{Stage(l)}} & {\footnotesize \bf{Stage(h)}} &  {\footnotesize \bf{Age}} & {\footnotesize \bf{Type(s)}} & {\footnotesize \bf{Type(a)}}  \\
\hline
\hline 
 {\small \bf Cancer samples in} & \multirow{2}{*}{\small 6} & \multirow{2}{*}{\small 50.00 \%}  & \multirow{2}{*}{\small 50.00 \%} & \multirow{2}{*}{\small 66.66 \%}  & \multirow{2}{*}{\small 33.34 \%} & \multirow{2}{*}{\small 58.60} &  \multirow{2}{*}{\small 33.34 \%} & \multirow{2}{*}{\small 66.66 \%}  \\
 {\small \bf circle cluster}   & & & & & & & & \\
\hline
{\small \bf Healthy samples in} & \multirow{2}{*}{\small 11} & \multirow{2}{*}{-}  & \multirow{2}{*}{-} & \multirow{2}{*}{-} & \multirow{2}{*}{-} & \multirow{2}{*}{\small 64.65} &  \multirow{2}{*}{-} & \multirow{2}{*}{-} \\
 {\small \bf circle cluster}  & & & & & & & & \\
\hline
{\small \bf Healthy samples in circle} & \multirow{2}{*}{\small 9} & \multirow{2}{*}{-}  & \multirow{2}{*}{-} & \multirow{2}{*}{-} & \multirow{2}{*}{-} & \multirow{2}{*}{\small 69.35} &  \multirow{2}{*}{-} & \multirow{2}{*}{-} \\
{\small \bf without outliers} &  &   &  &  &  &  &   &  \\
\hline
  {\small \bf Cancer samples} & \multirow{2}{*}{\small 95} &  \multirow{2}{*}{\small 61.05 \%} & \multirow{2}{*}{\small 38.95 \%} &  \multirow{2}{*}{\small 61.05 \%}  & \multirow{2}{*}{\small 38.95 \%} & \multirow{2}{*}{\small 66.39} &  \multirow{2}{*}{\small 47.37 \%} &  \multirow{2}{*}{\small 52.63 \%}  \\
{\small \bf excluding circle}  & & & & & & & & \\
 \hline
  {\small \bf Cancer samples} & {\small 101} & {\small  60.40 \%} & {\small 39.60 \%} &  {\small 61.39 \%}  &  {\small 38.61 \%} & {\small 65.93} & {\small 46.53 \%} &  {\small 53.47 \%}  \\
   \hline
\end{tabular}
\end{center}
\end{table}
\hspace*{-3.5mm} Table \ref{table2} includes the analysis of individuals in the square cluster in terms of covariates. From Table \ref{table1} (analysis of cancer samples in the diamond cluster) and Table \ref{table2} (analysis of cancer samples in the square cluster), we see percentages of cancer types adenocarcinoma and squamous cell carcinoma are different in these two clusters. We may conclude that the serum metabolites of individuals with adenocarcinoma cancer is different from squamous cell carcinoma cancer and they are different from healthy metabolites. This means not only some differences between the serum metabolites of healthy and cancer individuals are observed, but also the serum metabolites of different types of cancers are different. On the other hand, only six out of 101 cancer samples are categorized in the circle cluster, where more than one third of the healthy samples are categorized. This generates a hypothesis that these six cancer samples have distinctive features from other cancer samples. Table \ref{table3} compares the circle samples with the remaining cancer samples. We can see these six individuals are almost 8 years younger than the rest of cancer samples and more than 66 percent are adenocarcinoma patients. They are almost 11 years younger than the healthy samples in the same cluster. We may conclude younger people with lung cancer have metabolites similar to older healthy people. According to the properties of data types, critical samples follow the patterns of more than one cluster fully or partially. Therefore, critical samples obtain membership degrees from more than one cluster. As a result, they are  distinguishably separated from the other samples in a cluster according to their behaviour with respect to both clusters, see Fig. \ref{BFPM-F2}(b) for a critical sample indicated by a black circle. Fig. \ref{BFPM-F2}(b) shows that the critical sample obtained very similar membership degrees in both clusters in comparison with the other serum samples. In Table \ref{table4}, we compare the critical sample with the other cancer samples in the same (square) cluster. In the analysis, we noticed that one sample, number 20 in Fig. \ref{BFPM-F2}(b) obtained almost similar membership degrees from both clusters. It has distinguishably a lower membership degree compared to the other samples in the cluster, Fig. \ref{BFPM-F2}(a). Critical objects can be chosen with a predefined threshold that varies by approaches and datasets. The threshold can be calculated based on the lower boundary of the assigned membership degrees to object with respect to all clusters. The critical object presented here is selected as an example. The advantage is to know how to evaluate and treat those samples with potential ability to move from healthy to unhealthy cluster or vice-versa. 
\begin{table}[!ht]
\caption{The critical sample. Information for the critical sample and the other samples in the same (square) cluster.}
\label{table4}
\setlength\tabcolsep{1.5pt}
\begin{center}
\begin{tabular}{ | c | c | c | c | c | c | c | c | c |   }
\hline
{\small \bf{Serum metabolites}} & {\footnotesize \bf {Samples}} & {\footnotesize \bf {Survival(s)}} & {\footnotesize \bf{Survival(l)}} & {\footnotesize \bf{Stage(l)}} & {\footnotesize \bf {Stage(h)}} & {\footnotesize \bf {Age}} & {\footnotesize \bf{Type(s)}} & {\footnotesize \bf{Type(a)} } \\
\hline
\hline 
 {\small \bf The critical sample} & {\small 1} & {\small +}  & - & {\small +}  & - & {\small 71.60}  & {\small +} & -  \\
\hline
 {\small \bf Cancer samples in} & \multirow{2}{*}{\small 21} & \multirow{2}{*}{\small 40.00 \%} & \multirow{2}{*}{\small 60.00 \%} & \multirow{2}{*}{\small 61.91 \%}  & \multirow{2}{*}{\small 38.09 \%} & \multirow{2}{*}{\small 65.31} &  \multirow{2}{*}{\small 42.86 \%} & \multirow{2}{*}{\small 57.14 \%}  \\
 {\small \bf square cluster} & & & & & & & & \\
\hline
 {\small \bf Cancer samples excluding} & \multirow{2}{*}{\small 20} & \multirow{2}{*}{\small 36.84 \%} & \multirow{2}{*}{\small 63.16 \%} & \multirow{2}{*}{\small 60.00 \%}  & \multirow{2}{*}{\small 40.00 \%} & \multirow{2}{*}{\small 64.99} &  \multirow{2}{*}{\small 40.00 \%} & \multirow{2}{*}{\small 60.00 \%}  \\
 {\small \bf the critical sample} & &  &  &   &  &  &   &   \\
\hline
 {\small \bf All cancer samples} & {\small 101} &  {\small 60.40 \%} &  {\small 39.60 \%} &  {\small 61.39 \%}  & {\small 38.61 \%} & {\small 65.93} &  {\small 46.53 \%} &  {\small 53.47 \%}  \\
  \hline
\end{tabular}
\end{center}
\end{table}
The BFPM clustering approach was used on the metabolomic case and control dataset to evaluate dissimilarity between healthy and cancer samples according to the stage of their diseases. Through categorizing the serum samples in four clusters in Fig. \ref{BFPM-F3}, using BFPM, it was observed that the diamond cluster includes only the cancer samples and not healthy sample. Through further analyses of these samples and comparing with the rest of cancer samples with regard to covariates, it was noticed that the diamond samples have severe cancer, shorter survival time, shorter life time, and more squamous cell carcinoma rather than adeno carcinoma cancer. It should be mentioned that the results presented in Tables \ref{table5} to \ref{table7} are related to the 101 tissue samples matched with the 101 serum samples. From the analysis of the pathology data depicted in Table \ref{table5}, it was observed that the diamond samples have more than 7 percent cancer cell as compared with the other cancer samples. However, in terms of necrosis cells, the  diamond samples have 50 percent necrosis cells less than other samples.
\begin{table}[!ht]
\caption{Pathology data analysis. A comparison between the pathology data of the diamond samples from Fig. \ref{BFPM-F3} and the rest of cancer samples. The diamond samples have more cancer cells and less than $50 \%$ necrosis cells as compared with the rest of the cancer samples. Note that the diamond samples are found with sever cancer through the serum analysis.}
\label{table5}
\setlength\tabcolsep{2.0pt}
\begin{center}
\begin{tabular}{ | c | c | c | c | c | c | c | c | }
\hline 
{\small \bf Tissue metabolites} & {\footnotesize \bf  Cancer }  & {\footnotesize \bf  Necrosis } & {\footnotesize \bf  Necrosis=0} & {\footnotesize \bf  Necrosis$> 10 $} & {\footnotesize \bf  Necrosis$> 25$}   & {\footnotesize \bf  Necrosis$> 50$ }  & {\footnotesize \bf  Benign } \\
\hline
\hline
 {\small \bf Diamond samples} & {\small 26.47 \%} & {\small 6.56 \%} & {\small 66.67 \%} & {\small 16.67 \%} & {\small 8.33 \%} & {\small 0.00 \%} & {\small 66.97 \%} \\
\hline
 {\small \bf Tissue samples} &  \multirow{2}{*}{\small 19.17 \%} &  \multirow{2}{*}{\small 12.52 \%} & \multirow{2}{*}{\small 66.34 \%} &  \multirow{2}{*}{\small 22.78 \%} &  \multirow{2}{*}{\small 14.85 \%} & \multirow{2}{*}{\small  9.90 \%} & \multirow{2}{*}{\small 67.32 \%} \\
  {\small \bf excluding diamond} &   &  &  &   &   &   & \\
\hline
\end{tabular}
\end{center}
\end{table}
\begin{table}[!ht]
\caption{Analyzing the diamond samples through their pathology data. Almost $86 \%$ of the diamond samples with short survival time are diagnosed at stage low and almost $60 \%$ have short survival time while through pathology they have less than $20 \%$ cancer cell.  }
\label{table6}
\setlength\tabcolsep{1.5pt}
\hspace*{-4mm}
\begin{center}
\begin{tabular}{ | c | c | c | c | c |   }
\hline 
{\small \bf Tissue cells}  & {\footnotesize \bf Survival(s)-Stage(l)} & {\footnotesize \bf  Survival(s)-Benign$>85$} &  {\footnotesize \bf  Survival(s)-Cancer$<20 $} & {\footnotesize \bf  Survival(s)-Cancer$<50 $} \\
\hline
\hline
{\small \bf Diamond}  & \multirow{2}{*}{\small 85.79 \%} &  \multirow{2}{*}{\small 42.86 \%}  & \multirow{2}{*}{\small 57.14 \%} & \multirow{2}{*}{\small 71.43 \%} \\
{\small \bf samples} & & & & \\
\hline
\end{tabular}
\end{center}
\end{table}
In Tables \ref{table6} and \ref{table7}, the pathology data of the diamond samples has been analyzed. From Table \ref{table6}, it was observed that almost 86 percent of the diamond samples with short survival time are diagnosed at the stage low and almost 60 percent have short survival time, while through pathology they have less than 20 percent cancer cell. From Table \ref{table7}, we see more than 45 percent of the diamond samples have less than 10 percent necrosis, while their survival is short. Therefore, from Tables \ref{table5} and \ref{table7}, we may conclude that the diamond samples are misdiagnosed through pathology analysis. However, from serum analysis, the diamond samples with severe cancer are separated from healthy individuals. In Table \ref{table8}, we look at the pathology of the critical sample. We can see that 100 percent cells of the sample are benign and therefore, the sample is diagnosed at stage low, while the survival time of the  critical sample is low. Table \ref{table9} provides some information related to the star samples, cancer and healthy samples in star, presented by Fig. \ref{BFPM-F3}. according to covariant variables: five year survival (short/long), cancer stage (low/high), average age, smoked cigarette, and cancer type (s/a).

\begin{table}[!ht]
\caption{Analysing the diamond samples through necrosis cells from the pathology data. More than $45 \%$ of the diamond samples have less than $10 \%$ necrosis while their survival is short.}
\label{table7}
\setlength\tabcolsep{0.5pt}
\begin{center}
\begin{tabular}{ | c | c | c | c | c |   }
\hline 
{\small \bf Tissue cells}  & {\footnotesize \bf Survival(s)-Necrosis=0} & {\footnotesize \bf Survival(s)-Necrosis$\leq 10 $} & {\footnotesize \bf Survival(s)-Necrosis$\leq 25 $} & {\footnotesize \bf Survival(s)-Necrosis$\leq 50 $} \\
\hline
\hline
{\small \bf Diamond}  & \multirow{2}{*}{\small 36.36 \%} &  \multirow{2}{*}{\small 45.45 \%}  & \multirow{2}{*}{\small 54.55 \%} & \multirow{2}{*}{\small 63.63 \%} \\
{\small \bf samples} & & & & \\
\hline
\end{tabular}
\end{center}
\end{table}
\begin{table}[!ht]
\caption{The pathology data of the critical sample and comparison with its cluster, square, and the rest of samples.}
\label{table8}
\setlength\tabcolsep{2.5pt}
\begin{center}
\begin{tabular}{ | c | c | c | c | c | c | c | c | }
\hline 
{\small \bf Serum metabolites} & {\footnotesize \bf Cancer }  & {\footnotesize \bf Necrosis } & {\footnotesize \bf Necrosis=0} & {\footnotesize \bf Necrosis$> 10$} & {\footnotesize \bf Necrosis$> 25$}   & {\footnotesize \bf Necrosis$> 50 $ }  & {\footnotesize \bf Benign } \\
\hline
\hline
 {\small \bf The critical sample}  & \multirow{2}{*}{\small 0.00 \%} & \multirow{2}{*}{\small 0.00 \%} & \multirow{2}{*}{\small -} & \multirow{2}{*}{\small -} & \multirow{2}{*}{\small -} & \multirow{2}{*}{\small -} & \multirow{2}{*}{\small 100.00 \%} \\
 {\small \bf in square cluster} & &  &  &  &  &  &  \\
\hline
 {\small \bf Tissue samples in} & \multirow{2}{*}{\small 7.33 \%} & \multirow{2}{*}{\small 14.69 \%} & \multirow{2}{*}{\small 65.57 \%} & \multirow{2}{*}{\small 17.40 \%} & \multirow{2}{*}{\small 17.40 \%} & \multirow{2}{*}{\small 13.04 \%} & \multirow{2}{*}{\small 73.45 \%} \\
  {\small \bf square cluster} &  &  &  &  &  &  &  \\
\hline
 {\small \bf Tissue samples} &  {\small 19.17 \%} &  {\small 12.52 \%} & {\small 66.34 \%} &  {\small 22.78 \%} &  {\small 14.85 \%} &  {\small 9.90 \%} & {\small 67.32 \%} \\
\hline
\end{tabular}
\end{center}
\end{table}
\begin{table}[!ht]
\caption{ {Analyzing the star samples with regard to the covariant variables: five year survival (short/long), cancer stage (low/high), average age, smocked cigarette, and cancer type (squamous (s) / adenocarcinoma (a)). }}
\label{table9}
\setlength\tabcolsep{2.0pt}
\begin{center}
\begin{tabular}{ | c | c | c | c | c | c | c | c | c | c |}
\hline
{\small \bf Serum metabolites} &  {\footnotesize \bf Samples} & {\footnotesize \bf
 Survival(s)} & {\footnotesize \bf Survival(l)} & {\footnotesize \bf Stage(l)} & {\footnotesize \bf Stage(h)} &  {\footnotesize \bf Age} & {\footnotesize \bf Cigarette} & {\footnotesize \bf Type(s)} & {\footnotesize \bf Type(a)}  \\
\hline
\hline 
 {\small \bf Star samples} & {\small 77} & -  & - & -  & - & {\small 66.72} & {\small 54.58} & - & -  \\
\hline
 {\small \bf Cancer samples in} & \multirow{2}{*}{\small 62} & \multirow{2}{*}{\small 58.02 \%} & \multirow{2}{*}{\small 41.98 \%} & \multirow{2}{*}{\small 59.68 \%}  & \multirow{2}{*}{\small 40.32 \%} & \multirow{2}{*}{\small 66.01} & \multirow{2}{*}{\small 57.29} & \multirow{2}{*}{\small 51.61 \%} & \multirow{2}{*}{\small 48.39 \%}  \\
  {\small \bf star cluster} &  &  &  &   &  &  &  &  &   \\
\hline
 {\small \bf Healthy samples in} & \multirow{2}{*}{\small 15} &  \multirow{2}{*}{\small -}  &  \multirow{2}{*}{\small -} &  \multirow{2}{*}{\small -}  & \multirow{2}{*}{\scriptsize -} & \multirow{2}{*}{\small 69.45} & \multirow{2}{*}{\small 43.35} & \multirow{2}{*}{\small -} & \multirow{2}{*}{\small -}  \\
  {\small \bf star cluster} &  &    &  &   & &  &  &  &   \\
  \hline
\end{tabular}
\end{center}
\end{table}
%
\hspace*{-7.5mm} As a result, Bounded Fuzzy Possibilistic Method (BFPM) provides an opportunity to identify some differences between serum of healthy and lung cancer samples. Using BFPM, we can also evaluate the pathology. The method recognizes the critical samples in prevention strategies for those samples that are going to move to another cluster. The findings strengthened the hypothesis that there are some differences between serum metabolites of healthy and cancer samples.
%
%
\subsection{BFPM in Security, Banking Systems, and Risk Managements}
BFPM has been introduced to provide the most flexible search space in
partitioning problems in membership assignments to not only evaluate the current status of objects, but also to study their future movements from their own category to others. Objects' movement analysis makes the systems capable of keeping their environments safe and secure by applying prediction and prevention strategies at the same time. In banking, security, and financial systems, where any negligence leads to irreparable consequences, analysing the ability of any transactions (packets or activities) of harming the reputation of the organizations is the most suitable and reasonable solution. In risk management systems, we also need to know whether the objects are normal or risky in addition to have a better insight into what extent objects can be risky in the near future in order to keep the systems and the assets in a safe side. Security is the main issue for real time systems, where analysing the received packets should be handled on-line, otherwise the security of the systems cannot be guaranteed. To do so, BFPM has been applied on some datasets related to security and banking systems to evaluate the behaviour of objects.\\
The main goal of these experiments were to firstly, find out the harmful objects in the systems, and secondly, to extract the potential abilities of objects to move from their own class (cluster) to another. The method overcomes the issues with restrictions and limitations for samples in their freely participation in other clusters and classes. This property of the method helps to distinguish the objects that are willing to hide their properties to harm the systems. As mentioned earlier, by manipulating some features of each individual packets (transactions or objects), the methods might be trapped and consider that particular object in a wrong cluster, where in security and financial systems this misclassification or miss-assignment results in severe consequences. In crucial systems such as financial systems customers are expecting more from the organizations and the system cannot easily categorize them in a wrong category. In fact, these crucial systems need to evaluate the objects (customers) in the most accurate ways. On the other hand, paying less attention to the effects of risky objects and attacks leads to irreparable consequences. As a result, a comprehensive evaluation is needed to keep the customers happy and safe and also keep the systems more secure and safe.\\
BFPM with its properties allows the crucial systems consider the mentioned points in a good manner. BFPM evaluates the behaviour of objects by considering the history of the objects with respect to now and the future. This evaluation helps to detect the critical objects and shows how far the objects might cause damages to the systems. BFPM have been applied in both supervised and unsupervised methodologies to evaluate the ability of each object individually with respect to all classes or clusters. Two different algorithms introduced for classification and clustering concepts to cover the evaluation of risk management and both types of cyber attacks detection categories: misused-based and anomaly-based detection techniques, where the classification approach is processed by the BFPCM algorithm, and the clustering approach is handled by the BFPM algorithm.\\
BFPM also evaluates objects' movement (mutation) analysis in details by allowing objects to participate in much clusters and classes as they can. This condition, instead of restricting objects to participate in just one category, allows objects to show their potential abilities to participate in other categories. This facility clarifies in what extent on-line transactions (packets) can be risky (harmful) for the systems in the near future. The following sections pay more attention with more details in objects movement analysis in security, banking, and risk management systems. The functionality of the proposed method (BFPM) will be compared with fuzzy (probability) methods in objects' movement analysis to mathematically prove how BFPM is capable of providing such analysis.\\ 
%
\section{Experiments on Classification and Prediction Strategies}
Risk managements are taken into consideration, where the optimal systems have faced several risks on transactions (packets), reputation, information security, price, liquidity, foreign exchange, compliance risks, and so on. More specifically, default estimation, the amount of money that customers have difficulties to give back to the organizations, is considered by prediction systems by using different probability models to assess the risk, where default probabilities (PD), loss given default (LGD), exposure at default (EAD), and effective maturity for groups of homogeneous assets have been utilized to measure the risks alongside the historical data analysis. For instance, evaluating very safe assets based on their heuristic data using some models leads to severe consequences. Under uncertain conditions, probability distributions have been applied in default probability to predict the occurrence of default in the future. Default is also accompanied by the interest rate spikes and recessions which brings the attentions to more global parameters regarding the risk analysis.\\
In general, uncertainty has been covered by some methodologies while the probabilistic risk analysis (PRA), quantitative risk analysis (QRA), or probabilistic safety analysis (PSA) has been widely applied in risk assessments and risk managements. As a result, machine learning can help decision making, risk management, and financial systems to deal with uncertainty. Different learning methods, either supervised or unsupervised, have been utilized in uncertain conditions with different levels of accuracy and functionalities. Probability and fuzzy methods have been proposed to consider the likelihood or the level of happing of events with this condition $(\sum_{i=1}^c u_{ij} = 1)$ indicating that individuals $j^{th}$ can obtain partial degrees from categories $i^{th}$ in intervals: $(0 \leq u_{ij} \leq 1)$. Fuzzy (probability) condition allows the methods to cover uncertainty by providing more flexibilities in individuals analysis with some limitations.\\
Bounded Fuzzy Possibilistic Method (BFPM) allows the method to evaluate and study the behaviour of individuals with respect to both categories (risky and non-risky). BFPM presents new condition as $(0 < \frac{1}{c} \sum_{i=1}^c u_{ij} \leq 1)$. The condition proves that each individual can obtain degrees of one from all possible categories. This condition allows individuals to present their potential abilities in their participations in other categories which is very important for financial, decision making, and risk management systems. The BFPM methodology has been applied as a supervised method to evaluate whether the objects (customers) are risky or not. The proposed method aims to provide a relaxed search space to cover prevention and prediction techniques at the same time. In this experiment, the ability of the proposed method in classification and prediction strategies has been evaluated by this Thesis on a dataset from a large bank \cite{hundred-thirty-seven} as clustering is the unsupervised classification of patterns \cite{hundred-thirty-eight}. The aim of this experiment is to know how BFPM on membership assignments can be applied in other learning methodologies. In other words, the Thesis aims to know how allowing objects to participate in others classes, or solutions might affect the final results.\\
BFPM also allows to have different levels of risk categories which consequently assists learning methods to evaluate individuals with their real properties and assign different level of risks to individuals. One of the main and the most interesting achievement of this experiment will be a new view on how moving from crisp and fuzzy membership assignments to more suitable search space might increase the accuracy of the methods. For this purpose, the selected dataset from the international bank with more than one thousand branches has been chosen to evaluate the ability of the proposed method in risk assessment by considering supervised techniques. Due to the large amount of data, regarding corporate loans from the bank, just small amount of corporate loans from ten randomly selected branches of the bank has been considered for this experiment. Total cases of  corporate loans from these branches were almost 140000 samples, which was gathered for the period of 15 years from 1995-2010. From the available samples, just the samples from the last year have been considered in this experiment without considering the previous history of customers. One way of evaluating the credit risk is based on the analysis of historical data from customers which lead to get a better insight on customers' reaction in the near future. But, according to the economical issues and conditions, historical data might skew the results.\\
As a result, the samples of the last year including 16000 observations have been selected. Unfortunately, there was no possibility to evaluate all the samples owing to null and missing values. After applying preprocessing techniques to extract the useful dataset for precise evaluation, just $1486$ objects or examples have been considered suitable for this experiment, in which $1366$ objects were recognized as one class label. There are different techniques that are proposed to work with data, which pre-processing and feature selection approaches are some of them. This Thesis did not pay much attention to these techniques as these topics are far from the scope of this Thesis, but in general providing the most suitable dataset and feature sets should be well considered by learning methods. To obtain an acceptable result the dataset has been balanced, which led to $360$ objects. There is no possibility to trust the result without balancing the data. If the number of objects from one class label is much greater than the number of objects from other class labels, then the accuracy of the method is very high even if the proposed method falsely labels all objects by the major class label. In this case, the accuracy would be more than $85$ percent if we just work on unbalanced data and choose just the major class label for all objects in the testing dataset. The original objects had $23$ features and one of the features recognized as an objective function or class label. Five features that have less effects on the objective function have been removed.\\
Consequently, each object has $18$ features in which  $14$ features have discrete values and the rest have continuous values. The data has been used to introduce a new intelligent method to learn from the training dataset to classify a set of objects into categories whether customers would default on loan payments or not. Due to the fact that not normalized data may cause some non-realistic results to calculate the distance between objects, data objects are normalized into an interval $[0,1]$. In classification, the most direct measure of accuracy is misclassification error: the number of incorrectly classified objects (records) divided by the total number of objects. As a result of applying BFPCM on the datasets to predict the default, it was concluded that the accuracy of the algorithm is 99.82, or the error ratio is 0.18. This accuracy was obtained without considering the historical data analysis. In some cases, the historical data analysis might lead to better outcomes, while there is no economical crisis, otherwise the historical data analysis might skew the learning algorithms. The accuracy of BFPCM is better than the decision tree algorithm, which obtained 96.00 accuracy \cite{hundred-thirty-seven}.\\
%
%
BFPM can be applied in classification and prediction problems according to the presented experiment. BFPM performs better than the other classification approaches, which might use crisp or fuzzy membership assignments. This experiment brightened the path of applying BFPM on other learning strategies. The idea allows to use more flexible search space for membership assignments instead of restricting data objects to participate in less number of solutions and clusters. Allowing objects to participate in more clusters and solutions are the main ideas of distinguishing the critical objects and areas. The experiments show how BFPM methodology in different methods, either unsupervised or others, on different domains (medicine, security, risk management, and decision making systems) can lead to extracting useful information. The achievements are accompanied by the mutation analysis which results in analysing the behaviour of objects (individuals) in the near future which makes the systems more secure. Using BFPM methodology not only results in promising outcomes, but also facilitates the simultaneous procedure of prevention and prediction strategies with regard to cut the extra cost for further analysis and assessments.  \\
%
%
%
%
%
%
%
\section{Measuring Accuracy by Validity Indices}
As mentioned in chapter \ref{Measure-Chap}, different validity indices have been proposed to evaluate the accuracy of the clustering algorithms by measuring the compactness and separations of objects in intra-clusters and inter-clusters respectively. As BFPM has been applied in both types of problems, supervised and unsupervised, the Thesis aims to also check the functionalities of BFPM with regard to clustering problems. The Thesis evaluated the accuracy of the proposed method (BFPM) using some validity indices. Table \ref{Cluster Validity Functions} presents the results from validity functions: Partition Coefficient $(V_{pc})$, Partition Entropy $(V_{pe})$, and Xie-Beni $(V_{xb})$ obtained by BFPM, FCM, and wFCM algorithms on "Iris" and "Pima" datasets. Promising results from Table \ref{Cluster Validity Functions} and the previous sections indicate that BFPM leads to desirable outcomes. Table \ref{Val_clus_No} explores the results from different validity functions $V_{pc}$, $V_{pe}$, $DB$, $G$, and $CS$ for BFPM with respect to different values of fuzzification constant $m$. \\


%
\begin{table}[!h]
\caption{Results of clustering validity functions $ V_{pc}$, $V_{pe}$, and $V_{xb}$ for FCM, wFCM \cite{three}, and BFPM algorithms on Iris and Pima datasets.}
\label{Cluster Validity Functions}
\vspace{-4mm}
\begin{center}
\begin{tabular}{ | c | c | c | c | c | c | }
\hline
{\bf Dataset} & {\bf Method} & $\bf V_{pc} \uparrow$ & $\bf V_{pe}  \downarrow$ & $\bf V_{xb} \downarrow$ &  {\bf No. Clusters}  \\
\hline
\hline
& FCM  & 0.78 & 0.57 & 0.13 &   \\
{\bf Iris} & wFCM  & 0.86  & 0.36& {\bf 0.05} &  3 \\
& {\bf BFPM}  & {\bf 1.24}  & {\bf 0.23} & 0.20   &  \\
\hline
 & FCM  & 0.82 &0.43 & 0.12&  \\
{\bf Pima} & wFCM  & 0.87 &0.31 & {\bf 0.08}&  2\\
 & {\bf BFPM}  & {\bf 1.62}  & {\bf 0.08} & 0.11 &  \\
\hline
\end{tabular}
\end{center}
\end{table}

%
%
%
\begin{table}[!h]
\caption{Accuracy of BFPM algorithm, using $V_{pc}$, $V_{pe}$, $DB$, $G$, and $CS$ indices as fitness functions for normalized Iris and Pima datasets for different values of fuzzification constant $(m)$. }
\label{Val_clus_No}
\vspace{-4mm}
%
%
\begin{center}
\begin{tabular}{ | c | c | c | p{1cm} | p{1cm} | p{1cm} | p{1cm} |   }
\hline
{\footnotesize \bf Validity Index} & {\footnotesize \bf DataSet} & { \footnotesize \bf Cluster No.} &  {\footnotesize \bf m=1.4} & { \footnotesize \bf m=1.6} & { \footnotesize \bf m=1.8} & {\footnotesize \bf m =2} \\
\hline
\hline
\multirow{2}{*}{\textbf{$V_{pc}\uparrow$}}		
		&Iris	& 3 &   {\bf 0.76} & 0.74 & 0.73 &  0.74\\	
	\cline{3-7}
		&  Pima		& 2 &   {\bf 0.75} & 0.66 & 0.57& 0.57\\	
		\hline
		\hline
\multirow{2}{*}{\textbf{$V_{pe}  \downarrow$}}		
		& Iris 	& 3 &   0.33 & 0.27 & 0.23 & {\bf 0.20} \\	
		\cline{3-7}
		& Pima 		& 2 &  0.30 & 0.30 & 0.29& {\bf 0.26} \\	
\hline
\hline
\multirow{2}{*}{\textbf{$DB  \downarrow$}}
		&  Iris	& 3    & 0.30  & 0.24  & 0.23 & {\bf 0.22}  \\	
	\cline{3-7}
		& 	Pima	& 2  & 2.97 & 1.95  & 1.74  & {\bf 1.66}  \\
\hline
\hline
\multirow{2}{*}{\textbf{$G \uparrow$}}
	&  Iris	& 3  & 5.30 & 7.88  & 10.05 & {\bf 12.14}  \\	
		\cline{3-7}
		& 	Pima	& 2  & 1.44 & 1.83 & 2.10 & {\bf 2.23}   \\
\hline
\hline		
\multirow{2}{*}{\textbf{$CS  \downarrow$ }}
		&  	Iris	& 3  & 0.05 & {\bf 0.04}  & {\bf 0.04}  & {\bf 0.04} \\	
		\cline{3-7}
		& Pima		& 2  & 0.05 & {\bf 0.03} & {\bf 0.03} & {\bf 0.03}    \\
\hline
\end{tabular}
\end{center}
\end{table}

\hspace*{-7.5mm} With regard to the results, it is clear that the larger and smaller values of fuzzification constant leads to different outcomes, which should be precisely considered by the methods. It should be also noted that the concept of the Thesis for objects' movement analysis has some arguments for separation analysis, because the methods cannot precisely analyse the behaviour of objects in their participation in other clusters if the objects are completely separated.
\section{Complexity Analysis}
To evaluate the functionality of learning methods and algorithms different factors have been considered such as computational complexity alongside the good performance. In this section the Thesis aims to evaluate the computational complexity of the proposed algorithms to check whether the proposed algorithms are worthy enough in comparison with other centroid-based methods. Table \ref{complexity} shows the computational costs for each one of the algorithms presented in this Thesis, where the complexity of the algorithms strongly depends on the number of iterations according to reach the stopping conditions, but in general the table presents information regarding the complexity of functions in learning procedures with regard to CPU and memory usages called as computational complexity and space complexity respectively.\\
Computational complexity of BFPM is $O(it  \cdot  n  \cdot   c^{2}  \cdot d)$, where $it$ is the number of iterations, $n$ is the number of objects, $c$ is the number of clusters, and d is the number of dimensions or features. FPM-I complexity is $O(it \cdot n \cdot c^2 \cdot d + it \cdot c \cdot d)$ as in each iteration one more membership assignment with regard to features is calculated. FPM-II has less complexity in comparison with FPM-I as the extra membership assignments in each iteration reduced to just one assignment. FPM-II complexity is $O(it \cdot n \cdot c^2 \cdot d + c \cdot d)$. Regarding the weighted feature distance functions applied in BFPM, the complexity is increased in comparison with BFPM by multiplication of weights to all features in each iteration, so BFPM-WFD complexity is $O(it \cdot n \cdot c^2 \cdot 2d )$. In conclusion, BFPM has less computational complexity in comparison with other algorithms. According to space complexity, almost all of the proposed algorithms have the same complexity as $O( n \cdot d + n \cdot c)$ to store $n$ objects in $d$ dimensional search space and store memberships of $n$ objects with respect to $c$ clusters. But the space complexity of FPM-I is different as $O( n \cdot d + n \cdot 2c)$, while it needs to store one more membership for each object with respect to all clusters. The Thesis does not consider the space for FPM-II for extra memory for the second membership assignment as it can be calculated at run time without allocating any memory space.\\
According to the table and the information from other centroid-based clustering methods, it is clear that the cost of using BFPM is very reasonable and applying BFPM on other domains and other datasets will be beneficial if we also take into the consideration the performance of BFPM methodology in learning procedures.\\

\begin{table}[!ht]
\caption{Computational complexity and space complexity for the proposed algorithms FPM-I, FPM-II, BFPM, and BFPM-WFD in the Thesis, where $it$ is the number of iterations, $n$ is the number of objects, $d$ is the number of dimensions or the number of features, and $c$ is the number of clusters.}
\label{complexity}
\vspace{-4mm}
\begin{center}
\begin{tabular}{ | c | c | c |  }
\hline
{ \bf Method}  & {\bf Computational Complexity}  & {\bf Space Complexity} \\
\hline
\hline
{\footnotesize {\bf FPM-I}} 		& $O(it \cdot n \cdot c^2 \cdot d + it \cdot c \cdot d)$ & $O( n \cdot d + n \cdot 2c)$ \\
\hline
{\footnotesize {\bf FPM-II}} 		& $O(it \cdot n \cdot c^2 \cdot d + c \cdot d )$     & $O( n \cdot d + n \cdot c)$ \\
\hline
{\footnotesize {\bf BFPM}} 	 	&     $O(it \cdot n \cdot c^2 \cdot d)$ &  $O( n \cdot d + n \cdot c)$ \\
\hline
{\footnotesize {\bf BFPM-WFD}} 	 	&  $O(it \cdot n \cdot c^2 \cdot 2d )$ &  $O( n \cdot d + n \cdot c)$  \\
\hline
\end{tabular}
\end{center}
\end{table}

%
%
\chapter{Analysis of Obtained Results}
\label{Analysis-Chap}
This Thesis introduced a new method in learning procedures that can be applied in different types of methods (supervised, unsupervised, ...) in addition to propose new sets of similarity and membership functions to obtain better results in comparison with other methods. Introducing a new type of object "Critical" and a new type of feature "Dominant" by the Thesis provided a flexible environment for objects' movements analysis that can be simultaneously utilized for prevention and prediction strategies, besides partitioning purposes. The new types of object and feature lead to better accuracy by clarifying the causes of miss-assignments by learning methods. The proposed method and functions have been applied on different disciplines with remarkable achievements that are briefly concluded in follows.  \\
\begin{itemize}
\item {\bf Mathematically Proof of the Proposed Method:} \\[0.1cm]
The proposed method and similarity functions are mathematically proved in the Thesis as supersets of other related learning methods and similarity functions. The Thesis also presents the causes of miss-assignments by mathematically proposing the new type of object "Critical" and the new type of feature "Dominant". In addition, the Thesis  mathematically describes how to detect critical objects and handle the impact of dominant features in learning procedures. 
\item {\bf Membership Analysis:} \\[0.1cm]
BFPM is introduced to provide the most flexible search space for data objects in comparison with all crisp, probability, fuzzy, and possibilistic methods to allow the objects to obtain partial or full memberships of more, even all clusters (classes). The flexible search space results in having better insight in prevention, prediction, partitioning, and mutation analysis at the same time without redoing the learning procedures for many times. This property of BFPM is very crucial for Big-Data analysis as redoing the learner on huge amount of data is very costly and sometimes impossible. 
\item {\bf Similarity Analysis:}\\[0.1cm]
Weighted Feature Distance (WFD) and Prioritized Weighted Feature Distance (PWFD), as the supersets of other related similarity functions with respect to different norms, have been proposed and discussed in this Thesis. The proposed similarity functions not only cover the diversity in both the feature and the vector spaces, but also handle the impact of features on the final results of the methods. According to the results of BFPM-WFD in comparison with BFPM, it is clear that WFD leads to obtaining better accuracy in addition to handle the impact of features.
\item {\bf Diversity Analysis:}\\[0.1cm]
The Thesis proposed different algorithms (FPM-I, FPM-II, BFPM, and BFPM-WFD) by applying the proposed methodology and similarity functions to cover diversity either in the feature and the vector spaces or by using the most relaxed search space proposed by BFPM. Based on the results of BFPM and other methods that make use of different similarity functions in different norms (fuzzification constant), it is concluded that BFPM works better than the other methods with respect to their functionalities.  
\item {\bf Analysis on Impact of Dominant Features:}\\[0.1cm]
The new type of feature, "Dominant", introduced in this Thesis might skew the results of similarity functions, which consequently affects the accuracy of the methods. Weighted Feature Distance (WFD) is introduced to handle the impact of dominant features on similarity procedures. According to the results of BFPM and BFPM-WFD in comparison with the Euclidean distance function, it is concluded that BFPM-WFD can handle the impact of dominant features by assigning different weights to features to increase or decrease the influences of dominant features in the desirable directions.
\item {\bf Analysis on Critical Objects and Areas:}\\[0.1cm]
The critical object has been introduced as a new type of object in this Thesis to emphasize the necessity of considering and treating objects based on their behaviour. This type of object fits to more than two patterns of data with ability to move from one category to another, which can be interpreted either as causes of miss-assignments in learning procedure or some valuable objects that can lead to better achievements in some applications and systems such as medicine, security, risk management, and so on. The Thesis also presented a new taxonomy for data types by categorizing objects into Normal, Outlier, and Critical objects. According to the compared results and the figures on provided examples and datasets, none of the learning methods considered how to distinguish and also assign the proper memberships to critical objects. BFPM is able to distinguish and assign the proper memberships to critical objects. Knowing the exact type of objects in advance prevents the systems from further costs and unwanted consequences. 
\item {\bf Mutation Analysis:}\\[0.1cm]
BFPM is introduced to provide a flexible environment to monitor, track, and control the objects' movements in advance. This facilitates the procedure of providing the safe and secure environments for crucial systems, where any objects' movements might lead to irreparable consequences. The Thesis also compared BFPM with other  methods on their ability to study the mutation of data objects beforehand. The results and the presented figures show that BFPM is able to monitor the ability of data objects on their movements from one cluster or class to another by getting changes in their feature spaces, while other methods did not consider this strategy in their learning procedures. 
\item {\bf Applying the Method on Different Disciplines} \\[0.1cm]
The proposed methodology has been applied in different clustering, classification problems in different domains and disciplines such as finance, medicine, risk management, image processing, and anomaly detection systems. Promising results proved that the proposed method perform well in those domains, which consequently prove that BFPM methodology is capable of extracting knowledge from other disciplines and domains, and also being applied on different types of learning strategies. \\
\end{itemize}
%
%
\vspace*{-10mm}
\chapter{Conclusions}
\label{Conclusion-Chap}
\vspace*{-12mm}
The Thesis introduced a new comprehensive learning method that overcomes the drawbacks of the conventional methods that neglect the impact of critical objects and dominant features by introducing new membership and similarity functions. The Thesis declared and mathematically identified the critical object and dominant features that are not considered by the other methods. This Thesis introduced the Bounded Fuzzy Possibilistic Method (BFPM) as a new learning method and membership function. BFPM not only avoids decreasing the membership degrees assigned to data objects with respect to clusters, but also makes the search space wider for data objects to participate in more even all clusters as partial or full members. \\
BFPM relaxes the conditions provided by the other membership functions, besides defining upper and lower boundaries in their assignments to remove all the limitations and also to simplify the procedure of implementation in partitioning methods. Therefore, the BFPM method does not accept any null values for memberships with respect to the clusters, in contrast to possibilistic methods. Based on the definition of the proposed method, BFPM, and the proofs, it was concluded that all the crisp, fuzzy, probability, and possibilistic sets are the subsets of the BFPM set, as: 
\begin{equation}
\nonumber
\label{subsets-of-BFPM}
M_{hcn}\subset M_{fcn} \subset M_{pcn} \subset M_{bfpm}.
\end{equation}
Different types of data objects were explored in this Thesis. The importance of considering the type of objects in learning and mining algorithms has been mathematically proved by the Thesis. The Thesis also introduced a new type of data objects, {\it Critical}, that plays the main role in datasets. The new type of object follows the data models and obeys the rules by data patterns. Critical objects are able to affect the results of any learning and mining method as they can participate in more even all clusters (classes) as full or partial members. Critical objects can easily move from one cluster to another by getting mutation or small changes in their feature spaces. Removing or decreasing their membership degrees make the systems very vulnerable. The mutation analysis and critical objects (areas) analysis, as some of the most important factors of learning procedures, are proposed by this Thesis.\\
The Thesis explored some examples to demonstrate the issues that may happen by wrong membership assignments on critical objects. The Thesis also compares the accuracy of the most well-known partitioning methods in different domains and disciplines to evaluate the ability of the proposed methods on both partitioning results and dealing with critical objects. The Thesis also compares the functionalities of BFPM with fuzzy (probability) methods as well as other fuzzy and possibilistic methods on providing suitable search spaces for critical objects. BFPM allows investigators to study data objects' movements in most crucial systems and applications in advance, where any mutation should be handled and maintained beforehand.\\
A new type of feature, {\it Dominant}, has been introduced in this Thesis. Almost all of the similarity functions that perform in the vector space and most of the similarity functions that perform in both the feature and vector spaces are being trapped by dominant features. Dominant features can massively affect the final results in any learning algorithm. Similarity functions, the issues with similarity functions, and handling the impact of dominant features on similarity functions and learning methods are another concepts that have been explored in this Thesis. The Thesis also introduced Weighted Feature Distance (WFD) and Prioritized Weighted Feature Distance (PWFD) to cover diversity in the feature spaces as well as the vector space. Covering diversity in both the vector and the feature spaces prevent miss-assignments or misclassification by learning methods that might happen by dominant features. In other words, in WFD and PWFD the impact of dominant features on the final results has been precisely handled.\\
Some similarity functions have been compared with Weighted Feature Distance (WFD) function in their functionality on dominant features. The proposed method and functions are compared on different domains and disciplines on clustering problems. Promising results show that the proposed method and functions work better than the other algorithms. The results also prove that WFD makes learning algorithms more stable to deal with dominant features. Prioritized weighted feature distance (PWFD) has been proposed to demonstrate how features can skew the results by obtaining the right or wrong weights. The proposed method, similarity functions, and algorithms have been applied on different datasets from different domains and disciplines such as medicine, security, physic, risk management, anomaly detection, images, video, economy, finance, and so on with respect to the well consideration of critical objects and dominant features. The promising results by BFPM proved that the proposed method is capable of providing the most flexible environment for data objects and learning algorithms in supervised and unsupervised strategies. \\
In conclusion, BFPM is introduced to overcome the issues with other methods in membership assignments. WFD is proposed to cover diversity in search spaces, and also prevents learning algorithms of being influenced by dominant features. Critical objects are defined and studied to predict the future movements of data objects. As a result, BFPM and BFPM-WFD lead to better accuracy in comparison with other fuzzy and possibilistic methods in their membership assignments. WFD performs better than the other discussed functions on handling the impact of dominant features. In brief the Thesis covered the following concepts:
\vspace*{-2mm}
\begin{itemize}
\item  introducing a new method, Bounded Fuzzy Possibilistic Method (BFPM);
\vspace*{-2mm}
\item  introducing a new set of similarity functions Weighted Feature Distance (WFD(s)) and Prioritized Feature Distance (PWFD(s)) in different norms;
\vspace*{-2mm}
\item  introducing a new type of data object (critical);
\vspace*{-2mm}
\item  introducing the dominant features and handling the impact of such features to cover diversity in both the feature and the vector spaces;
\vspace*{-2mm}
\item  providing the most flexible environment for mutation's analysis;
\vspace*{-2mm}
\item comparing the accuracy of the other methods with BFPM, in terms of fuzzification constant, different similarity functions, dealing with critical objects, uncertainty, and mutation. 
\end{itemize}
As a result, the Thesis found the proper answer for the hypothesis, whether some advanced parameters affect the accuracy of the methods. The parameters that affect the accuracy of the methods and cause misclassification and miss-assignments in learning approaches have been presented in the Thesis. The analysis and the experiments have shown that critical objects and dominant features are very crucial in learning methods, membership functions, and similarity functions that have been neglected by most of other methods. Precise processing of critical objects and dominant features using the proper membership and similarity functions prevents further costs and leads to obtaining better results either in description or prediction strategies.\\[0.2cm]
The better achievements have been obtained by the new method (BFPM) and algorithms proposed in this Thesis with regard to accuracy, providing the most flexible membership assignments, and covering diversity in both the vector and the feature spaces. Due to the new learning method proposed in this Thesis, the accuracy of partitioning methods has been improved.

%
\addcontentsline{toc}{chapter}{Bibliography}
\bibliography{biblio}

\begin{thebibliography}{100}

\bibitem{fifteen}
F.~Yang, T.~Sun, and C.~Zhang, ``An efficient hybrid data clustering method
  based on k-harmonic means and particle swarm optimization,'' {\em Expert
  Systems with Applications, vol. 36, no. 6, pp. 9847-9852}, 2009.

\bibitem{eight}
J.~Han, J.~Pei, and M.~Kamber, {\em Data mining: concepts and techniques}.
\newblock Elsevier, 2011.

\bibitem{one}
I.~H. Witten and E.~Frank, {\em Data mining practical machine learning tools
  and techniques}.
\newblock Elsevier, Morgan Kaufmann, 2016.

\bibitem{fifty-seven}
X.~Wang, Y.~Wang, and L.~Wang, ``Improving fuzzy c-means clustering based on
  feature-weight learning,'' {\em Pattern Recognition Letters, vol. 25, no. 10,
  pp. 1123-1132}, 2004.

\bibitem{three}
C.~C. Aggarwa, {\em Outlier analysis}.
\newblock Kluwer Academic Publishers, 2004.

\bibitem{two}
P.~N. Tan, M.~Steinbach, and V.~Kumar, {\em Introduction to data mining}.
\newblock Pearson Addison Wesley, 2014.

\bibitem{four}
C.~Borgelt, {\em Prototype-based classification and clustering}.
\newblock doctoral dissertation, Otto-von-Guericke-Universitat Magdeburg, 2005.

\bibitem{five}
J.~C. Bezdek and N.~R. Pal, ``Some new indexes of cluster validity,'' {\em
  IEEE, Transactions on Systems, Man, and Cybernetics - Part B: Cybernetics,
  vol. 28, no. 3, pp. 301-315}, 1998.

\bibitem{six}
O.~Linda and M.~Manic, ``General type-2 fuzzy c-means algorithm for uncertain
  fuzzy clustering,'' {\em IEEE, Transactions on Fuzzy Systems, vol. 20, no. 5,
  pp. 883-897}, 2012.

\bibitem{seven}
R.~Xu and D.~Wunsch, {\em Clustering}.
\newblock IEEE Press Series on Computational Intelligence, 2009.

\bibitem{nine}
H.~Yazdani, ``Bounded fuzzy possibilistic method,'' {\em Elsevier, Fuzzy Sets
  and Systems, vol. 389, pp. 51-65}, 2020.

\bibitem{hundred-six}
H.~Yazdani and K.~Choros, ``Intrusion detection and risk evaluation in online
  transactions using partitioning methods,'' {\em Springer, International
  Conference on Multimedia $\&$ Network Information Systems, vol. 833, pp.
  190-200}, 2018.

\bibitem{hundred-ten}
H.~Yazdani, D.~Ortiz-Arroyo, and H.~Kwasnicka, ``New similarity functions,''
  {\em IEEE, International Conference on Artificial Intelligence and Pattern
  Recognition, pp. 47-52}, 2016.

\bibitem{hundred-eleven}
H.~Yazdani and H.~Kwasnicka, ``Issues on critical objects in mining
  algorithms,'' {\em IEEE, International Conference on Artificial Intelligence
  and Pattern Recognition, pp. 53-58}, 2016.

\bibitem{hundred-twelve}
H.~Yazdani, L.~Cheng, D.~C. Christiani, and A.~Yazdani, ``Bounded fuzzy
  possibilistic method reveals information about lung cancer through analysis
  of metabolomics,'' {\em IEEE, Transactions on Computational Biology and
  Bioinformatics, vol. 17, no. 2, pp. 526-535}, 2020.

\bibitem{hundred-thirteen}
H.~Yazdani, ``Fuzzy possibilistic on different search spaces,'' {\em IEEE,
  International Symposium on Computational Intelligence and Informatics, pp.
  283-288}, 2016.

\bibitem{hundred-fourteen}
H.~Yazdani, D.~Ortiz-Arroyo, K.~Choros, and H.~Kwasnicka, ``On high dimensional
  searching space and learning methods,'' {\em Springer Verlag, Data Science
  and Big Data: An Environment of Computational Intelligence, pp. 29-48}, 2016.

\bibitem{hundred-fifteen}
H.~Yazdani, D.~Ortiz-Arroyo, K.~Choros, and H.~Kwasnicka, ``Applying bounded
  fuzzy possibilistic method on critical objects,'' {\em IEEE, International
  Symposium on Computational Intelligence and Informatics, pp. 271-276}, 2016.

\bibitem{hundred-sixteen}
H.~Yazdani, H.~Kwasnicka, and D.~Ortiz-Arroyo, ``Multiobjective particle swarm
  optimization using fuzzy logic,'' {\em Springer, International Conference on
  Computational Collective Intelligence, Lecture Notes in Computer Science,
  vol. 6922, pp. 224-233}, 2011.

\bibitem{hundred-seventeen}
H.~Yazdani and H.~Kwasnicka, ``Fuzzy classification method in credit risk,''
  {\em Springer, International Conference on Computational Collective
  Intelligence, Lecture Notes in Computer Science, vol. 7653, pp. 495-505},
  2012.

\bibitem{hundred-thirty-six}
H.~Yazdani and K.~Choros, ``Comparative analysis of accuracy of fuzzy
  clustering methods applied for image processing,'' {\em Springer,
  International Conference on Multimedia $\&$ Network Information Systems, vol.
  833, pp. 89-99}, 2018.

\bibitem{ten}
W.~W.~Pedrycz, V.~Loia, and S.~Senatore, ``Fuzzy clustering with viewpoints,''
  {\em IEEE, Transactions on Fuzzy Systems, vol. 18, no. 2, pp. 274-284}, 2010.

\bibitem{eleven}
H.~E. Lee, K.~H. Park, and Z.~Z. Bien, ``Iterative fuzzy clustering algorithm
  with supervision to construct probabilistic fuzzy rule base from numerical
  data,'' {\em IEEE, Transactions on Fuzzy Systems, vol. 16, no. 1, pp.
  263-277}, 2008.

\bibitem{twelve}
L.~A. Zadeh, ``Fuzzy sets,'' {\em Information and Control, vol. 8. no. 3, pp.
  338-353}, 1965.

\bibitem{thirteen}
R.~Krishnapuram and J.~M. Keller, ``A possibilistic approach to clustering,''
  {\em IEEE, Transactions on Fuzzy System, vol. 1, no. 2, pp. 98-110}, 1993.

\bibitem{fourteen}
F.~Hoppner, {\em Fuzzy cluster analysis: Methods for classification, data
  analysis and image recognition}.
\newblock John Wiley $\&$ Sons, 1999.

\bibitem{sixteen}
T.~M. Mitchell, {\em Machine learning}.
\newblock Mc-Graw Hill, 1997.

\bibitem{seventeen}
J.~Zhou, C.~L.~P. Chen, L.~Chen, and H.~X. Li, ``A collaborative fuzzy
  clustering algorithm in distributed network environments,'' {\em IEEE,
  Transactions on Fuzzy Systems, vol. 22, no. 6, pp. 1443-1456}, 2014.

\bibitem{eighteen}
H.~P. Kriegel, P.~P.~Kroger, and A.~Zimek, ``Clustering high-dimensional data:
  A survey on subspace clustering, pattern-based clustering, and correlation
  clustering,'' {\em ACM, Transactions on Knowledge Discovery from Data, vol.
  3, no. 1, pp. 1-58}, 2009.

\bibitem{nineteen}
R.~Xu and D.~Wunsch, ``Survey of clustering algorithms,'' {\em IEEE,
  Transactions on Neural Networks, vol. 16, no. 3, pp. 645-678}, 2005.

\bibitem{twenty}
D.~Vanisri, ``Spatial bias correction based on gaussian kernel fuzzy c-means in
  clustering,'' {\em International Journal of Computer Science and Network
  Solutions, vol. 2, no. 12, pp. 1-8}, 2014.

\bibitem{twenty-one}
A.~Fahad, N.~Alshatri, A.~Tari, A.~Alamri, I.~Khalil, A.~Y. Zomaya, S.~Foufou,
  and A.~Bouras, ``A survey of clustering algorithms for big data: Taxonomy and
  empirical analysis,'' {\em IEEE, Transactions on Emerging Topics in
  Computing, vol. 2, no. 3, pp. 267-279}, 2014.

\bibitem{twenty-two}
S.~Sardari and M.~Eftekhari, ``A fuzzy decision tree approach for imbalanced
  data classification,'' {\em IEEE, International Conference on Computer and
  Knowledge Engineering, pp. 292-297}, 2016.

\bibitem{twenty-three}
P.~Liu and H.~Li, ``Efficient learning algorithms for three-layer regular
  feedforward fuzzy neural networks,'' {\em IEEE, Transactions on Neural
  Networks, vol. 15, no. 3, pp. 545-558}, 2004.

\bibitem{twenty-four}
A.~G. Bakirtzis, J.~B. Theocharis, S.~J. Kiartzis, and K.~J. Satsios, ``Short
  term load forecasting using fuzzy neural networks,'' {\em IEEE, Transactions
  on Power Systems, vol. 10, no. 3, pp. 1518-1524}, 1995.

\bibitem{twenty-five}
L.~A. Zadeh, ``Fuzzy sets as a basis for a theory of possibility,'' {\em
  Elsevier, Fuzzy Sets and Systems, vol. 100, pp. 9-34}, 1999.

\bibitem{twenty-six}
G.~Chen and T.~T. Pham, {\em Introduction to fuzzy sets, fuzzy logic and fuzzy
  control systems}.
\newblock CRC Press, 2000.

\bibitem{twenty-seven}
L.~Yu, K.~K. Lai, S.~Wang, and L.~Zhou, ``A least squares fuzzy svm approach to
  credit risk assessment,'' {\em Fuzzy Information and Engineering, pp.
  865-874}, 2007.

\bibitem{twenty-eight}
G.~Carlsson and F.~M'emoli, ``Characterization, stability and convergence of
  hierarchical clustering methods,'' {\em Journal of Machine Learning Research,
  vol. 11, pp. 1425-1470}, 2010.

\bibitem{twenty-nine}
L.~A. Zadeh, ``Toward extended fuzzy logic - a first step,'' {\em Elsevier,
  Fuzzy Sets and Systems, vol. 160, no. 21, pp. 3175-3181}, 2009.

\bibitem{thirty}
J.~J. Kli and B.~Yuan, {\em Fuzzy sets and fuzzy logic theory and
  applications}.
\newblock Printice Hall PTR, 1995.

\bibitem{thirty-one}
M.~Russo, ``Genetic fuzzy learning,'' {\em IEEE, Transactions on Evolutionary
  Computation, vol. 4, no. 3, pp. 259-273}, 2000.

\bibitem{thirty-two}
A.~P. Engelbrecht, {\em Computational intelligence: An introduction}.
\newblock John Wiley $\&$ Sons, 2007.

\bibitem{thirty-three}
W.~Sheng, S.~Swift, L.~Zhang, and X.~Liu, ``A weighted sum validity function
  for clustering with a hybrid niching genetic algorithm,'' {\em IEEE,
  Transactions on Systems, Man, and Cybernetics - Part B (Cybernetics), vol.
  35, no. 6, pp. 1156-1167}, 2005.

\bibitem{thirty-four}
L.~Bottolo and S.~Richardsony, ``Evolutionary stochastic search for bayesian
  model exploration,'' {\em International Society for Bayesian Analysis, vol.
  5, no. 3, pp. 583-618}, 2010.

\bibitem{thirty-five}
O.~Poleshchuk and E.~Komarov, ``A fuzzy linear regression model for interval
  type-2 fuzzy sets,'' {\em IEEE, Annual Meeting of the North American Fuzzy
  Information in Processing of the Society, pp. 1-5}, 2012.

\bibitem{thirty-six}
A.~A. Marquez, F.~A. Marquez, and A.~Peregrin, ``An efficient multi-objective
  evolutionary adaptive conjunction for high dimensional problems in linguistic
  fuzzy modelling,'' {\em IEEE, International Conference on Fuzzy Systems, pp.
  1-8}, 2012.

\bibitem{thirty-seven}
J.~Wang, G.~Shan, X.~Duan, and B.~Wen, ``Improved svm-rfe feature selection
  method for multi-svm classifier,'' {\em IEEE, International Conference on
  Electrical and Control Engineering, pp. 1592-1595}, 2011.

\bibitem{thirty-eight}
C.~Olaru and L.~Wehenkel, ``A complete fuzzy decision tree technique,'' {\em
  Elsevier, Fuzzy Sets and Systems, vol. 138, no. 2, pp. 221-254}, 2003.

\bibitem{thirty-nine}
F.~Hoppner and F.~Klawonn, ``Improved fuzzy partitions for fuzzy regression
  models,'' {\em International Journal of Approximate Reasoning, vol. 32, pp.
  85-102}, 2003.

\bibitem{forty}
S.~Piramuthu, ``Financial credit-risk evaluation with neural and neurofuzzy
  systems,'' {\em European Journal of Operational Research, vol. 112, no. 2,
  pp. 310-321}, 1997.

\bibitem{forty-one}
P.~Hajek, D.~Harmancova, and R.~Verbrugge, ``A qualitative fuzzy possibilistic
  logic,'' {\em International Journal of Approximate Reasoning, vol. 12, no. 1,
  pp. 1-19}, 1995.

\bibitem{forty-two}
V.~Torra, ``Fuzzy c-means for fuzzy hierarchical clustering,'' {\em IEEE,
  International Conference on Fuzzy Systems, pp. 646-651}, 2005.

\bibitem{forty-three}
H.~Ishibuchi and Y.~Nojima, ``Analysis of interpretability-accuracy tradeoff of
  fuzzy systems by multiobjective fuzzy genetics-based machine learning,'' {\em
  Elsevier, International Journal of Approximate Reasoning, vol. 44, no. 1, pp.
  4-31}, 2007.

\bibitem{forty-four}
I.~Inza, P.~Larranaga, R.~Etxeberria, and B.~Sierra, ``Feature subset selection
  by bayesian network-based optimization,'' {\em Artificial Intelligence, vol.
  123, no. 1-2, pp. 157-184}, 2000.

\bibitem{forty-five}
C.~Hwang and F.~C.~H. Rhee, ``Uncertain fuzzy clustering: Interval type-2 fuzzy
  approach to c-means,'' {\em IEEE, Transactions on Fuzzy Systems, vol. 15, no.
  1, pp. 107-120}, 2007.

\bibitem{forty-six}
T.~C. Havens, J.~C. Bezdek, C.~Leckie, L.~O. Hall, and M.~Palaniswami, ``Fuzzy
  c-means algorithms for very large data,'' {\em IEEE, Transactions on Fuzzy
  Systems, vol. 20, no. 6, pp. 1130-1146}, 2012.

\bibitem{forty-seven}
J.~Wu, H.~Xiong, and J.~Chen, ``Adapting the right measures for k-means
  clustering,'' {\em Knowledge Discovery Data Mining, pp. 877-886}, 2009.

\bibitem{forty-eight}
M.~S. Yang and H.~S. Tsai, ``A gaussian kernel-based fuzzy c-means algorithm
  with a spatial bias correction,'' {\em Pattern Recognition Letters, vol. 29,
  no. 12, pp. 1713-1725}, 2008.

\bibitem{forty-nine}
M.~Y. Su, C.~Y. Lin, S.~W. Chien, and H.~C. Hsu, ``Genetic-fuzzy association
  rules for network intrusion detection systems,'' {\em IEEE, International
  Conference on Fuzzy Systems, pp. 2046-2052}, 2011.

\bibitem{fifty}
V.~Chandola, A.~Banerjee, and V.~Kumar, ``Anomaly detection: A survey,'' {\em
  ACM, Computing Surveys, vol. 41, no. 3, p. 15}, 2009.

\bibitem{fifty-one}
C.~Cooper and A.~Frieze, ``A general model of web graphs,'' {\em Algorithms,
  vol. 22, no. 3, pp. 311-335}, 2003.

\bibitem{fifty-two}
R.~Agrawal, R.~Srikant, ``Mining sequential patterns,'' {\em IEEE,
  International Conference on Data Engineering, vol. 95, pp. 3-14}, 1995.

\bibitem{fifty-three}
A.~Agresti, {\em An introduction to categorical data analysis}.
\newblock John Wiley and Sons, 1996.

\bibitem{fifty-four}
V.~Barnett and T.~Lewis, {\em Outliers in statistical data}.
\newblock John Wiley and Sons, 1994.

\bibitem{fifty-five}
A.~Kannan, G.~Q. Maguire, A.~Sharma, and P.~Schoo, ``Genetic algorithm based
  feature selection algorithm for effective intrusion detection in cloud
  networks,'' {\em IEEE, International Conference on Data Mining Workshops, pp.
  416-423}, 2012.

\bibitem{fifty-six}
Y.~Yu and H.~Wu, ``Anomaly intrusion detection based upon data mining
  techniques and fuzzy logic,'' {\em IEEE, International Conference on Systems,
  Man and Cybernetics, pp. 514-517}, 2012.

\bibitem{fifty-eight}
F.~Critchely, ``On a framework for dissimilarity analysis,'' {\em Springer,
  Data Analysis: Scientific Modelling and Practical Application, pp. 121-134},
  2000.

\bibitem{fifty-nine}
P.~Cunningham, ``A taxonomy of similarity mechanisms for case-based
  reasoning,'' {\em IEEE, Transactions on Knowledge and Data Engineering, vol.
  21, no. 11, pp. 1532-1543}, 2009.

\bibitem{sixty}
A.~E. Sarlyuce, B.~Gedik, G.~Jacques-Silva, K.~L. Wu, and U.~V. Catalyurek,
  ``Sonic: streaming overlapping community detection,'' {\em Data Mining
  Knowledge Discovery, vol. 30, no. 4, pp. 819-847}, 2016.

\bibitem{sixty-one}
J.~Xie, S.~Kelley, and B.~Szymanski, ``Overlapping community detection in
  networks: The state-of-the-art and comparative study,'' {\em ACM, Computer
  Survey, vol. 45, no. 4, pp. 1-43}, 2013.

\bibitem{sixty-two}
S.~Das, S.~Chaudhuri, and A.~K. Das, ``Cluster analysis for overlapping
  clusters using genetic algorithm,'' {\em IEEE, International Conference on
  Computational Intelligence and Communication Networks, pp. 6-11}, 2016.

\bibitem{sixty-three}
J.~Whang, D.~Gleich, and I.~Dhillon, ``Overlapping community detection using
  seed set expansion,'' {\em ACM, International Conference on Information and
  Knowledge Management, vol. 13, pp. 2099-2108}, 2013.

\bibitem{sixty-four}
G.~O. Campos, A.~Zimek, J.~Sander, R.~J. G.~B. Campello, B.~Micenkova,
  E.~Schubert, I.~Assent, and M.~E. Houle, ``On the evaluation of unsupervised
  outlier detection: measures, datasets, and an empirical study,'' {\em Data
  Mining Knowledge Discovery, vol. 30, no. 4, pp. 891-927}, 2016.

\bibitem{sixty-five}
H.~D. Nguyen and Q.~Cheng, ``An efficient feature selection method for
  distributed cyber attack detection and classification,'' {\em IEEE,
  Conference on Information Sciences and Systems, pp. 1-6}, 2011.

\bibitem{sixty-six}
J.~C. Dunn, ``A fuzzy relative of the isodata process and its use in detecting
  compact well-separated clusters,'' {\em pp. 32-57}, 1973.

\bibitem{sixty-seven}
J.~Bezdek, {\em Pattern recognition with fuzzy objective function algorithm}.
\newblock Springer, Science $\&$ Business Media, 2013.

\bibitem{sixty-eight}
H.~C. Huang, Y.~Y. Chuang, and C.~S. Chen, ``Multiple kernel fuzzy
  clustering,'' {\em IEEE, Transactions on Fuzzy Systems, vol. 20, no. 1, pp.
  120-134}, 2012.

\bibitem{sixty-nine}
S.~Eschrich, J.~Ke, L.~O. Hall, and D.~B. Goldgof, ``Fast accurate fuzzy
  clustering through data reduction,'' {\em IEEE, Transactions on Fuzzy
  Systems, vol. 11, no. 2, pp. 262-270}, 2003.

\bibitem{seventy}
R.~J. Hathawaya and J.~C. Bezdek, ``Extending fuzzy and probabilistic
  clustering to very large data sets,'' {\em Computational Statistics $\&$ Data
  Analysis, vol. 51, no. 1, pp. 215-234}, 2006.

\bibitem{seventy-one}
B.~Kosko, ``Fuzziness vs. probability,'' {\em International Journal of General
  Systems, vol. 17, no. 2-3, pp. 211-240}, 1990.

\bibitem{seventy-two}
X.~Wu, V.~Kumar, J.~R. Quinlan, J.~Ghosh, Q.~Yang, H.~Motoda, G.~J. McLachlan,
  A.~Ng, B.~Liu, P.~S. Yu, Z.~H. Zhou, D.~J. Steinbach, . nd~Hand, and
  D.~Steinberg, ``Top 10 algorithms in data mining,'' {\em Knowledge and
  Information Systems, vol. 14, no. 1, pp. 1-37}, 2008.

\bibitem{seventy-three}
R.~L. Cannon, J.~V. Dave, and J.~C. Bezdek, ``Efficient implementation of the
  fuzzy c-means clustering algorithms,'' {\em IEEE, Transactions on Pattern
  Analysis and Machine Intelligence, vol. PAMI-8, no. 2, pp. 248-255}, 1986.

\bibitem{seventy-four}
R.~Xu and D.~Wunsch, ``Recent advances in cluster analysis,'' {\em
  International Journal of Intelligent Computing and Cybernetics, vol. 1, no.
  4, pp. 484-508}, 2008.

\bibitem{seventy-five}
H.~K. Kwan and Y.~Cai, ``A fuzzy neural network and its application to pattern
  recognition,'' {\em IEEE, Transactions on Fuzzy Systems, vol. 2, no. 3, pp.
  185-193}, 1994.

\bibitem{seventy-six}
B.~Taskar, E.~Segal, and D.~Koller, ``Probabilistic classification and
  clustering in relational data,'' {\em International Joint Conference on
  Artificial Intelligence, vol. 17, no. 1, pp. 870-878}, 2001.

\bibitem{seventy-seven}
N.~R. Pal, K.~Pal, J.~M. Keller, and J.~C. Bezdek, ``A possibilistic fuzzy
  c-means clustering algorithm,'' {\em IEEE, Transactions on Fuzzy Systems,
  vol. 13, no. 4, pp. 517-530}, 2005.

\bibitem{seventy-eight}
D.~T. Anderson, J.~C. Bezdek, M.~Popescu, and J.~M. Keller, ``Comparing fuzzy,
  probabilistic, and possibilistic partitions,'' {\em IEEE, Transactions on
  Fuzzy Systems, vol. 18, no. 5, pp. 906-918}, 2010.

\bibitem{seventy-nine}
T.~C. Havens, R.~Chitta, A.~K. Jain, and R.~Jin, ``Speed-up of fuzzy and
  possibilistic kernel c-means for large-scale clustering,'' {\em IEEE,
  International Conference on Fuzzy Systems, pp. 463-470}, 2011.

\bibitem{eighty}
M.~H.~F. Zarandi, M.~Zarinbal, and I.~B. Turksen, ``Type-ii fuzzy possibilistic
  c-mean clustering,'' {\em IFSA-EUSFLAT Conference, pp. 30-35}, 2009.

\bibitem{eighty-one}
M.~Barni, V.~Cappellini, and A.~Mecoccie, ``Comments on a possibilistic
  approach to clustering,'' {\em IEEE, Transaction on Fuzzy Systems, vol. 4,
  no. 3, pp. 393-396}, 1996.

\bibitem{eighty-two}
S.~D. Xenaki, K.~D. Koutroumbas, and A.~A. Rontogiannis, ``A novel adaptive
  possibilistic clustering algorithm,'' {\em IEEE, Transactions on Fuzzy
  Systems, vol. 24, no. 4, pp. 791-810}, 2016.

\bibitem{eighty-three}
M.~S. Yang and C.~Y. Lai, ``A robust automatic merging possibilistic clustering
  method,'' {\em IEEE, Transactions on Fuzzy Systems, vol. 19, no. 1, pp.
  26-41}, 2011.

\bibitem{eighty-four}
F.~Masulli and S.~Rovetta, ``Soft transition from probabilistic to
  possibilistic fuzzy clustering,'' {\em IEEE, Transactions on Fuzzy Systems,
  vol. 14, no. 4, pp. 516-527}, 2006.

\bibitem{eighty-five}
K.~Honda, H.~Ichihashi, A.~Notsu, F.~Masulli, and S.~Rovetta, ``Several
  formulations for graded possibilistic approach to fuzzy clustering,'' {\em
  Springer, Lecture Notes in Computing, pp. 939-948}, 2006.

\bibitem{eighty-six}
U.~L. Altintakan, A.~Yazici, and M.~Koyuncu, ``A novel fuzzy visual object
  classification approach,'' {\em IEEE, International Conference on Fuzzy
  Systems, pp. 1-6}, 2012.

\bibitem{eighty-seven}
X.~Chen, X.~Li, B.~Ma, and P.~M.~B. Vitanyi, ``The similarity metric,'' {\em
  IEEE, Transactions on Information Theory, vol. 50, no. 12, pp. 3250-3264},
  2004.

\bibitem{eighty-eight}
S.~H. Cha, ``Comprehensive survey on distance/similarity measures between
  probability density functions,'' {\em International Journal of Mathematical
  Models and Methods in Applied Sciences, vol. 4, no. 1, pp. 1}, 2007.

\bibitem{eighty-nine}
M.~Deza and E.~Deza, {\em Encyclopaedia of distances}.
\newblock Springer, 2014.

\bibitem{ninety}
J.~Z. Huang, M.~K. Ng, H.~Rong, and Z.~Li, ``Automated variable weighting in
  k-means type clustering,'' {\em IEEE, Transaction on Pattern Analysis and
  Machine Intelligence, vol. 27, no. 5, pp. 657-668}, 2005.

\bibitem{ninety-one}
L.~Jing, M.~K. Ng, and J.~Z. Huang, ``An entropy weighting k-means algorithm
  for subspace clustering of high-dimensional sparse data,'' {\em IEEE,
  Transaction on Knowledge and Data Engineering, vol. 19, no. 8, pp.
  1026-1041}, 2007.

\bibitem{ninety-two}
L.~Wang, C.~Leckie, R.~Kotagiri, and J.~Bezdek, ``Approximate pairwise
  clustering for large data sets via sampling plus extension,'' {\em Pattern
  Recognition, vol. 44, no. 2, pp. 222-235}, 2011.

\bibitem{ninety-three}
J.~Neyman, ``Contributions to the theory of the $x^2$ test,'' {\em Berkeley
  Symposium on Mathematical Statistics and Probability, vol. 1, pp. 239-273},
  1949.

\bibitem{ninety-four}
D.~G. Gavin, W.~W. Oswald, E.~R. Wahl, and J.~W. Williams, ``A statistical
  approach to evaluating distance metrics and analog assignments for pollen
  records,'' {\em Quaternary Research, vol. 60, no. 3, pp. 356-367}, 2003.

\bibitem{ninety-five}
T.~F. Cox and M.~A.~A. Cox, {\em Multidimensional scaling}.
\newblock Chapman and Hall/CRC 2nd Edition, 2001.

\bibitem{ninety-six}
E.~F. Krause, {\em Taxicab geometry an adventure in non-Euclidean geometry}.
\newblock Dover, 1987.

\bibitem{ninety-seven}
D.~M.~J. Tax, R.~Duin, and D.~de~Ridder, {\em Classification, parameter
  estimation, and state estimation: An engineering approach using MATLAB}.
\newblock John Wiley and Sons, 2017.

\bibitem{ninety-eight}
J.~Looman and J.~B. Campbell, ``Adaptation of sorensen's k for estimating unit
  affinities in prairie vegetation,'' {\em Ecology, vol. 41, no. 3, pp.
  409-416}, 1960.

\bibitem{ninety-nine}
J.~C. Gower, ``A general coefficient of similarity and some of its
  properties,'' {\em Biometrics, vol. 27, no. 4, pp. 857-871}, 1971.

\bibitem{hundred}
V.~Monev, ``Introduction to similarity searching in chemistry,'' {\em
  Communications in Mathematical and Computer Chemistry, vol. 51, pp. 7-38},
  2004.

\bibitem{hundred-one}
P.~Kumar and A.~Johnson, ``On a symmetric divergence measure and information
  inequalities,'' {\em Journal of Inequalities in Pure and Applied Mathematics,
  vol. 6, no. 3}, 2005.

\bibitem{hundred-two}
D.~S. Modha and W.~S. Spangler, ``Feature weighting in k-means clustering,''
  {\em Machine Learning, vol. 52, no. 3, pp. 217-237}, 2003.

\bibitem{hundred-three}
S.~Salzberg, ``A nearest hyperrectangle learning method,'' {\em Machine
  Learning, vol. 6, no. 3, pp. 251-276}, 1991.

\bibitem{hundred-four}
G.~Strang, {\em Introduction to linear algebra}.
\newblock Wellesley-Cambridge Press, 2014.

\bibitem{hundred-five}
D.~J. Weller-Fahy, B.~J. Borghetti, and A.~A. Sodemann, ``A survey of distance
  and similarity measures used within network intrusion anomaly detection,''
  {\em IEEE, Communication Surveys and Tutorials, vol. 17, no. 1, pp. 70-91},
  2015.

\bibitem{hundred-seven}
L.~Parsons, E.~Haque, and H.~Liu, ``Subspace clustering for high dimensional
  data: a review,'' {\em ACM, Sigkdd Explorations Newsletter, vol. 6, no. 1,
  pp. 90-105}, 2004.

\bibitem{hundred-eight}
H.~P. Kriegel, P.~Kroger, and A.~Zimek, ``Clustering high-dimensional data: A
  survey on subspace clustering, pattern-based clustering, and correlation
  clustering,'' {\em ACM, Transactions on Knowledge Discovery from Data, vol.
  3, no. 1, pp. 1}, 2009.

\bibitem{hundred-nine}
C.~W. Tsai, C.~F. Lai, H.~C. Chao, and A.~V. Vasilakos, ``Big data analytics: A
  survey,'' {\em Journal of Big Data, vol. 2, no. 1, pp. 21}, 2015.

\bibitem{hundred-eighteen}
D.~W. Aha, D.~Kibler, and M.~K. Albert, ``Instance-based learning algorithms,''
  {\em Machine Learning, vol. 6, no. 1, pp. 37-66}, 1991.

\bibitem{hundred-nineteen}
R.~Nisbet, J.~Elder, and G.~Miner, {\em Statistical analysis and data mining
  applications}.
\newblock Elsevier, 2009.

\bibitem{hundred-twenty}
T.~Hastie, R.~Tibshirani, and J.~Friedman, ``The elements of statistical
  learning: data mining, inference, and prediction,'' {\em Journal of the Royal
  Statistical Society: Series A (Statistics in Society), vol. 173, no. 3, pp.
  693-694}, 2010.

\bibitem{hundred-twenty-one}
R.~J. G.~B. Campello, ``A fuzzy extension of the rand index and other related
  indexes for clustering and classification assessment,'' {\em Pattern
  Recognition Letters, vol. 28, no. 7, pp. 833-841}, 2007.

\bibitem{hundred-twenty-two}
C.~Chou, M.~Su, and E.~Lai, ``A new cluster validity measure and its
  application to image compression,'' {\em Pattern Analysis and Applications,
  vol. 7, no. 2, pp. 205-220}, 2004.

\bibitem{hundred-twenty-three}
M.~R. Rezaee, B.~P.~F. Lelieveldt, and J.~H.~C. Reiber, ``A new cluster
  validity index for the fuzzy c-mean,'' {\em Pattern Recognition Letters, vol.
  19, no. 3, pp. 237-246}, 1998.

\bibitem{hundred-twenty-four}
D.~L. Davies and D.~W. Bouldin, ``A cluster separation measure,'' {\em IEEE,
  Transactions on Pattern Analysis and Machine Intelligence, vol. 2, pp.
  224-227}, 1979.

\bibitem{hundred-twenty-five}
R.~XU, J.~Xu, and D.~C. Wunsch, ``A comparison study of validity indices on
  swarm-intelligence-based clustering,'' {\em IEEE, Transaction on Systems, Man
  and Cybernetics, vol. 42, no. 4, pp. 1243-1256}, 2012.

\bibitem{hundred-twenty-six}
N.~S. Rhee and K.~W. Oh, ``A validity measure for fuzzy clustering and its use
  in selecting optimal number of clusters,'' {\em IEEE, International
  Conference on Fuzzy Systems, vol. 2, pp. 1020-1025}, 1996.

\bibitem{hundred-twenty-seven}
A.~Asuncion and D.~Newman, ``Uci machine learning repository,'' 2007.

\bibitem{hundred-thirty-one}
H.~Parvin and B.~Minaei-Bidgoli, ``A clustering ensemble framework based on
  selection of fuzzy weighted clusters in a locally adaptive clustering
  algorithm,'' {\em Springer, Pattern Analysis and Applications, vol. 18, no.
  1, pp. 87-112}, 2015.

\bibitem{hundred-thirty-two}
D.~Graves and W.~Pedrycz, ``Kernel-based fuzzy clustering and fuzzy clustering:
  A comparative experimental study,'' {\em Elsevier, Fuzzy Sets and Systems,
  vol. 161, no. 4, pp. 522-543}, 2010.

\bibitem{hundred-thirty-three}
R.~C. de~Amorim, ``An empirical evaluation of different initializations on the
  number of k-means iterations,'' {\em Springer Verlag, International
  Conference on Artificial Intelligence, pp. 15-26}, 2012.

\bibitem{hundred-thirty-four}
S.~P. Chatzis and G.~Tsechpenakis, ``A possibilistic clustering approach toward
  generative mixture models,'' {\em Elsevier, Pattern Recognition, vol. 45, no.
  2, pp. 1819-1825}, 2012.

\bibitem{hundred-thirty-five}
A.~Lensen, B.~Xue, and M.~Zhang, ``Particle swarm optimisation representations
  for simultaneous clustering and feature selection,'' {\em IEEE, Symposium
  series in Computational Intelligence, pp. 1-8}, 2016.

\bibitem{hundred-twenty-eight}
Q.~Song, J.~Ni, and G.~Wang, ``A fast clustering-based feature subset selection
  algorithm for high dimensional data,'' {\em IEEE, Transactions on Knowledge
  and Data Engineering, vol. 25, no. 1, pp. 1-14}, 2013.

\bibitem{hundred-twenty-nine}
C.~Alzate and J.~A.~K. Suykens, ``Multiway spectral clustering with
  out-of-sample extensions through weighted kernel pca,'' {\em IEEE,
  Transactions on Pattern Analysis and Machine Intelligence, vol. 32, no. 2,
  pp. 335-347}, 2010.

\bibitem{hundred-thirty}
I.~J. Sledge, J.~C. Bezdek, T.~C. Havens, and J.~M. Keller, ``Relational
  generalizations of cluster validity indices,'' {\em IEEE, Transactions on
  Fuzzy Systems, vol. 18, no. 4, pp. 771-786}, 2010.

\bibitem{hundred-thirty-seven}
Z.~Yazdani, M.~M. Sepehri, and B.~Teimourpour, {\em Selecting best features for
  predicting bank loan default}, pp.~229--245.
\newblock Elsevier, 2014.

\bibitem{hundred-thirty-eight}
A.~K. Jain, M.~N. Murty, and P.~J. Flynn, ``Data clustering a review,'' {\em
  ACM, Computing Surveys, vol. 31, no. 3, pp. 264-323}, 1999.

\end{thebibliography}
\bibliographystyle{ieeetr}
\newpage
{\small 
{\bf \Large{Appendix A:}}\\[0.3cm]
{\bf \Large{Abbreviations}}\\[0.3cm]
\parbox{3cm}{ACC} Accuracy \\%
\parbox{3cm}{AM-PCM}  \parbox{10cm}{Automatic Merging Possibilistic Clustering Methods} \\%
\parbox{3cm}{ANN}  Artificial Neural Network\\%
\parbox{3cm}{APCM} Adaptive Possibilistic C-Means\\%
\parbox{3cm}{AVG} Average\\%
%
\parbox{3cm}{BFPM}  Bounded Fuzzy Possibilistic Method\\%
%
%
\parbox{3cm}{CD-FCM} Collaborative Distributed Fuzzy C-Means\\%
\parbox{3cm}{CART} Classification And Regression Tree\\%
%
%
%
\parbox{3cm}{DBSCAN} Density-Based Spatial Clustering of Applications with Noise\\%
\parbox{3cm}{DENCLUE} DENsity based CLUstEring\\%
\parbox{3cm}{D-Medoid} Dynamic Medoid \\%
\parbox{3cm}{DOM} Document Object Model\\
\parbox{3cm}{DS-Centroid} Dataset Seeded Centroid\\%
\parbox{3cm}{DTree} Decision Tree\\%
%
%
%
%
%
\parbox{3cm}{ERR}  Error\\%
%
%
%
\parbox{3cm}{FCM}  Fuzzy C-Means \\%
\parbox{3cm}{FWLAC} Fuzzy Weighted Locally Adaptive Clustering\\%
%
%
%
\parbox{3cm}{$H \& W$} Hartigan and Wong\\%
\parbox{3cm}{HTML} HyperText Markup Language\\%
%
%
\parbox{3cm}{ID3} Iterative Dichotomiser 3\\%
\parbox{3cm}{IDS} Intrusion Detection System\\%
\parbox{3cm}{IPS} Intrusion Prevention System\\ %
\parbox{3cm}{ISP} Internet Service Provider\\%
%
%
%
%
\parbox{3cm}{KCD-FCM} Kernel-based Collaborative Distributed Fuzzy C-Means\\%
\parbox{3cm}{KFCM-F} Kernel-based Fuzzy C-Means and Fuzzy clustering\\%
\parbox{3cm}{KMS-Centroid} K-means Seeded Centroid\\%
%
%
%
\parbox{3cm}{LAC} Locally Adaptive Clustering\\%
%
%
%
\parbox{3cm}{MoG} Mixtures of Gaussian\\%
%
%
\parbox{3cm}{NGE}  Nested Generalized Examplar\\%
%
%
\parbox{3cm}{OPTICS} Ordering Points to Identify the Clustering Structure\\%
%
%
\parbox{3cm}{PAM} Partition Around Medoids\\%
\parbox{3cm}{PCM} Possibilistic Clustering Method\\%
\parbox{3cm}{PFCM} Possibilistic Fuzzy Clustering Method\\%
\parbox{3cm}{PMoG} Possibilistic Mixtures of Gaussian\\%
\parbox{3cm}{PWFD}  Prioritized Feature Distance Function\\%
%
%
%
%
%
\parbox{3cm}{SONIC} Streaming Overlapping community detection\\%
\parbox{3cm}{STING}  STatistical Information Grid\\%
\parbox{3cm}{SVM} Support Vector Machine\\%
%
%
%
%
%
\parbox{3cm}{WEFCM} Weighted Entropy-regularized Fuzzy C-Means\\%
\parbox{3cm}{WFCM} Weighted Fuzzy C-Means\\%
\parbox{3cm}{WFD}  Weighted Feature Distance\\%
\parbox{3cm}{WLAC}  Weighted Locally Adaptive Clustering\\%
\parbox{3cm}{XML}  eXtensible Markup Language\\%
}
%
%
\newpage
\vspace*{-2mm}
{\bf \Large{Appendix B:}}\\[0.3cm]
{\bf \Large{Mathematical symbols}}\\[0.4cm]
\parbox{2.0cm}{$c$}  the number of clusters. \\
\parbox{2.0cm}{$c_i$}  the $i^{th}$ cluster. \\
\parbox{2.0cm}{$ d$}   the dimensionality of the search space. \\
\parbox{2.0cm}{$D(X_1,X_2)$}   the distance between two objects $X_1$ and $X_2$. \\
\parbox{2.0cm}{$D(P_j,Q_j)$}   distance between two objects or probability measures. \\
\parbox{2.0cm}{$ \mathscr{D}_i$}  the $i^{th}$ department for the student (teacher) example. \\
\parbox{2.0cm}{$Err$}   error. \\
\parbox{2.0cm}{$G(V,E)$} a graph with a vertex set $V$ and a edge set $E$. \\
\parbox{2.0cm}{$J_m(U,V)$}   Fuzzy C-Means (FCM) function. \\
\parbox{2.0cm}{$ L_1$}   the 1-norm. \\
\parbox{2.0cm}{$ L_2$}   the 2-norm. \\
\parbox{2.0cm}{$ L_p$}   the p-norm. \\
\parbox{2.0cm}{$ \mathscr{L}_i$}  the $i^{th}$ line for the crossing line example. \\
\parbox{2.0cm}{$ m$}   the fuzzification constant or different norms. \\
\parbox{2.0cm}{$ m_i$}  the $i^{th}$ centroid in k-means algorithms. \\
\parbox{2.0cm}{$ M_{hcn}$}   the crisp membership function. \\
\parbox{2.0cm}{$M_{fcn}$}   the fuzzy membership function. \\
\parbox{2.0cm}{$M_{pcn}$}   the possibilistic membership function. \\
\parbox{2.0cm}{$M_{bfpm}$}   the BFPM membership function. \\
\parbox{2.0cm}{$n$}  the number of objects in each dataset.\\
\parbox{2.0cm}{$N_v$} null vector. \\
\parbox{2.0cm}{$N_{v}^{T}$}   the transpose of the vector. \\
\parbox{2.0cm}{$O_j$}  the $j^{th}$ object.\\
\parbox{2.0cm}{$O$}  a set of objects.\\
\parbox{2.0cm}{$SP$}   sequential pattern. \\
\parbox{2.0cm}{$u_{ij}$} membership of the $j^{th}$ object with respect to the $i^{th}$ cluster.\\
\parbox{2.0cm}{$u(O_j)$}  the membership of the $j^{th}$ object in fuzzy sets.\\
\parbox{2.0cm}{$ U$}  a set of membership degrees. \\
\parbox{2.0cm}{$ v_{i}$}  the $i^{th}$ centroid or prototype. \\
\parbox{2.0cm}{$ V$}  a set of centroids or prototypes from clusters. \\
\parbox{2cm}{$V_{pc}$} Partition Coefficient Validity index\\%
\parbox{2cm}{$V_{pe}$}  Partition Entropy Validity index\\%
\parbox{2cm}{$V_{xb}$}  Xie-Beni Validity index\\%
\parbox{2.0cm}{$x_f$}  the $f^{th}$ feature of the object X.\\
%
%
\parbox{2.0cm}{$w_d$}  the weight assigned to the $d^{th}$ feature. \\
\parbox{2.0cm}{$ || X ||$}   norms: $\sqrt{(X,X)}$. \\
\parbox{2.0cm}{$ || . ||_{A}$}   the A norm. \\
\parbox{2.0cm}{$\delta(X_j,X_l)$} a distance metric space between two objects. \\
\end{document}